%% file: main.tex
\pdfoutput=1
\documentclass[10pt,twocolumn,letterpaper]{article}
\PassOptionsToPackage{table}{xcolor}
\ifdefined\cvprsubmission
  \usepackage[review]{cvpr}
\else
  \usepackage[pagenumbers]{cvpr}
\fi
\usepackage[T1]{fontenc}
\usepackage{microtype}
\usepackage{multirow,makecell}
\usepackage{tikz}
\usetikzlibrary{arrows.meta,positioning,fit,backgrounds,calc,shapes.geometric,shapes.arrows}
\usepackage{tabularx}
\usepackage{needspace}
\usepackage{pifont}
\usepackage{float}
\floatstyle{ruled}\newfloat{algorithm}{tbp}{loa}\floatname{algorithm}{Algorithm}\floatstyle{plain}
\definecolor{cvprblue}{rgb}{0.21,0.49,0.74}
\usepackage[pagebackref,breaklinks,colorlinks,allcolors=cvprblue]{hyperref}
\def\confName{CVPR}
\def\confYear{2027}
\definecolor{ours}{RGB}{232,242,252}
\definecolor{citedgray}{gray}{0.5}
\newcommand{\cited}[1]{\textcolor{citedgray}{#1}}
\definecolor{trk}{RGB}{218,236,219}
\definecolor{vlm}{RGB}{252,234,214}
\definecolor{sym}{RGB}{235,230,245}
\input{figs/walk_common.tex}
\input{figs/gallery_common.tex}

\newcommand{\blfootnote}[1]{\begingroup\renewcommand{\thefootnote}{}\footnote{#1}\addtocounter{footnote}{-1}\endgroup}

\title{Ask About the Right Moment: Training-Free Object Identity Reasoning\\for Video-Language Models}
\author{Di Yang$^{1, \dagger}$\qquad Xiaochang Li$^{1}$\qquad Muyang Liu$^{1}$\qquad Yufei Wang$^{2,*}$\qquad Yanhai Xiong$^{1,*}$\\[4pt]
$^{1}$College of William \& Mary\qquad $^{2}$SparcAI Inc.\\
{\tt\small dyang06@wm.edu\quad xli59@wm.edu\quad mliu22@wm.edu\quad yxiong05@wm.edu\quad yufei.wang@sparclab.ai}}

\newif\iftairfilelist
\iftairfilelist
\includegraphics{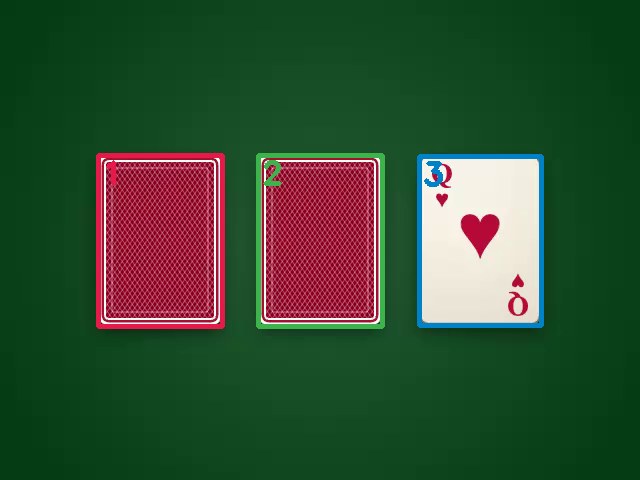}
\includegraphics{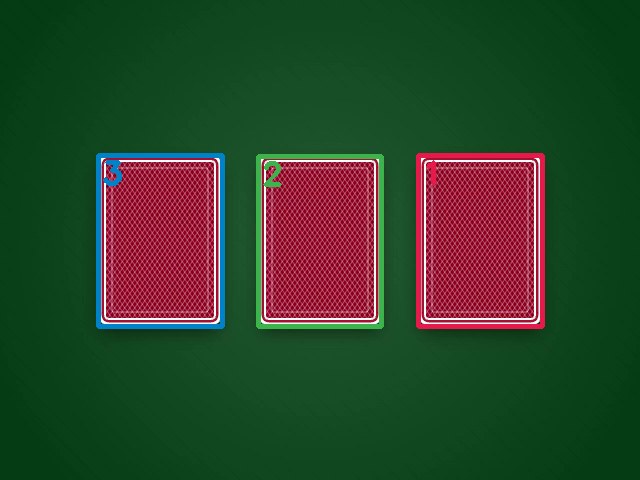}
\includegraphics{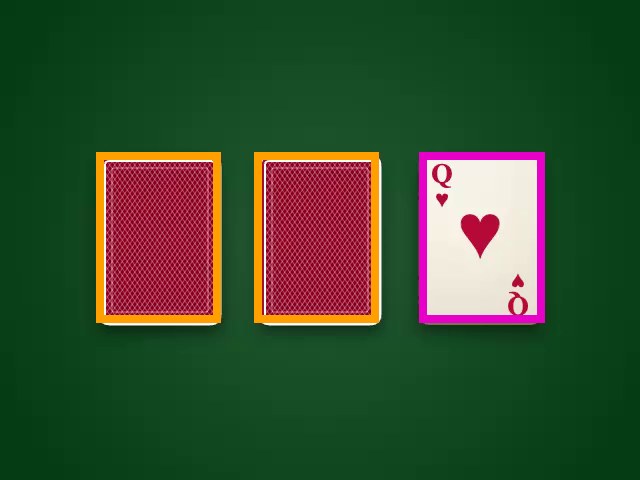}
\includegraphics{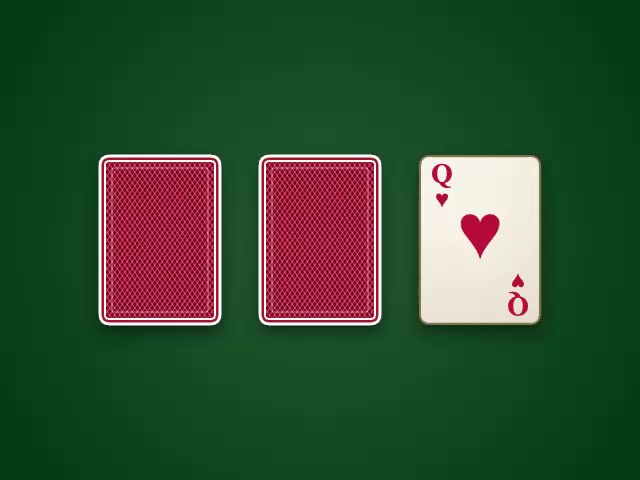}
\includegraphics{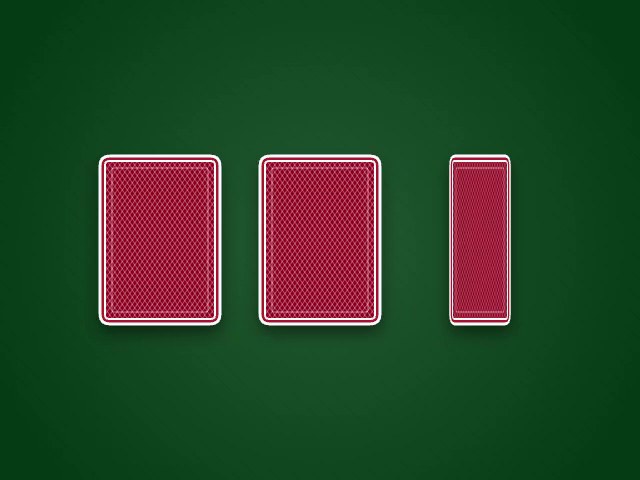}
\includegraphics{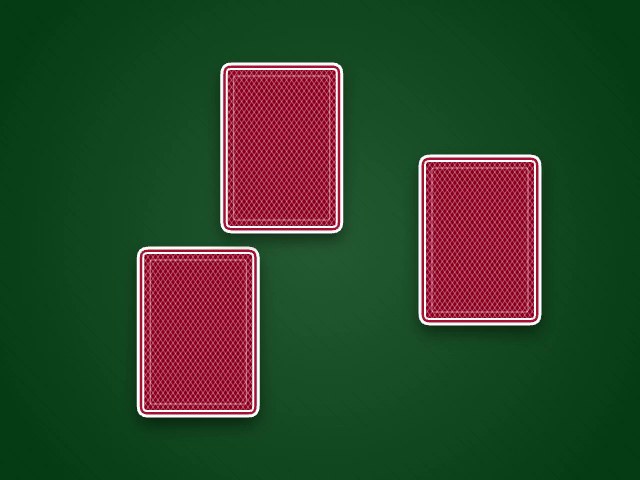}
\includegraphics{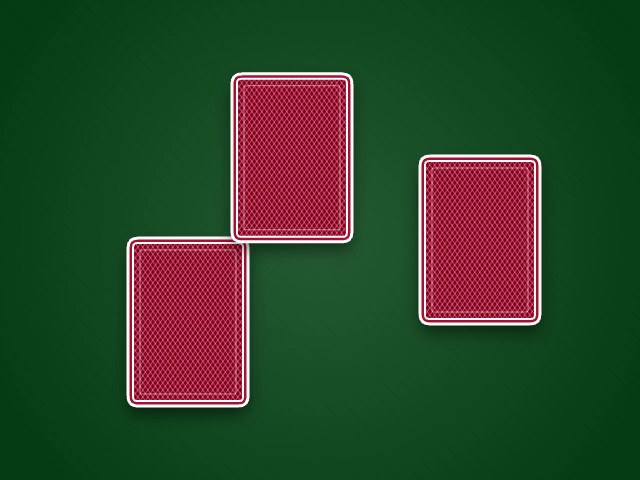}
\includegraphics{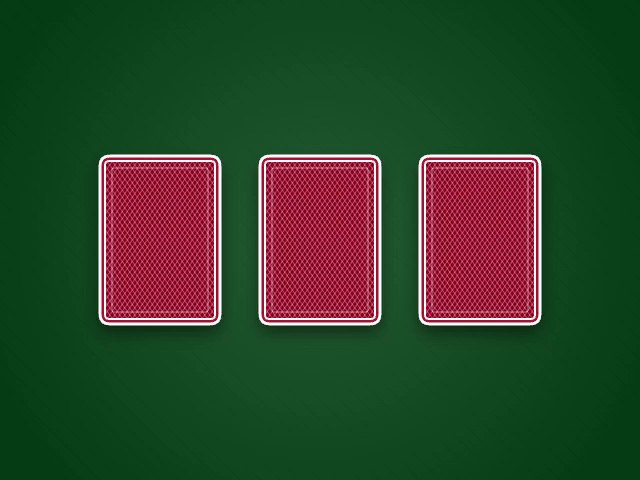}
\includegraphics{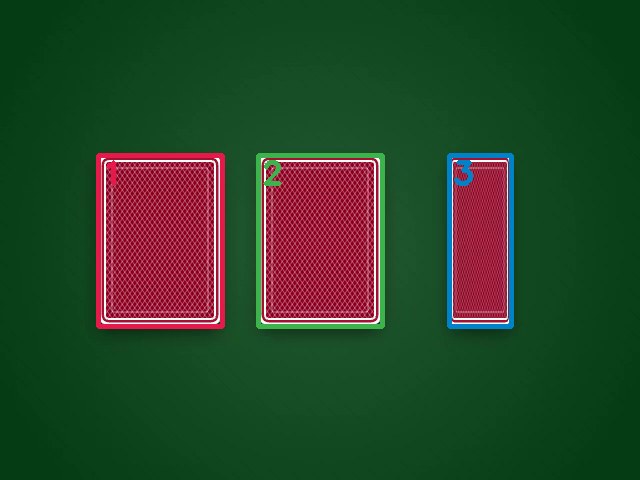}
\includegraphics{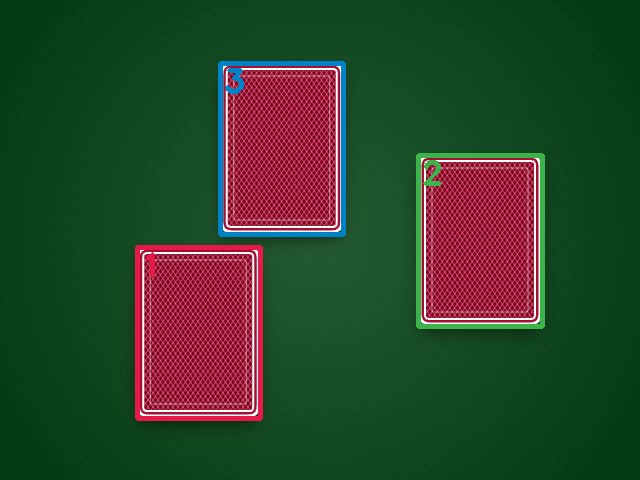}
\includegraphics{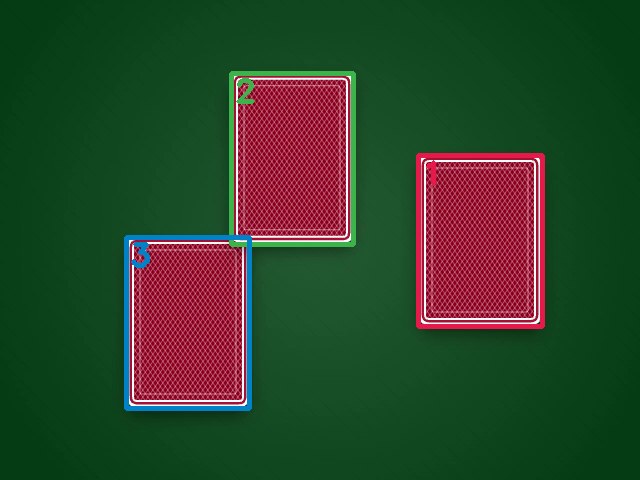}
\includegraphics{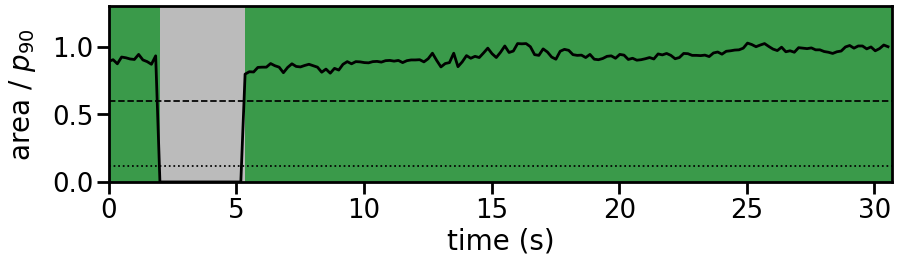}
\includegraphics{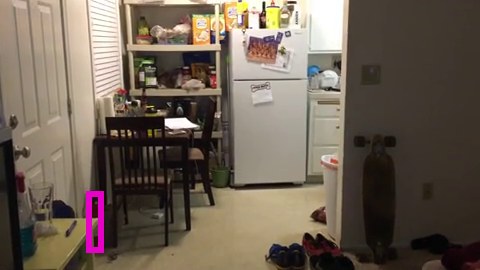}
\includegraphics{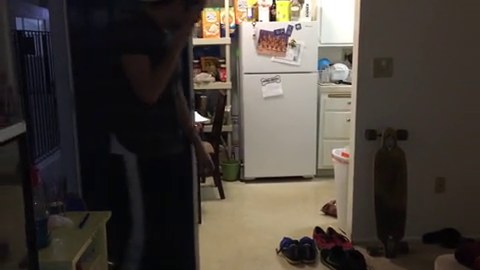}
\includegraphics{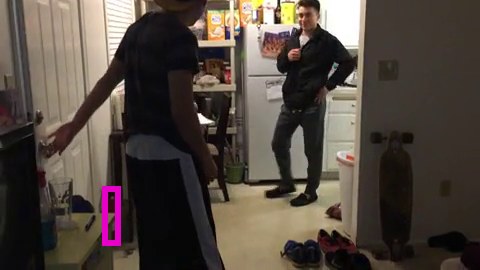}
\includegraphics{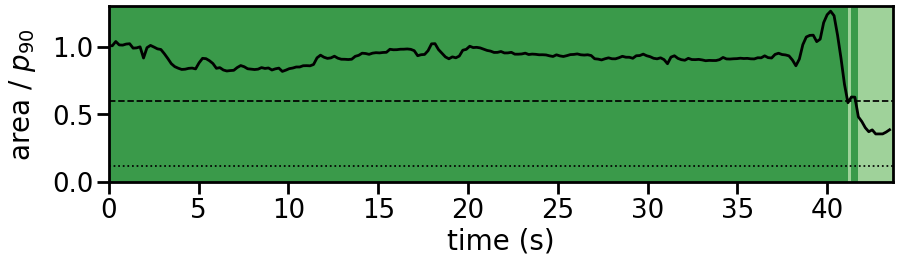}
\includegraphics{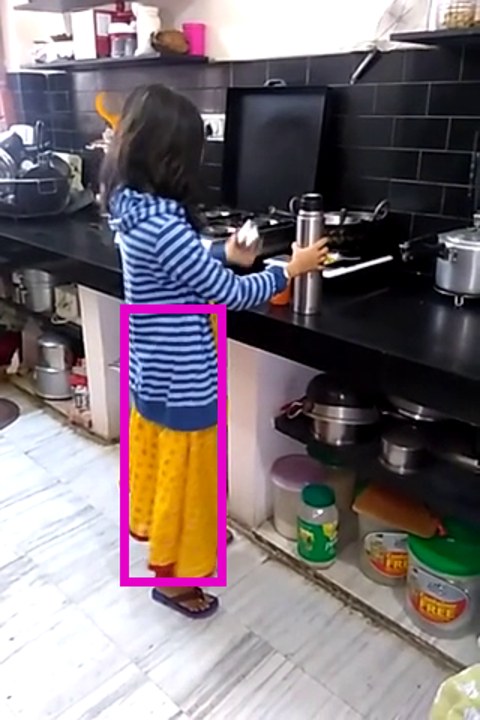}
\includegraphics{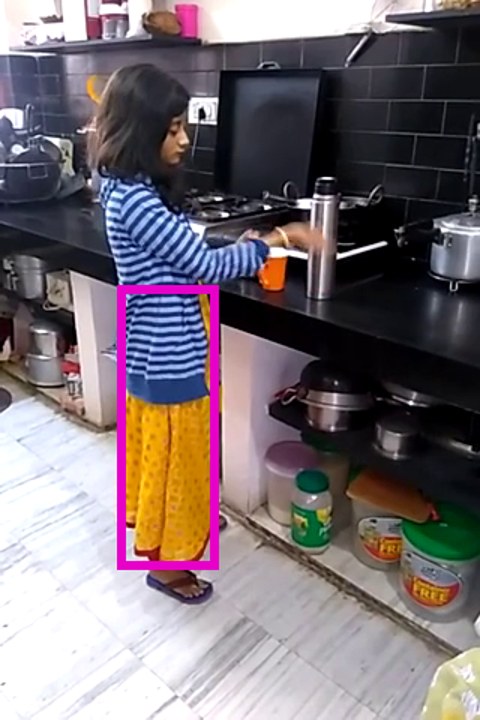}
\includegraphics{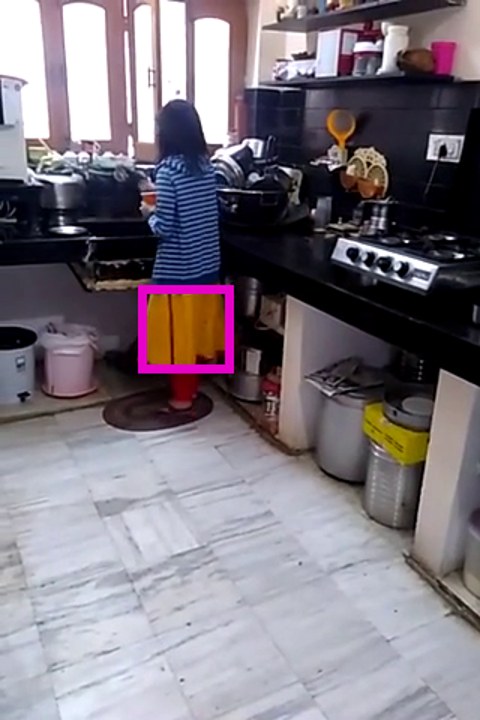}
\includegraphics{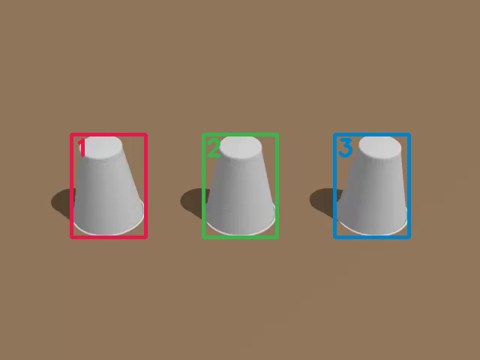}
\includegraphics{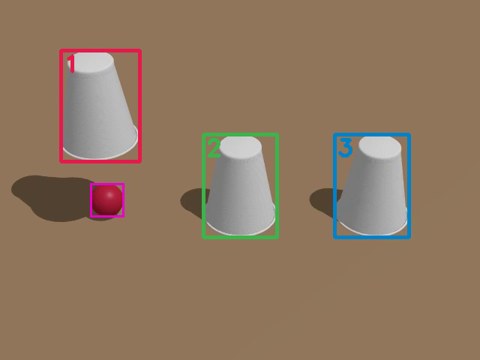}
\includegraphics{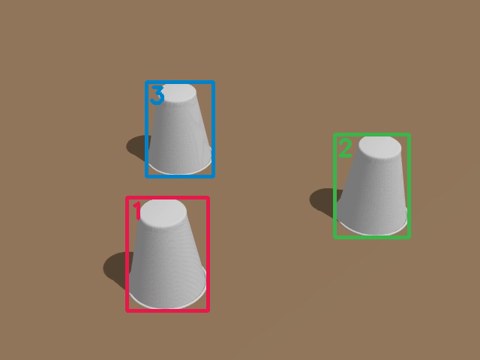}
\includegraphics{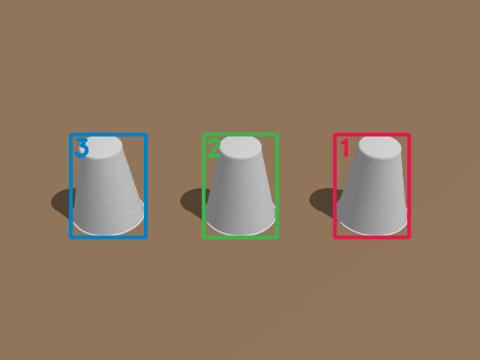}
\includegraphics{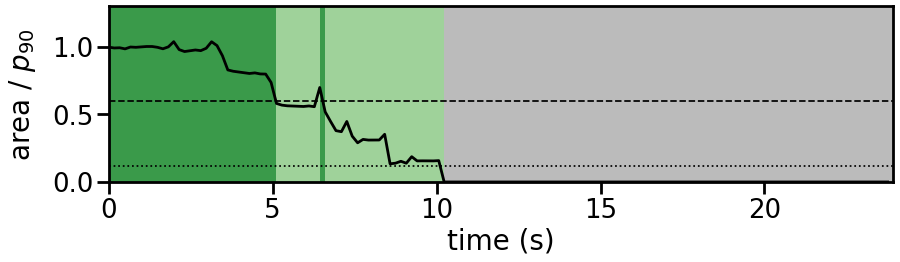}
\includegraphics{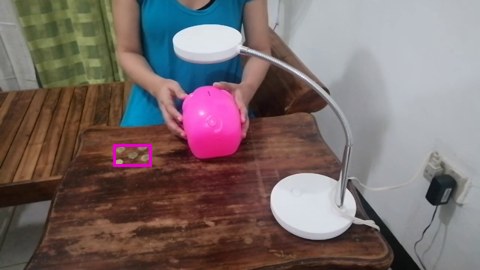}
\includegraphics{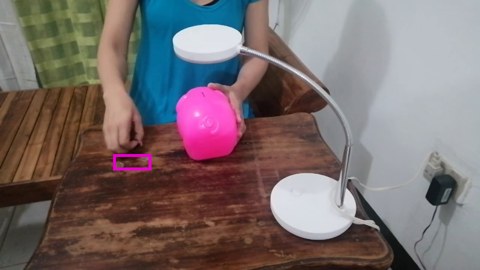}
\includegraphics{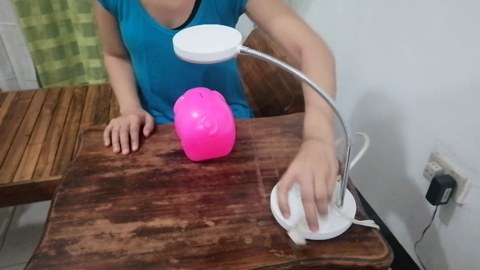}
\includegraphics{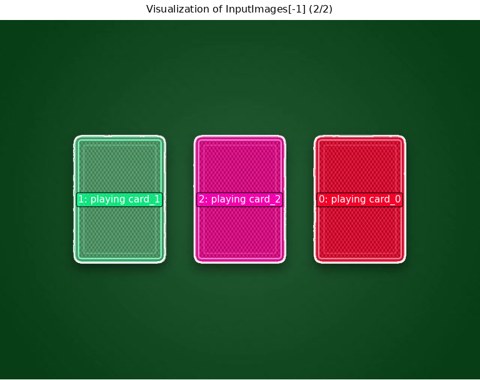}
\includegraphics{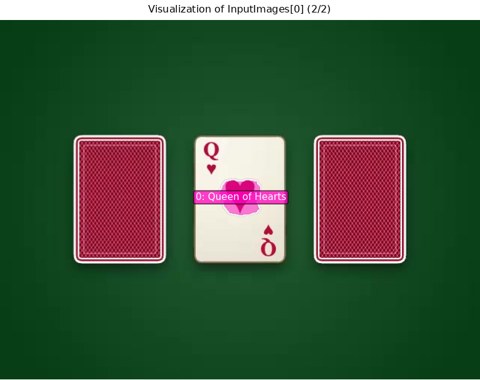}
\includegraphics{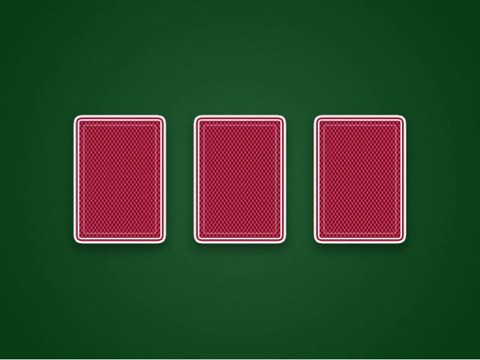}
\includegraphics{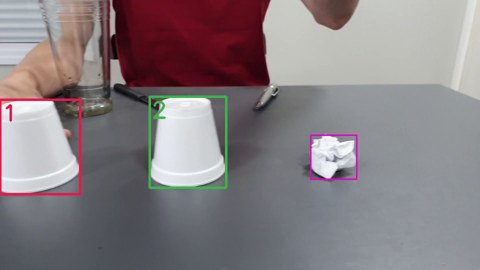}
\includegraphics{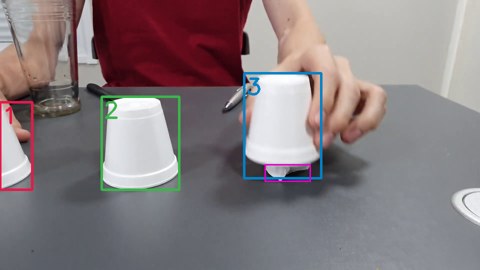}
\includegraphics{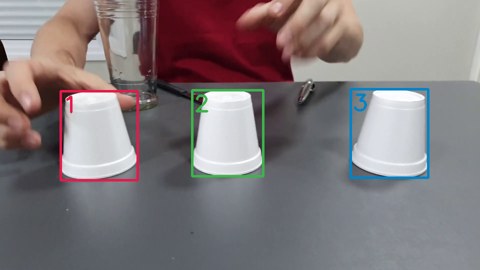}
\includegraphics{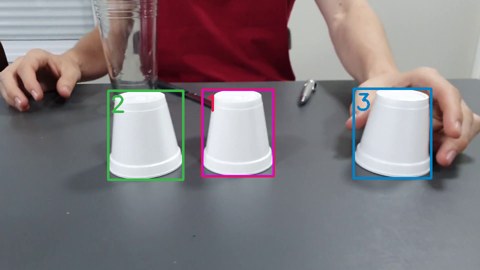}
\includegraphics{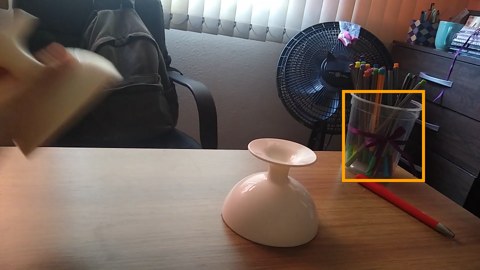}
\includegraphics{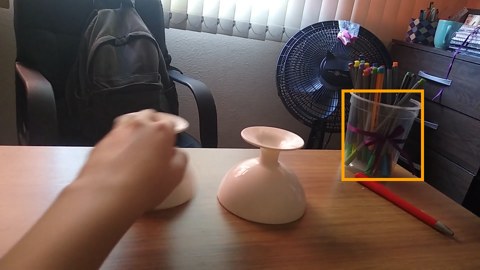}
\includegraphics{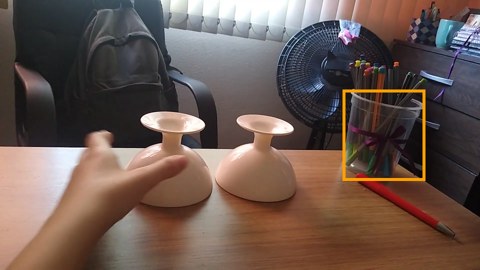}
\includegraphics{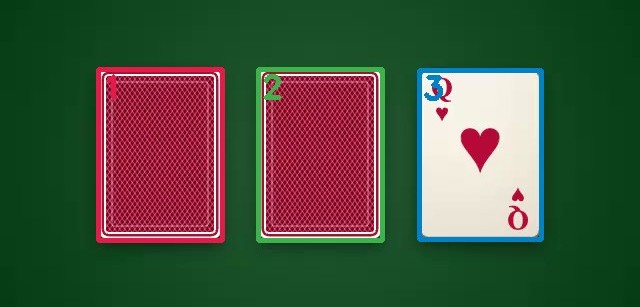}
\includegraphics{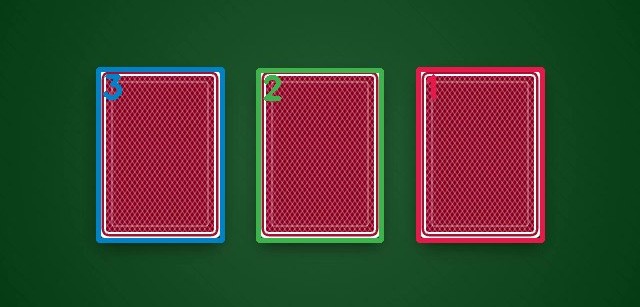}
\includegraphics{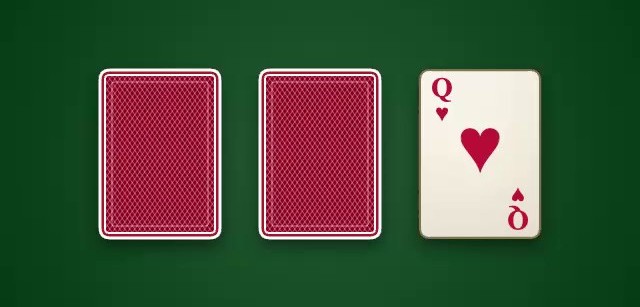}
\includegraphics{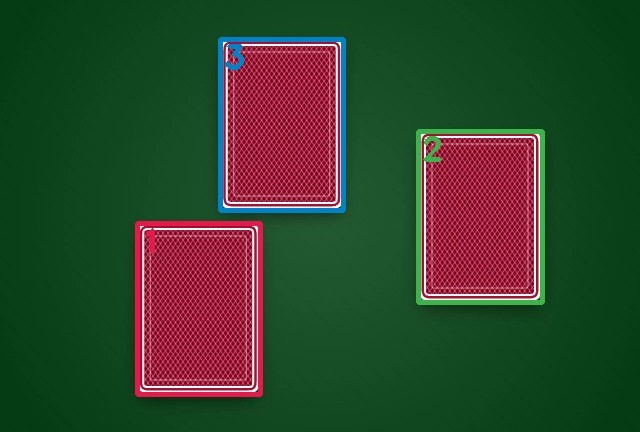}
\includegraphics{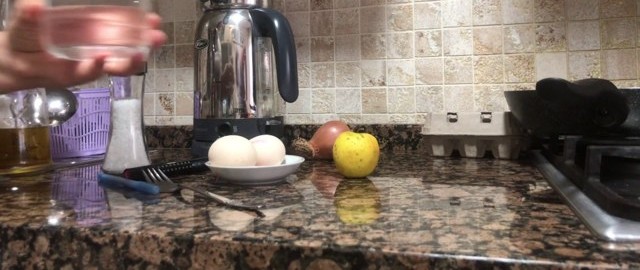}
\includegraphics{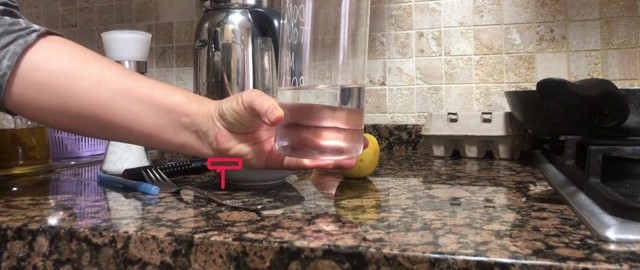}
\includegraphics{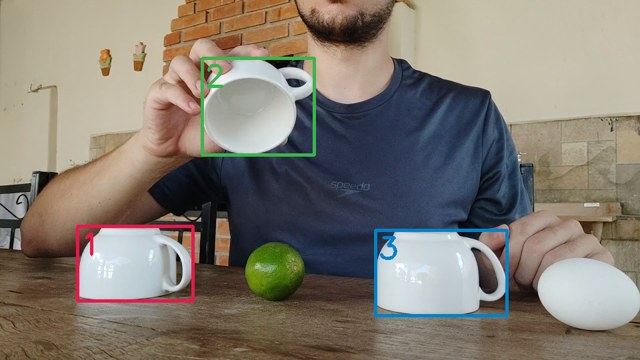}
\includegraphics{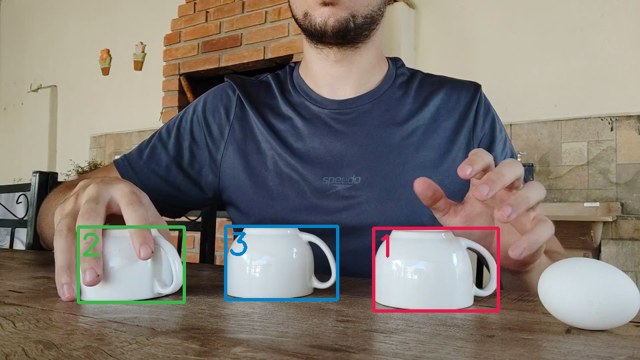}
\includegraphics{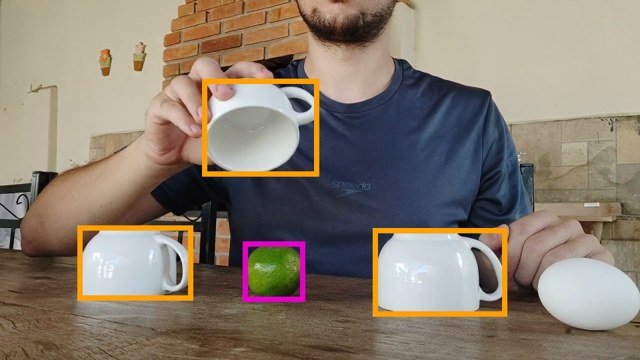}
\includegraphics{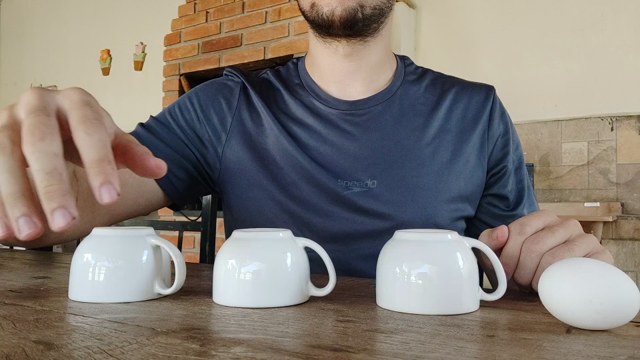}
\includegraphics{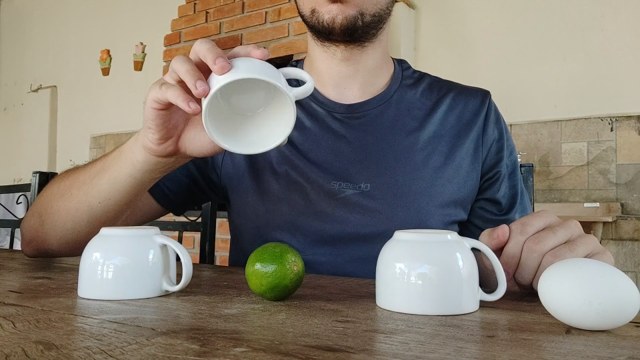}
\includegraphics{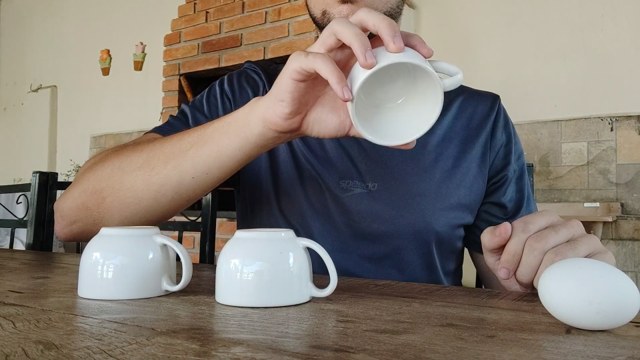}
\includegraphics{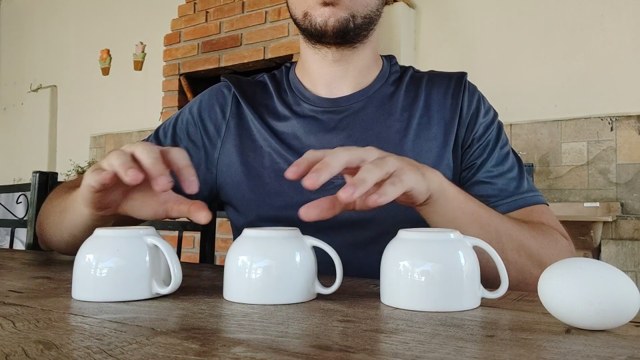}
\includegraphics{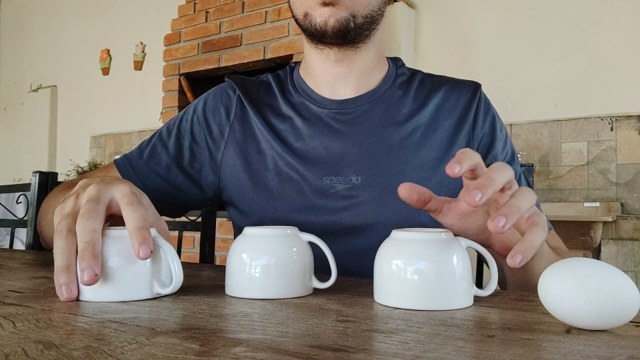}
\includegraphics{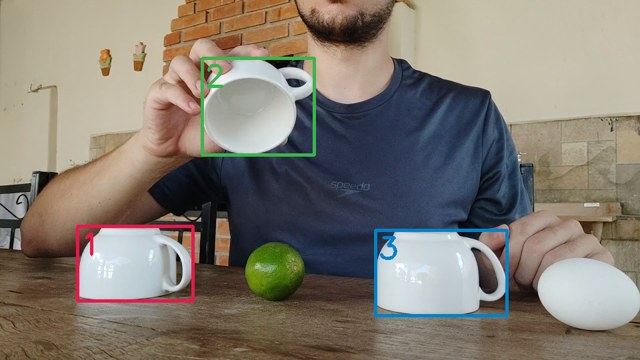}
\includegraphics{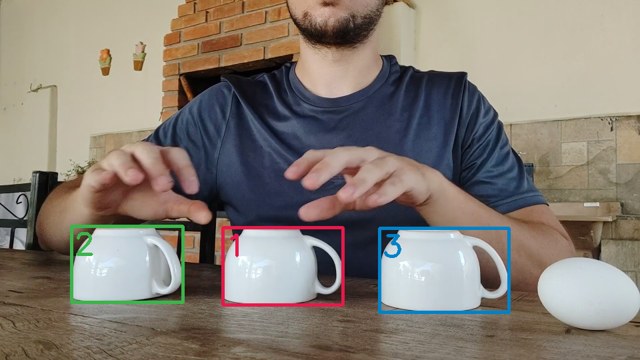}
\includegraphics{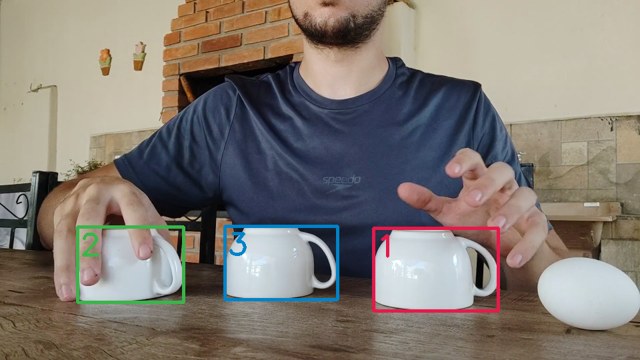}
\includegraphics{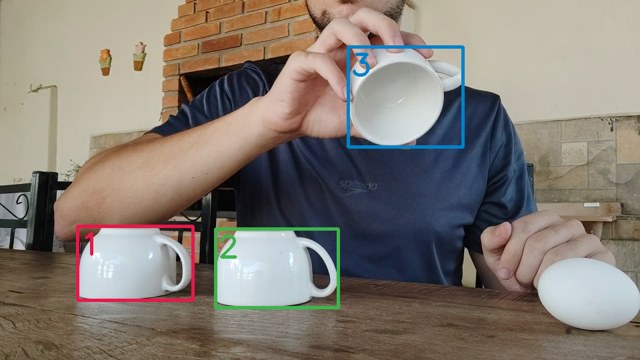}
\includegraphics{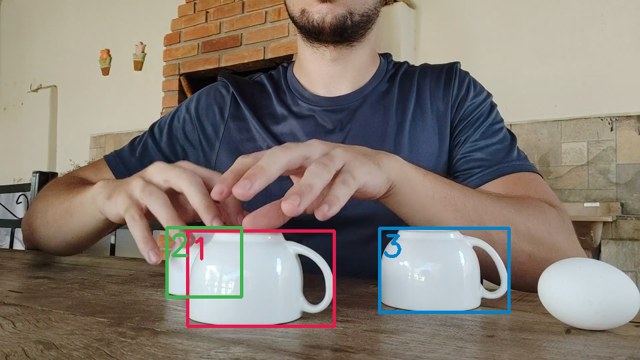}
\includegraphics{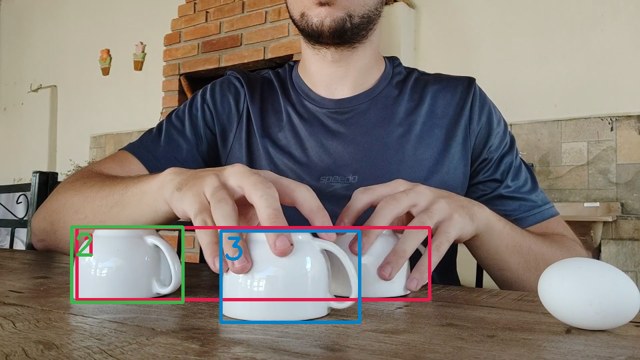}
\includegraphics{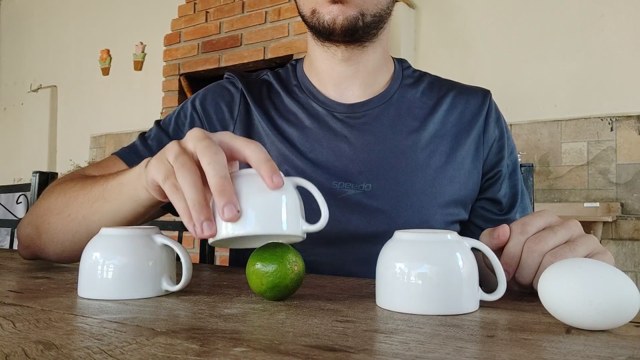}
\includegraphics{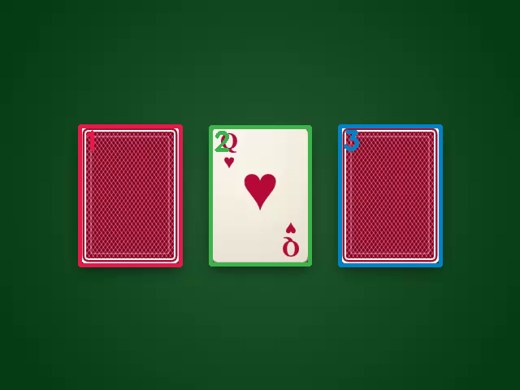}
\includegraphics{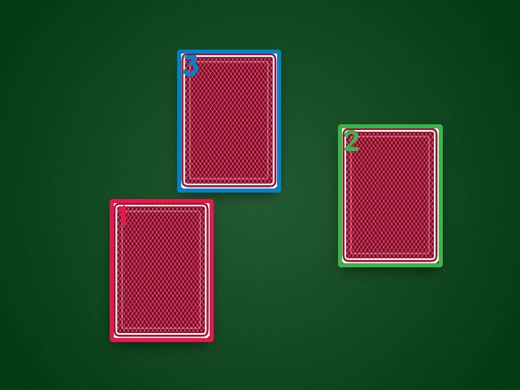}
\includegraphics{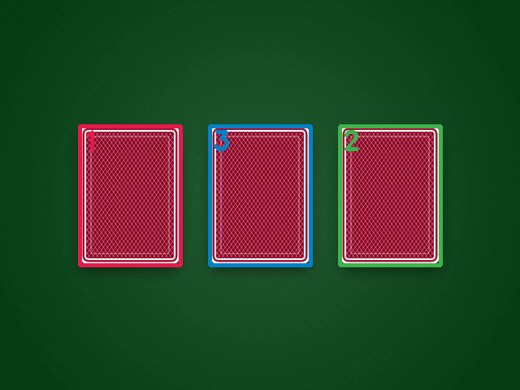}
\includegraphics{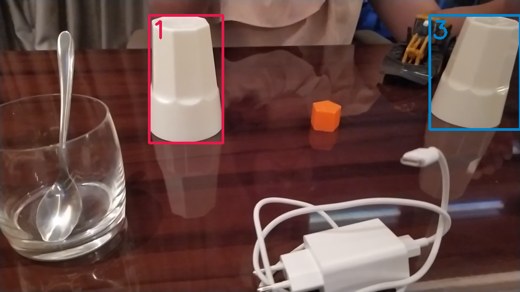}
\includegraphics{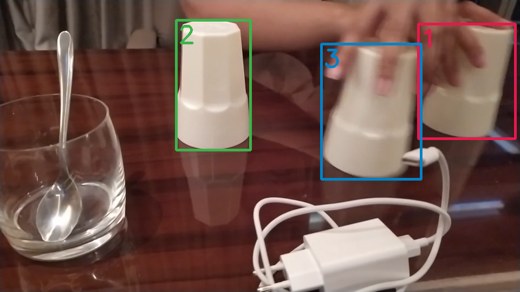}
\includegraphics{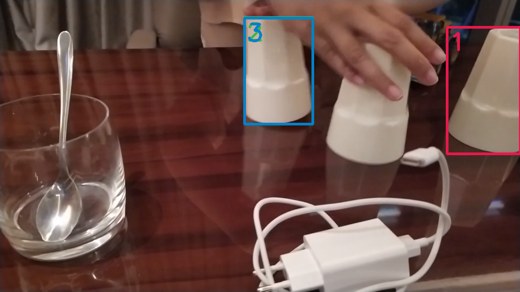}
\includegraphics{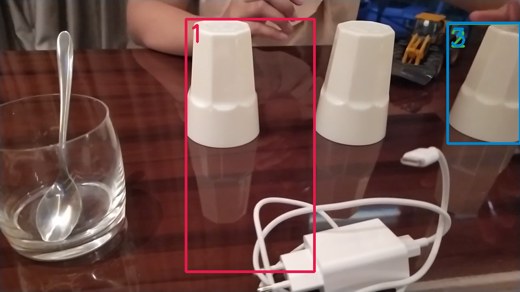}
\includegraphics{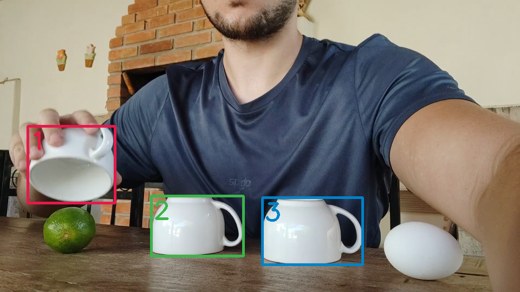}
\includegraphics{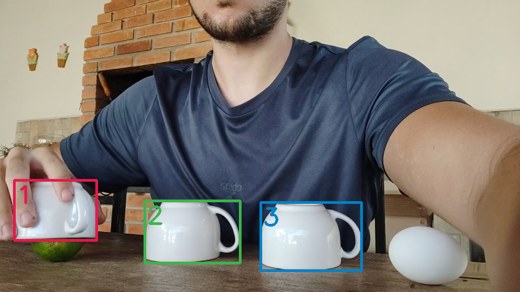}
\includegraphics{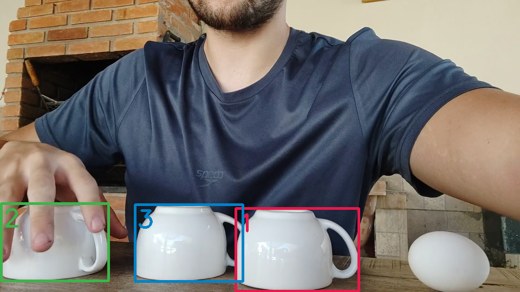}
\includegraphics{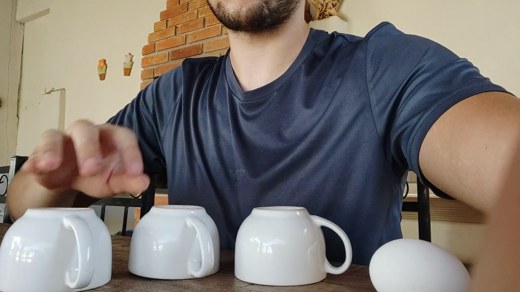}
\includegraphics{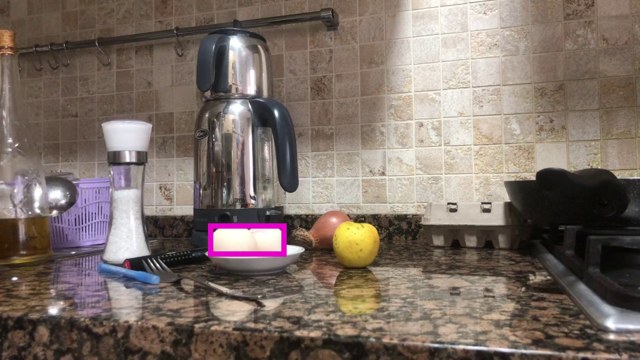}
\includegraphics{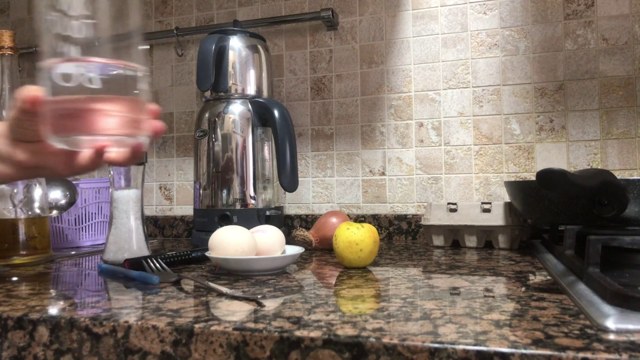}
\includegraphics{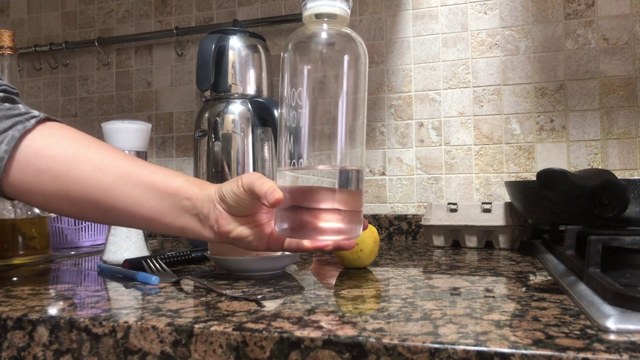}
\includegraphics{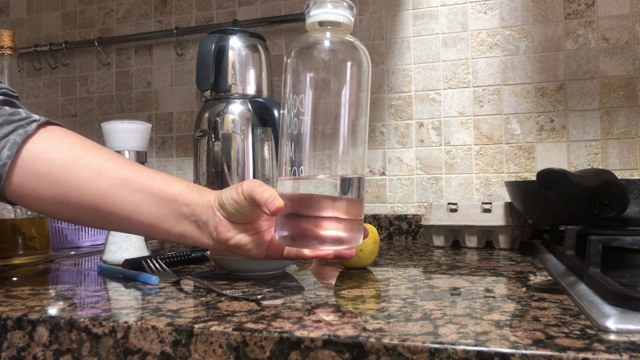}
\includegraphics{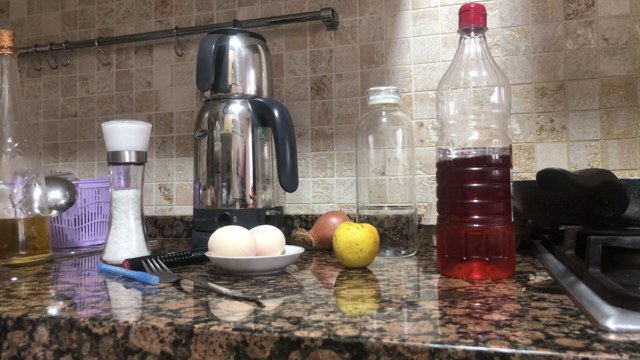}
\includegraphics{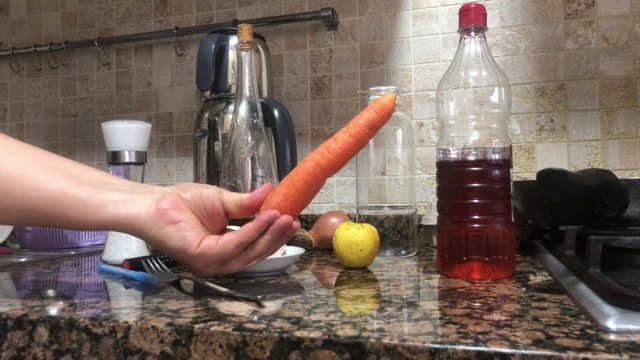}
\includegraphics{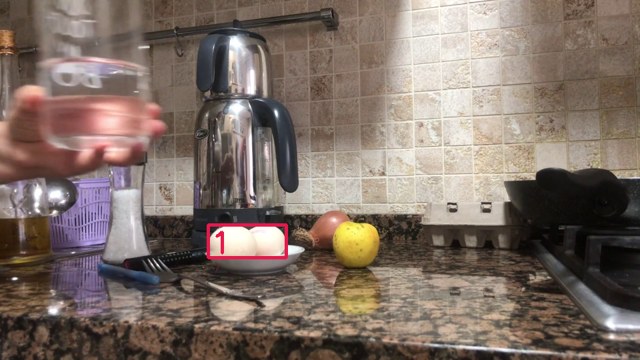}
\includegraphics{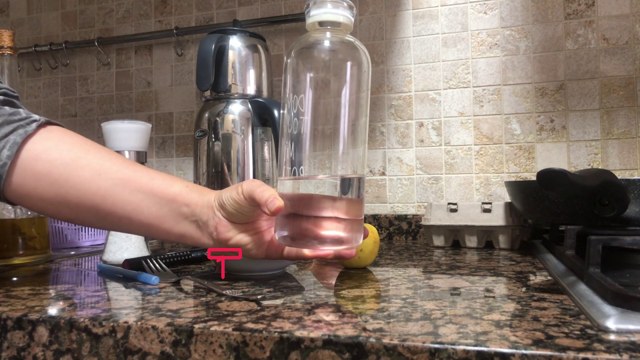}
\includegraphics{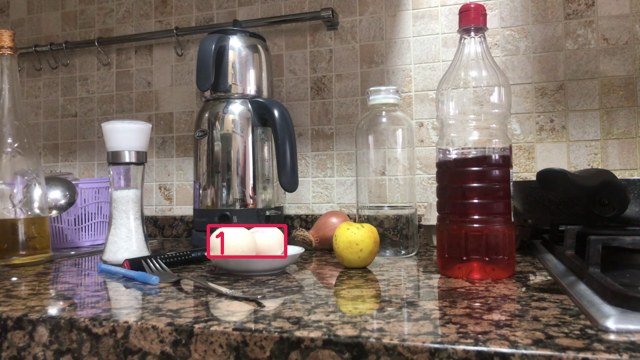}
\fi
\begin{document}
\twocolumn[{\renewcommand\twocolumn[1][]{#1}\maketitle
\begin{center}
\resizebox{0.95\textwidth}{!}{\input{figs/overview.tex}}
\captionof{figure}{\textbf{Bind identity, then follow the track} (Perception Test \texttt{video\_6159}, test set, Qwen3-VL-8B). A lime is visible under the middle mug before three identical mugs are shuffled. The direct answer of a video-language model (Video-LM) is that it stayed in the middle. Our agent asks which mug held the lime just after it was hidden, binds the answer to track \#2, and reads its final position from that track. Fig.~\ref{fig:pipeline} shows the full method.}
\label{fig:overview}
\end{center}\vspace{0.5em}}]
\ifdefined\cvprsubmission\else
\blfootnote{$^{*}$Corresponding authors.\quad $^{\dagger}$Project lead.}
\blfootnote{\textit{Work in progress.}}
\fi

\begin{abstract}
Many questions about a video concern one physical object. Where does the ball end up after the cups are shuffled? Which card is the queen? How often was the mug covered? Open video-language models (Video-LMs) recognise such objects in single frames but do not keep them apart over time. On shuffled shell games, open models from eight families stay near chance, and so do published tool agents. We separate semantic identity binding from temporal identity maintenance. Binding should happen at a moment when the evidence for the object is visible, and maintenance should be left to a tracker. \emph{Track, Then Bind} implements this principle without training. A promptable video tracker follows every candidate object and draws it with a track number. A frozen vision-language model (VLM) answers one question about the identifying evidence, which binds the target to a track. The answer is then derived from a text description of the tracks. With Qwen3-VL-32B, the agent solves both VET-Bench tasks, where a fine-tuned trajectory model reaches 92\% and 90\%. On real cup games from the Perception Test, it raises accuracy from 35.5\% to 52.3\%. Without retuning, the same configuration also solves VET-Bench with Gemma 4 and GPT-5.5. Its track-derived visibility timelines add 8.8--17.9 points on TOC-Bench for open models. GPT-6.1 already answers 92.5\% of the real cup games directly but 71.0\% through our pipeline, while our timelines still raise its TOC-Bench accuracy from 52.5\% to 57.7\%. The benefit of external tracking thus depends on the underlying model and the task.
\end{abstract}

\begin{figure*}[t]
\centering
\resizebox{0.95\textwidth}{!}{\input{figs/pipeline.tex}}
\caption{\textbf{Method.} A frozen vision-language model (VLM) plans the question (1) and boxes its entities, which SAM2.1 tracks as numbered instances (2). If a target must be told apart from look-alikes or containers, the VLM answers one question on a clip with numbered tracks (3--4). The answer is then read out from a text description of the tracks (5a; VET-Bench card monte \texttt{card\_003}). Otherwise the tracks become a visibility timeline that is appended to the benchmark prompt (5b; TOC-Bench \texttt{8b0c33d38075}). Both examples are logged test runs with Qwen3-VL-8B.}
\label{fig:pipeline}
\end{figure*}

\begin{table*}[t]
\centering
\small
\setlength{\tabcolsep}{5pt}
\caption{\textbf{Accuracy (\%) on the test sets.} VET: VET-Bench, 50 videos per task. PT: Perception Test cup games filtered by the VET-Bench authors (107 videos); PT-65: its 65 videos with three cups and at least one swap. TOC: TOC-Bench test split (1,845 questions). \cited{Gray numbers are cited} from other papers. The VET, PT and PT-65 accuracies of Gemini-3-Pro, Doubao-Seed-2.0-mini and GLM-4.6V-Flash are from VET-Bench~\cite{vetbench}, rescored from the predictions its authors released. The TOC accuracy of Gemini is from TOC-Bench~\cite{tocbench}. All black numbers are our runs with each benchmark's prompt, frames and scorer. Best result per column in bold, best without a closed model underlined.}
\label{tab:main}
\begin{tabular}{llccccc}
\toprule
Method & Type & VET cup & VET card & PT & PT-65 & TOC \\
\midrule
Random & -- & 33.3 & 33.3 & 33.3 & 33.3 & 30.3\\
\midrule
\multicolumn{7}{l}{\emph{Closed Video-LMs}}\\
Gemini-3-Pro~\cite{gemini3pro} & thinking (API default)$^{\S}$ & \cited{34.0} & \cited{40.0} & \cited{36.4} & \cited{30.8} & \cited{43.9}$^{\ast}$\\
Doubao-Seed-2.0-mini~\cite{seed2} & API default$^{\S}$ & \cited{30.0} & \cited{32.0} & \cited{46.7} & \cited{40.0} & --\\
GPT-5.5~\cite{gpt55} & reasoning, medium effort & 94.0 & 96.0 & 68.2 & 72.3 & 47.6\\
GPT-6.1~\cite{gpt61} & reasoning, medium effort & \textbf{100} & \textbf{100} & \textbf{92.5} & \textbf{90.8} & \textbf{52.5}\\
\midrule
\multicolumn{7}{l}{\emph{Open Video-LMs}}\\
Qwen3-VL-8B~\cite{qwen3vl} & direct answer & 28.0 & 28.0 & 33.6 & 27.7 & 25.1\\
Qwen3-VL-32B~\cite{qwen3vl} & direct answer & 32.0 & 38.0 & 35.5 & 33.8 & 37.1\\
Qwen3-VL-235B-A22B~\cite{qwen3vl} & direct answer & 22.0 & 38.0 & 38.3 & 27.7 & 32.8\\
InternVL3.5-8B~\cite{internvl35} & direct answer & 32.0 & 32.0 & 43.0$^{\ddagger}$ & 38.5$^{\ddagger}$ & 31.3\\
Gemma 4 31B~\cite{gemma4} & direct answer & 36.0 & 28.0 & 36.4 & 32.3 & 40.0\\
GLM-4.6V-Flash~\cite{glm46v} & thinking (API default)$^{\S}$ & \cited{34.0} & \cited{28.0} & \cited{43.9} & \cited{41.5} & --\\
Molmo2-8B~\cite{molmo2} & direct answer & 30.0 & 40.0 & 33.6 & 30.8 & 32.3\\
\midrule
\multicolumn{7}{l}{\emph{Reasoning-enhanced and tool-using}}\\
Qwen3-VL-8B-Thinking~\cite{qwen3vl} & thinking & 28.0 & 24.0 & 37.4 & 36.9 & 30.0\\
Qwen3-VL-32B-Thinking~\cite{qwen3vl} & thinking & 24.0 & 22.0 & 33.6 & 32.3 & 36.9\\
Molmo2-SGCoT~\cite{vetbench} & trajectory chain-of-thought, fine-tuned & 92.0 & 90.0 & 32.7 & 38.5 & 39.2\\
Open-o3-Video-v2-8B~\cite{openo3v2} & reinforcement-learned crop tool & 26.0 & 30.0 & 38.3 & 33.8 & 33.9$^{\dagger}$\\
SpatialClaw~\cite{spatialclaw} (Qwen3-VL-32B) & code agent, SAM3 + depth & 28.0 & 36.0 & 26.2 & 30.8 & 35.7\\
VideoAgent~\cite{videoagent} (Qwen3-VL-32B) & object-memory agent & 40.0 & 28.0 & 37.4 & 36.9 & 33.0\\
VideoINSTA~\cite{videoinsta} (Qwen3-VL-32B) & clip-reasoning agent & 42.0 & 28.0 & 42.1 & 38.5 & 34.5\\
\midrule
\rowcolor{ours}\textbf{Track, Then Bind} (Qwen3-VL-8B) & training-free agent & \textbf{100} & \textbf{100} & 51.4 & 52.3 & 43.0\\
\rowcolor{ours}\textbf{Track, Then Bind} (Qwen3-VL-32B) & training-free agent & \textbf{100} & \textbf{100} & \underline{52.3} & \underline{55.4} & 48.6\\
\rowcolor{ours}\textbf{Track, Then Bind} (Gemma 4 31B) & training-free agent & \textbf{100} & \textbf{100} & 48.6 & 49.2 & \underline{48.8}\\
\bottomrule
\end{tabular}\\[2pt]
\begin{minipage}{0.97\textwidth}\scriptsize
--: not evaluated. $^{\ast}$Reported on all 2,323 TOC-Bench questions (Gemini: version 3.1-Pro~\cite{gemini31pro}). $^{\dagger}$The official evaluator cannot express the 89 ordering questions, which count as wrong (35.6\% on the other 1,756). Random: uniform guessing over each question's answer space. $^{\S}$Run by the VET-Bench authors with each API's default thinking setting. $^{\ddagger}$InternVL3.5 omits the required \texttt{\textbackslash boxed\{\}} answer in 83 of 107 direct answers, so we read its first option letter; the VET-Bench scorer gives 9.3 (PT) and 7.7 (PT-65). Baseline settings: Appendices~\ref{app:baselines} and~\ref{app:families}.
\end{minipage}
\end{table*}

\section{Introduction}
Many questions about a video are questions about one physical object. Which cup hides the ball after the shuffle? Which card is the queen once all three are face down? Did the mug leave the frame or was it covered before it reappeared? Such questions arise wherever objects are moved, stacked or put away. Examples are assistants that remember where an item was left, robots whose own arm hides the object they manipulate, and laboratory procedures with rows of identical tubes. Answering them requires following one object through occlusion, containment and look-alike copies.

Open Video-LMs fail at this. On VET-Bench~\cite{vetbench}, open models from eight families answer 22--40\% of the shell games, around the chance level of 33\% (Appendix~\ref{app:families}). They include Qwen3-VL up to 235B parameters, Qwen3.5-397B, Gemma 4 and InternVL3.5. Only a model fine-tuned on synthetic trajectories (SGCoT) escapes, and it returns to chance on real cup-game videos from the Perception Test~\cite{perceptiontest}. Thinking variants, a reinforcement-learned tool agent and three training-free tool agents also stay near chance (Table~\ref{tab:main}). Closed models differ. Gemini-3-Pro stays near chance, while GPT-5.5 answers 94--96\% of VET-Bench and 68.2\% of the real cup games. GPT-6.1 answers 100\% and 92.5\%. It also answers all card-monte videos that we render ourselves. On the cup games mirrored left to right it keeps 91.3\%, which argues against memorised answers. The capability can therefore be learned end to end, but none of the open models we evaluated has it.

A central difficulty is that the model must re-derive which object is which from a sparse sample of frames. Identical objects that cross or disappear leave almost no visual trace from which to do so. A promptable video tracker such as SAM2~\cite{sam2} has the opposite strengths. It follows each object mask through motion and brief occlusion, but does not know which object a question refers to. We study two design choices in combining them. \emph{When} should the system ask the VLM about an object's identity? A hidden ball can only be recognised before it is hidden, a card only while it is face up. Once the identity is fixed, \emph{what} should decide the answer: the VLM's reading of the video or the tracked state?

We answer them with two design rules and test each in isolation. First, the VLM is asked about an object's identity only through a static question about the moment at which the identity can be seen. We call this moment the evidence moment. A tracker then carries the identity, and the VLM is not asked again. Second, once the target is bound, the answer is read from the tracked state as text, without showing the video again. The tracked state is the position of each track at the end of the video. \emph{Track, Then Bind} (Fig.~\ref{fig:overview}) implements both rules without training. Every candidate carries a track number on the clip that the VLM sees, so its answer binds the target to one track. A plan made by the VLM decides which objects to follow and how the target can be recognised. As an extension to visibility and event questions that need no identity binding, the same framework supplies a visibility timeline built from the tracks. One program and one set of thresholds serve all benchmarks.

\noindent Our contributions are:
\begin{itemize}
\item \textbf{A diagnosis across model families.} Open Video-LMs from eight families, their thinking variants, a reinforcement-learned tool agent and three training-free ones stay near chance on shuffled cup games. Among closed models, GPT-5.5 and GPT-6.1 are far above chance; GPT-6.1 keeps its accuracy on mirrored videos.
\item \textbf{Binding through a static question about the evidence moment.} With the tracks and the clip held fixed, asking which cup the target is under right after it is hidden, instead of where it is at the end, raises accuracy on every test set (Table~\ref{tab:mech}). At 8B it lifts VET-Bench from near chance to 100\% and the real cup games by 15.0 points. Drawing the same tracks and asking the question directly stays at 32.7--44.0\% at 8B, against 51.4--100\% for the agent, and 12.1 points below the agent on real cup games at 32B.
\item \textbf{Answering from the tracked state.} Given the same tracked state together with the shuffled video, the VLM follows its own reading of the video. The synthetic cup game drops from 100\% to 38.0\% at 8B and to 76.0\% at 32B, and real cup games drop by 3.7--4.7 points.
\item \textbf{One configuration across benchmarks and VLM families.} The agent is tuned once on development videos and runs unchanged with Qwen3-VL, InternVL3.5, Gemma 4, GPT-5.5 and GPT-6.1. It solves VET-Bench with three of these families. Its track-derived timelines raise TOC-Bench accuracy for every open model, by 8.8--17.9 points.
\end{itemize}

\section{Related Work}
\textbf{Video-LMs and their temporal limits.} Video-LMs pair a visual encoder with a language model and are tuned on video instructions~\cite{videochatgpt,videollava,llavaov}. Current open families such as Qwen2.5-VL~\cite{qwen25vl}, InternVL3~\cite{internvl3}, Qwen3-VL~\cite{qwen3vl} and Molmo2~\cite{molmo2} add dynamic frame rates, timestamps and grounding outputs. Benchmarks that isolate time find that much of their accuracy comes from single frames. MVBench~\cite{mvbench} and NExT-QA~\cite{nextqa} probe temporal and causal actions, EgoSchema~\cite{egoschema} and Video-MME~\cite{videomme} long videos, TempCompass~\cite{tempcompass} finds that many models barely exceed chance once single-frame and language shortcuts are removed, and TOMATO~\cite{tomato} measures a 57-point gap to humans on questions that need the frame order. VELOCITI~\cite{velociti} finds weak binding of agents to actions across events, and VSI-Bench~\cite{vsibench} limited visual-spatial reasoning. VET-Bench~\cite{vetbench} and TOC-Bench~\cite{tocbench} target object identity directly, the setting of this paper.

\textbf{Reasoning-enhanced Video-LMs.} Chain-of-thought prompting~\cite{cot} and its video variants~\cite{vot} write intermediate steps before the answer, and self-consistency~\cite{selfconsistency} votes over sampled chains. On VSI-Bench neither improves spatial answers~\cite{vsibench}. Reinforcement-learned models such as Video-R1~\cite{videor1} and Open-o3-Video-v2~\cite{openo3v2} add textual or grounded reasoning traces. SGCoT~\cite{vetbench} fine-tunes Molmo2 to write object trajectories before answering, and RGA3~\cite{rga3} trains a Video-LM to refer to and ground objects with spatio-temporal prompts. In all of these, identity is held in learned representations or generated tokens, which our diagnosis finds unreliable in open models once objects look alike or are hidden. Our call-matched baseline (Table~\ref{tab:budget}) is self-consistency with one sample per agent VLM call.

\textbf{Object permanence.} Infants expect a hidden object to persist~\cite{baillargeon1985} and parse scenes into cohesive bodies that move continuously~\cite{spelke1990}. In vision, CATER~\cite{cater} and CLEVRER~\cite{clevrer} test these abilities on synthetic videos with containment and collisions. Learned models localise contained and carried objects~\cite{shamsian2020}, keep tracks through full occlusion~\cite{tokmakov2021}, learn intuitive physics from object-centric codes~\cite{plato}, and segment objects inside containers~\cite{tcow}. Xie and Alanwar~\cite{trackingnotperm} find that video world models keep visible objects but drop hidden ones. We use a frozen tracker and VLM to maintain explicit identity associations across occlusion and containment.

\textbf{Segmentation and tracking.} Promptable segmentation~\cite{sam} extends to video with a streaming memory in SAM2~\cite{sam2} and to concept prompts in SAM3~\cite{sam3}. Memory-based video object segmentation~\cite{xmem,cutie}, point tracking~\cite{cotracker} and tracking-by-detection~\cite{bytetrack} keep identities through motion and short occlusion, and open-vocabulary detectors~\cite{groundingdino} supply the initial boxes. None of them knows which object a question refers to. Our agent obtains the boxes from the VLM and the identities from SAM2.1.

\textbf{Visual prompting with tracks.} Marks drawn on an image steer a model to a region: a red circle for CLIP~\cite{redcircle}, numbered regions in Set-of-Mark~\cite{som}, and free-form prompts learned by ViP-LLaVA~\cite{vipllava}. Coarse Correspondences~\cite{coarsecorr} draws tracker IDs on video frames and lets the model answer the whole question. Elysium~\cite{elysium}, VideoRefer~\cite{videorefer} and Sa2VA~\cite{sa2va} instead train Video-LMs to track, refer to or segment objects. We draw the same kind of numbered tracks, but the model only answers a static question about one moment, and the answer comes from the tracks. Section~\ref{sec:mech} shows that the overlay alone recovers little of the gain at 8B and part of it at 32B.

\textbf{Tool-using agents.} VisProg~\cite{visprog} and ViperGPT~\cite{vipergpt} compose vision modules through generated programs, and ReAct~\cite{react} interleaves reasoning with tool calls. For video, agents retrieve frames iteratively~\cite{videoagentwang}, summarise clip captions~\cite{llovi,videoinsta}, plan over a symbolic memory~\cite{doraemongpt}, or store object-centric tracking states that a large language model (LLM) queries~\cite{videoagent}. STRAND~\cite{strand} links per-chunk object states by attribute and position similarity, making identical objects difficult to separate. SpatialClaw~\cite{spatialclaw} writes code that calls SAM3 and Depth Anything 3~\cite{da3}. AgentRVOS~\cite{agentrvos} reasons over SAM3 tracks to segment a referred, visible object, and Dual-Agent~\cite{whentrackersfail} and ACTrack~\cite{actrack} use a VLM to verify or re-prompt a tracker. These agents, like ours, keep object state outside the model, and they leave the final judgement of which object is which to the model. Our design differs in when the model is asked about identity and in what decides the answer once identity is fixed.

\section{Problem Formulation}\label{sec:problem}
\textbf{Setting.} A video $V=(I_1,\dots,I_T)$ shows a set of physical objects $\mathcal{O}$. A question $q$ with answer space $\mathcal{A}$ concerns a \emph{target} $o^\star\in\mathcal{O}$. It may ask where the target is at the end, which of several objects it is, or when and how often an object is visible. The objects that hide, cover or resemble the target are its \emph{carriers} $\mathcal{C}=\{c_1,\dots,c_K\}\subset\mathcal{O}$. A Video-LM $f$ answers directly from a sparse frame sample $V_S=(I_t)_{t\in S}$:
\begin{equation}\label{eq:direct}
\hat a_{\mathrm{direct}} = f(V_S,\,q),\qquad \hat a_{\mathrm{direct}}\in\mathcal{A}.
\end{equation}
Our agent uses the same frozen model $f$, which we call the \emph{VLM} when it is asked about single frames, short clips or text.

\textbf{Evidence moments.} An \emph{evidence moment} $t$ has a local observation window $W_t=V_{[t-\delta,\,t+\delta]}$ in which the target or its containing carrier can be identified:
\begin{equation}\label{eq:evidence}
\mathcal{E}=\{\,t : o^\star \text{ is identifiable among } \mathcal{C} \text{ from } W_t\,\}.
\end{equation}
The target is identified either \emph{by its look}, as the queen while it is face up, or as \emph{hidden in a carrier}, as the ball at the moment it is covered. In shell and card games, this evidence often disappears before the final frame. A direct answer (Eq.~\ref{eq:direct}) must then carry identity from an evidence moment $t_e\in\mathcal{E}$ to $T$ through a sparse sample of frames in which identical carriers cross. When $q$ concerns occlusion over time, it asks when an object disappears or returns, and these events may fall between the sampled frames.

\textbf{Principle.} A tracker follows each carrier $c_k$ with a mask track $\tau_k=(m_{k,1},\dots,m_{k,T})$. We fix identity once, at an evidence moment, and keep it outside the model afterwards. The answer factors into a \emph{bind} and a \emph{read-out}:
\begin{align}
k^\star &= f\big(W_{\mathrm{b}}\oplus\{\tau_k\}_{k=1}^{K},\;q_{\mathrm{b}}\big),\label{eq:bind}\\
\hat a &= f\big(q,\;\mathcal{F}(k^\star,\{\tau_k\}_{k=1}^{K})\big).\label{eq:readout}
\end{align}
In Eq.~\ref{eq:bind}, the bind clip $W_{\mathrm{b}}$ is $W_{t_e}$ when a moment is located, or the sampled whole video otherwise. The operator $\oplus$ draws each track with its number $k$ on this clip, and $q_{\mathrm{b}}$ asks about the evidence moment in it. The VLM returns the track index $k^\star$ of the target. In Eq.~\ref{eq:readout}, $\mathcal{F}$ states the tracked state in text, for example the left-to-right order of the tracks in the last frame. The VLM answers $q$ from these facts, without the video. Identity thus crosses the interval $(t_e,T]$ only through the track $\tau_{k^\star}$, and the VLM never has to recognise the target after the evidence is gone. Questions without a target to bind, such as how often an object is hidden, receive a timeline:
\begin{equation}\label{eq:timeline}
\hat a = f\big(V_S,\;q\oplus\mathcal{L}(\{\tau_k\})\big),
\end{equation}
where $\mathcal{L}$ lists per-object visibility segments and events computed from mask areas; $\oplus$ appends them to the prompt.

\section{Method}\label{sec:method}
Track, Then Bind organises inference into five steps (Fig.~\ref{fig:pipeline}; Algorithm~\ref{alg:ttb} in Appendix~\ref{app:impl}). A plan decides which steps run, and all thresholds are the same for every benchmark.

\textbf{Step 1: Plan.} The VLM watches $V$ at 2\,fps and fills a fixed schema. It lists the entities in $q$ with their roles: the target, its carriers, and other \emph{subjects}. It states how the target is recognised, by its look or as hidden in a carrier, and whose left and right the question uses. A separate text-only question asks whether $q$ is about the target's final location. If the target is recognised by its look and no carrier is named, all objects of the target's category become its carriers. The plan selects these operations from the question using the same schema across benchmarks.

\textbf{Step 2: Perceive.} The VLM boxes all entities in one call per frame, on 8 frames spaced evenly over the video. Entities found on none of them receive a second round of 7 frames in between. SAM2.1~\cite{sam2} is prompted on the earliest frame that shows the most common number of instances, or on the largest box for a single object. It propagates each instance forward and backward at 6\,fps. Carrier tracks that overlap, move more than a quarter of their width per frame, or end early trigger a second pass. This pass tracks the carriers and the target on every $k$-th native frame. Here $k$ is the native frame count over 300, rounded, and at least 1. The 12\,s VET-Bench videos are thus tracked at their native 30\,fps. Tracking all candidates jointly separates identical objects by their trajectories, and the dense pass is spent only where they cross. All questions on one video share its boxes and tracks.

\textbf{Step 3: Find evidence moments.} The bind question of step~4 always names the moment it is about: the target in view, or the target right after it is hidden. Step~3 tries to locate that moment $t_e$ in the tracks, so that the VLM sees a short clip (Eq.~\ref{eq:bind}). For a target recognised by its look, the moments are the frames on which the VLM found it in step~2. A mask track alone cannot tell a face-up card from its back. For a hidden target, the moment is its last view before its own track vanishes inside the frame. Small hidden objects are often lost by the tracker or never vanish cleanly. In that case no moment is located, and the clip covers the whole video. The VLM then finds the moment named by the question itself. Section~\ref{sec:ablation} reports how often each case occurs (Table~\ref{tab:branches}) and when a located moment matters (Table~\ref{tab:mech}).

\textbf{Step 4: Bind.} Each carrier track is drawn as a numbered box on a 4\,fps clip, with at most 96 frames. The clip spans 3\,s around a located moment ($\delta=1.5$\,s), or the whole video. The VLM answers one static question $q_{\mathrm{b}}$ and returns $k^\star$. For a target recognised by its look it names the numbered carrier that is the target. For a hidden target it names the carrier the target is under right after it is hidden. When several moments exist, the last one with a valid number decides. The question concerns a single moment, not the history after it, so the VLM never has to follow the target through the shuffle.

\textbf{Step 5: Read out.} A bound target is answered from facts. The agent states them in plain words, for example ``the hidden object was hidden under mug \#2 \ldots\ so the hidden object is under the first mug from the camera's left''. If $q$ uses a person's point of view, one more VLM call on four frames asks whether that person is opposite the camera. Its answer is added as a viewpoint fact. The facts then also give the person's left-to-right position. These facts are $\mathcal{F}$ in Eq.~\ref{eq:readout}. The VLM answers $q$ from them alone, without the video, making the tracked facts the evidence for read-out. Questions without a bound target use the timeline $\mathcal{L}$ of Eq.~\ref{eq:timeline}. Per-subject visibility segments, an ordered list of events and the coarse direction of motion are appended to the benchmark's prompt. The VLM then answers with the benchmark's own frames, template and parser. An answer that cannot be parsed falls back to the direct answer.

\section{Experiments}\label{sec:experiments}

\subsection{Experimental Setup}\label{sec:setup}

\textbf{Benchmarks.} We evaluate on three benchmarks about object identity and temporal object consistency. VET-Bench~\cite{vetbench} has 50 synthetic shell-game videos (\emph{cup}) and 50 card-monte videos (\emph{card}). The Perception Test~\cite{perceptiontest} cup-game subset (\emph{PT}) has 107 real test videos. The VET-Bench authors filtered it to remove single-frame shortcuts. Following them, we also report its 65 videos with three cups and at least one swap (\emph{PT-65}). TOC-Bench~\cite{tocbench} asks about object consistency in real videos along ten dimensions, such as event counts, durations and reappearance. Almost none of its questions needs binding (Table~\ref{tab:branches}), so it tests visibility timelines. We score each benchmark with its own answer parser.

\textbf{Development and test data.} Design choices and thresholds are selected on development data. These are 195 cup-game videos from the Perception Test validation split, 56 card-monte videos rendered with the VET-Bench generator, and 478 TOC-Bench questions. We split the 2,323 TOC-Bench questions by video into these 478 development questions and 1,845 test questions. The configuration is frozen and then run once on the test sets. Each ablation on the test sets changes one step of the frozen configuration.

\textbf{Baselines.} Open Video-LMs answer directly with each benchmark's prompt, frames and parser. We run them in vLLM~\cite{vllm}. Appendix~\ref{app:results} also gives Qwen3-VL answering in Hugging Face Transformers, the engine of our agent. The reasoning-enhanced baselines are the Qwen3-VL thinking variants and two trained models. Molmo2-SGCoT~\cite{vetbench} is fine-tuned to write object trajectories before it answers. Open-o3-Video-v2~\cite{openo3v2} is trained with reinforcement learning to crop and inspect frames. We run the official code of three training-free agents with Qwen3-VL-32B-Instruct, the backbone of our 32B agent. SpatialClaw~\cite{spatialclaw} writes Python code calling SAM3~\cite{sam3} and Depth Anything 3~\cite{da3}. VideoAgent~\cite{videoagent} builds an object memory from a detector, a tracker and re-identification, and queries it with an LLM. VideoINSTA~\cite{videoinsta} reasons over captions and object lists of event clips. Section~\ref{sec:budget} adds call-matched and increased-input baselines. Section~\ref{sec:families} runs the agent with InternVL3.5-8B~\cite{internvl35}, Gemma 4 31B~\cite{gemma4}, GPT-5.5 and GPT-6.1~\cite{gpt61}. Appendices~\ref{app:baselines} and~\ref{app:families} list all settings.

\textbf{Implementation details.} The VLM is Qwen3-VL-8B- or 32B-Instruct~\cite{qwen3vl} with greedy decoding, and the tracker is SAM2.1-hiera-large~\cite{sam2}. A timeline marks an object as fully visible when its mask covers at least 0.6 of its 90th-percentile area, and as partly hidden down to 0.12 of it. Segments shorter than 0.5\,s are merged. A disappearance counts as leaving the frame if the last box lies within 2\% of the border. All prompts are listed in Appendix~\ref{app:prompts}. Paired comparisons report item-level bootstrap 95\% confidence intervals (CI) and exact McNemar tests.

\subsection{Main Results}\label{sec:main}

\textbf{Comparison with existing methods.} Table~\ref{tab:main} compares Track, Then Bind with direct answers, reasoning-enhanced models and tool agents. On VET-Bench, the only open method substantially above chance on both tasks is SGCoT, which is fine-tuned on synthetic trajectories and reaches 92.0\% and 90.0\%. Our agent solves both tasks with both backbones and without training. Among closed models, GPT-5.5 answers 94.0\% and 96.0\%, and GPT-6.1 solves both. On the 107 real cup games, SGCoT falls back to 32.7\%. The agent reaches 51.4\% at 8B and 52.3\% at 32B. That is 17.8 (CI [3.7, 30.8]) and 16.8 points (CI [3.7, 29.9]) above the same models answering directly. The shuffled three-cup subset separates the methods most clearly. Every open baseline stays within ten points of chance on it, and the 32B agent reaches 55.4\%. GPT-5.5 and GPT-6.1 reach 72.3\% and 90.8\%. On TOC-Bench the agent reaches 43.0\% at 8B and 48.6\% at 32B, 17.9 (CI [15.4, 20.4]) and 11.5 points (CI [9.0, 14.0]) above the direct answers. The 32B agent is also 11.7 points above the 32B thinking variant and 14.7 points above Open-o3-Video-v2. GPT-5.5 answers 47.6\% and GPT-6.1 52.5\% of these questions.

\textbf{Comparison with tool-using agents.} SpatialClaw, VideoAgent and VideoINSTA run with our 32B backbone. All three stay within ten points of chance on the cup games and below the direct answer on TOC-Bench. Our 32B agent is 10.3--72.0 points higher on every set (Appendix~\ref{app:results}). Only the 10.3-point gap to VideoINSTA on real cup games is not significant. The tools alone do not carry identity. SpatialClaw propagates masks through only a quarter of the VET-Bench videos, VideoAgent's detector misses playing cards, and VideoINSTA keeps no object track.

\begin{table}[t]
\centering
\scriptsize
\setlength{\tabcolsep}{2.4pt}
\caption{\textbf{One configuration across VLM families} (test sets, \%). D: the VLM answering directly (GPT-5.5 and GPT-6.1 reason at medium effort); T: Track, Then Bind with the same VLM and the frozen configuration. $^{\ddagger}$First option letter, as in Table~\ref{tab:main}; the VET-Bench scorer gives 9.3 (D) and 34.6 (T).}
\label{tab:families}
\begin{tabular}{lcccccccc}
\toprule
 & \multicolumn{2}{c}{VET cup} & \multicolumn{2}{c}{VET card} & \multicolumn{2}{c}{PT} & \multicolumn{2}{c}{TOC}\\
\cmidrule(lr){2-3}\cmidrule(lr){4-5}\cmidrule(lr){6-7}\cmidrule(lr){8-9}
VLM & D & T & D & T & D & T & D & T\\
\midrule
Qwen3-VL-8B & 28.0 & 100 & 28.0 & 100 & 33.6 & 51.4 & 25.1 & 43.0\\
Qwen3-VL-32B & 32.0 & 100 & 38.0 & 100 & 35.5 & 52.3 & 37.1 & 48.6\\
InternVL3.5-8B & 32.0 & 48.0 & 32.0 & 32.0 & 43.0$^{\ddagger}$ & 35.5$^{\ddagger}$ & 31.3 & 41.7\\
Gemma 4 31B & 36.0 & 100 & 28.0 & 100 & 36.4 & 48.6 & 40.0 & 48.8\\
GPT-5.5 & 94.0 & 100 & 96.0 & 100 & 68.2 & 61.7 & 47.6 & 49.2\\
GPT-6.1 & 100 & 100 & 100 & 100 & 92.5 & 71.0 & 52.5 & 57.7\\
\bottomrule
\end{tabular}
\end{table}

\textbf{Across VLM families.}\label{sec:families} Table~\ref{tab:families} runs the frozen configuration with four VLM families. Each backend adapts only its model's input format and box convention (Appendix~\ref{app:families}). With Qwen3-VL, Gemma 4, GPT-5.5 and GPT-6.1, the agent solves both VET-Bench tasks. The plan of InternVL3.5 names no carrier on card monte, so nothing is bound. Given the plans of Qwen3-VL-8B, InternVL3.5 answers 98\% of card monte. On TOC-Bench the agent answers from track-derived timelines and binds almost no question (Table~\ref{tab:branches}). These timelines narrow the gap between VLMs: direct answers span 25.1--52.5\%, and the agent spans 41.7--57.7\%. The gain is largest for the weakest direct answer, 17.9 points with Qwen3-VL-8B, and smallest with GPT-5.5, 1.6 points (CI [$-0.7$, 4.0]); GPT-6.1 gains 5.2 points (52.5\% to 57.7\%). On real cup games a stronger VLM still helps the agent: 48.6--52.3\% with Qwen3-VL and Gemma 4, 61.7\% with GPT-5.5 and 71.0\% with GPT-6.1. InternVL3.5 gains nothing there (43.0\% direct, 35.5\% agent). For the strongest model, the external pipeline becomes the bottleneck. GPT-6.1 answers 92.5\% directly, 21.5 points above its agent (CI [13.1, 29.9]). Most of the agent's 31 errors with GPT-6.1 come from the bound track (Appendix~\ref{app:families}).

\begin{table}[t]
\centering
\scriptsize
\setlength{\tabcolsep}{1.6pt}
\caption{\textbf{Call-matched and increased-input baselines} (test sets, \%). \emph{Self-consistency}~\cite{selfconsistency}: $k$ answers sampled at the model's recommended temperature and a majority vote, with $k$ the agent's mean VLM calls on that benchmark (at most 16). \emph{More input tokens}: one answer from 16\,fps and larger frames. \emph{SpatialClaw}: as in Table~\ref{tab:main}. Calls: VLM calls per question on cup/card/PT/TOC. Tokens: input tokens per PT question. Transformers runs: Appendix~\ref{app:results}.}
\label{tab:budget}
\input{figs/tab_budget.tex}
\end{table}

\textbf{Comparison with larger inference budgets.}\label{sec:budget} The agent issues 11--19 VLM calls and 7--27k input tokens per question (Appendix~\ref{app:impl}), while a direct answer issues one call. Table~\ref{tab:budget} increases the direct answer's call or input budget. Self-consistency samples as many answers as the agent makes calls and takes the majority vote. The second baseline increases input tokens in a single answer, with 16\,fps and larger frames. On VET-Bench the voted answers stay at 28--32\%. After the shuffle no frame shows where the target is, so more samples of the same frames add no evidence. On the real cup games, the 8B vote over 13 samples reaches 46.7\%, against 33.6\% for a single answer. The agent is 4.7 points higher (CI [$-9.3$, 18.7]) and reads one sixth of the input tokens (25k against 157k). At 32B the vote reaches 37.4\%, and the agent is 15.0 points higher (CI [2.8, 27.1]). On TOC-Bench the vote changes the direct answer by less than one point at both sizes. A single answer with more input tokens reads 34k tokens per real cup game, and its accuracy falls below that of the plain direct answer everywhere except 8B card monte. SpatialClaw spends the largest budget, about 19 times the agent's input tokens per real cup game.

\subsection{Ablation Study}\label{sec:ablation}

\begin{table}[t]
\centering
\scriptsize
\setlength{\tabcolsep}{3pt}
\caption{\textbf{Ablation of the three steps} (test sets, accuracy \%). Each variant keeps the plan, boxes and tracks and changes one step. \emph{Tracks drawn, question asked directly}: the numbered whole-video clip with the benchmark question, as in visual prompting with track IDs. \emph{Bind question on the whole video}: no located moment. \emph{Bind question about the end}: the same clip, asked where the target is at the end. \emph{Facts read out with the video}: the same facts are appended to the benchmark prompt, which is answered with the video.}
\label{tab:mech}
\input{figs/tab_mech.tex}
\end{table}

The principle has three parts. A static question binds the target to a track, it is asked about the evidence moment, and the answer is read from the tracks. Table~\ref{tab:mech} removes or replaces each part on the test sets, with the plan, the boxes and the tracks fixed, and also removes the located clip.

\textbf{Binding.}\label{sec:mech} Without the bind question, the numbered tracks are drawn on the whole video and the benchmark question is asked directly, as in Coarse Correspondences~\cite{coarsecorr}. The 8B model then answers 44.0\%, 40.0\% and 32.7\% of the cup, card and real cup games, close to its direct answer. The 32B model reads the numbers better. It solves card monte and reaches 78.0\% on the cup game. On the real cup games it stays at 40.2\%, 12.1 points below the agent (CI [0.0, 24.3]).

\begin{figure*}[t]
\centering
\resizebox{0.95\textwidth}{!}{\input{figs/qual.tex}}
\caption{\textbf{Qualitative results} (test videos, Qwen3-VL-8B unless noted). (a)~A success that no method without binding reproduces. (b, c)~Two failures on real cup games: a wrong viewpoint fact with correct tracks, and a cup whose track is lost when both hands cross the cups. Numbered boxes are the agent's tracks as the VLM sees them.}
\label{fig:qual}
\end{figure*}

\textbf{Evidence moment.} With the same tracks and the same whole-video clip, asking where the target is at the end, instead of right after it is hidden, lowers accuracy on every test set at both scales. At 8B, VET-Bench falls to 28.0\% and 36.0\%, near chance, and the real cup games fall by 15.0 points (CI [$-25.2$, $-4.7$]).

\textbf{Located clip.} Asking the bind question on the numbered clip of the whole video keeps both VET-Bench tasks at 100\% at 8B. On the real cup games it gives 53.3\%, 1.9 points above the agent (CI [$-1.9$, 5.6]). At 32B, card monte stays at 100\% and the real cup games reach 54.2\%, but the cup game drops to 72.0\%. A located moment shortens the bind clip from about 47 to 12 frames on the cup game, and the 32B model needs the short clip there.

\textbf{Text-only read-out.} Appending the same facts to the benchmark prompt and answering with the video lowers accuracy on all three sets. At 8B the cup game drops from 100\% to 38.0\% and card monte to 72.0\%. At 32B they drop to 76.0\% and 96.0\%. Reintroducing the shuffled video can reduce accuracy even when the tracked facts are unchanged.

\begin{table}[t]
\centering
\scriptsize
\setlength{\tabcolsep}{2.6pt}
\caption{\textbf{Bind clips selected by step~3} (test sets). \emph{Located}: the VLM sees a 3\,s clip around the evidence moment. \emph{Whole video}: no moment is found, and the bind question is asked on the numbered clip of the whole video. \emph{No binding}: the plan finds no target to tell apart, or fewer than two carrier tracks, and the question is answered with a timeline. Each cell gives the share of questions and, in parentheses, the accuracy on them. Frames: mean frames per bind call; --: no bind call is made.}
\label{tab:branches}
\begin{tabular}{llccc}
\toprule
Test set & Bind clip & 8B & 32B & Frames\\
\midrule
VET cup & located (hidden) & 100\% (100) & 100\% (100) & 12\\
VET card & located (in view) & 100\% (100) & 100\% (100) & 6\\
PT & located (hidden) & 20.6\% (40.9) & 11.2\% (33.3) & 12\\
PT & whole video & 74.8\% (55.0) & 81.3\% (57.5) & 60\\
PT & no binding & 4.7\% (40.0) & 7.5\% (25.0) & --\\
TOC & no binding & 100\% (43.0) & 99.9\% (48.7) & --\\
\bottomrule
\end{tabular}
\end{table}

\textbf{Bind clips.}\label{sec:branches} Table~\ref{tab:branches} shows which bind clip step~3 selects. On VET-Bench every question is bound on a located clip: 12 frames around the hiding of the ball, or 6 frames while the queen is face up. In real cup games the hidden object is small. Its own track yields a moment in only 21\% (8B) and 11\% (32B) of the videos. The other videos are bound on the whole numbered clip, about 60 frames over 15\,s. They are answered more often than the located ones (55.0\% against 40.9\% at 8B). On real videos the gain thus comes from the static question and the tracks; a short window adds little.

\subsection{Qualitative Analysis}\label{sec:qual}

\textbf{Success case.} In the card-monte video of Fig.~\ref{fig:qual}(a), the queen is face up for one second. Every method that leaves the final judgement to the VLM answers that the queen stayed in the middle. SpatialClaw (32B) loses the queen's mask once the card turns over and then guesses the middle (Fig.~\ref{fig:gallery_sc}). Our agent asks which card is the queen only while it is face up. The tracks then carry the queen to the right.

\textbf{Failure cases.} We annotated whether the person in each real cup game faces the camera (Appendix~\ref{app:viewpoint}) and attribute each wrong answer under this viewpoint. At 8B, 7 of the 52 errors come from a wrong viewpoint fact (Fig.~\ref{fig:qual}b) and 38 from binding or tracking. We inspected these 38 by hand: the bound track never moves to another cup. The errors arise in the bind answer, or when the track of another cup is lost, which shifts the count from the left (Fig.~\ref{fig:qual}c). At 32B, 14 errors come from the viewpoint fact and 22 from binding or tracking. Most of the remaining errors at both scales are questions that are never bound.

\section{Limitations}
The agent inherits the errors of its tracker and of its VLM's plan. Identical cups crossed by hand can still confuse the tracker, very small hidden objects are rarely tracked, and a plan that names no carrier binds nothing. Real-video evidence for binding comes from the Perception Test cup games, the only real benchmark we know of in which a hidden object moves among look-alike carriers. For a model that already solves these games directly, such as GPT-6.1, the external pipeline adds errors. The agent also makes several VLM calls per question (Table~\ref{tab:cost}).

\section{Conclusion}
Open Video-LMs lose object identity when the identifying evidence disappears or the objects look alike, while GPT-6.1 solves the same games directly. Track, Then Bind asks the VLM one question about the moment the evidence is visible, then carries the identity with tracks and reads the answer from the tracked state. Without training, it solves both VET-Bench tasks with three VLM families and raises Qwen3-VL-32B from 35.5\% to 52.3\% on real cup games. Its track-derived timelines add 8.8--17.9 points on TOC-Bench to open models. The benefit of external tracking depends on the model and the task: GPT-6.1 already solves these games directly and loses accuracy through the pipeline, yet still gains from the timelines on TOC-Bench.

\section*{LLM Usage Statement}
Vision-language models are components of the method and the baselines studied in this paper. Beyond these, LLMs were used as coding assistants and to polish the wording of the manuscript. The authors verified all code and results and take full responsibility for the content.

\clearpage
\onecolumn
\appendix
\section{Algorithm and Implementation Details}\label{app:impl}
Algorithm~\ref{alg:ttb} gives the control flow of Track, Then Bind. The same program and the values of Table~\ref{tab:hyper} run on every benchmark. The plan decides which branch a question takes; no branch is selected by the benchmark name.

\begin{algorithm}[h!]
\caption{\textbf{Track, Then Bind.} $f$: the frozen VLM; $\mathcal{T}$: the SAM2.1 video tracker. Read-outs use the benchmark's own prompt template and answer parser.}
\label{alg:ttb}
\small
\begin{enumerate}[label=\arabic*:, leftmargin=2.2em, itemsep=1.5pt, topsep=3pt]
\item \textbf{Input:} video $V$, question $q$. \textbf{Output:} answer $\hat a$.
\item $P \gets f(V \text{ at 2\,fps}, \textsc{Plan}(q))$;\enspace $P.\textit{final} \gets f(\textsc{FinalLocation}(q))$ \hfill$\triangleright$ step 1: entities, roles, cue, viewpoint
\item \textbf{if} $P.\textit{cue}=\text{visible}$, $P.\textit{final}$, and no carrier is named \textbf{then} add the target's category as the carriers
\item \textbf{if} $P$ has no entities \textbf{then return} $\textsc{Parse}(f(V_S, q))$ \hfill$\triangleright$ direct answer
\item \textbf{for each} entity $e$ \textbf{do} \hfill$\triangleright$ step 2
\item \hspace*{1.2em}boxes $\gets f(\textsc{Ground})$ on 8 uniform frames, then on the 7 frames in between if $e$ was found on none
\item \hspace*{1.2em}$\tau_e \gets \mathcal{T}(V \text{ at 6\,fps})$ from the earliest frame with the modal box count (one box: the largest)
\item \textbf{if} carrier tracks overlap, jump, or are not all tracked after 90\% of the video \textbf{then}
\item \hspace*{1.2em}re-track carriers and target on every $k$-th frame, $k=\max(1,\operatorname{round}(N_{\text{native}}/300))$ \hfill$\triangleright$ dense pass
\item $\hat a \gets \varnothing$
\item \textbf{if} $P.\textit{final}$, $P.\textit{cue} \in \{\text{visible}, \text{contained}\}$, a target track and $\geq 2$ carrier tracks exist \textbf{then}
\item \hspace*{1.2em}$\mathcal{M} \gets$ evidence moments \hfill$\triangleright$ step 3
\item \hspace*{2.4em}visible: frames on which the target's box was found, merged within 3\,s
\item \hspace*{2.4em}contained: the last frame before the target's own track vanishes inside the frame
\item \hspace*{1.2em}\textbf{for each} window $W \in \mathcal{M}$, or the whole video if $\mathcal{M}=\varnothing$ \textbf{do} \hfill$\triangleright$ step 4
\item \hspace*{2.4em}$k \gets f(W \oplus \{\tau_c\} \text{ at 4\,fps, } \leq 96 \text{ frames},\ q_{\mathrm{b}}(P.\textit{cue}))$;\enspace keep the last valid $k$ as $k^\star$
\item \hspace*{1.2em}\textbf{if} $k^\star$ exists \textbf{then} \hfill$\triangleright$ step 5a
\item \hspace*{2.4em}$\mathcal{F} \gets$ the bound track and the left-to-right track order in the last frame where all carriers are tracked
\item \hspace*{2.4em}\textbf{if} $P.\textit{viewpoint}=\text{person}$ \textbf{then} $\mathcal{F} \gets \mathcal{F} \cup$ viewpoint fact from $f(4 \text{ frames}, \textsc{Opposite})$
\item \hspace*{2.4em}$\hat a \gets \textsc{Parse}(f(q, \mathcal{F}))$ \hfill$\triangleright$ text only
\item \textbf{if} $k^\star$ does not exist \textbf{then} $\hat a \gets \textsc{Parse}(f(V_S,\ q \oplus \mathcal{L}(\{\tau_e\})))$ \hfill$\triangleright$ step 5b: timeline
\item \textbf{if} $\hat a=\varnothing$ \textbf{then} $\hat a \gets \textsc{Parse}(f(V_S, q))$ \hfill$\triangleright$ unparsable read-out
\item \textbf{return} $\hat a$
\end{enumerate}
\end{algorithm}

\begin{table}[h!]
\centering
\scriptsize
\caption{\textbf{Settings of the single configuration} (all benchmarks). $p_{90}$: the 90th percentile of a track's mask area over the video.}
\label{tab:hyper}
\begin{tabularx}{\textwidth}{@{}>{\raggedright\arraybackslash}p{2.0cm}>{\raggedright\arraybackslash}p{4.6cm}>{\raggedright\arraybackslash}X@{}}
\toprule
Step & Setting & Value\\
\midrule
1 Plan & video rate for the plan call & 2\,fps\\
 & entities per plan & at most 3\\
\midrule
2 Perceive & grounding frames & 8 uniform, plus 7 in between for entities found on none\\
 & tracking resolution & 960\,px on the long side\\
 & first tracking pass & 6\,fps, at most 900 frames\\
 & dense pass triggers & intersection over union (IoU) of two instances $>0.05$; centre shift $>0.25$ box widths per sampled frame; carriers not all tracked after 90\% of the video\\
 & dense pass rate & target rate $r=\min(\text{native},\ 300/\text{duration})$; the pass runs if $r$ exceeds $1.5\times$ the first-pass rate and reads every $k$-th native frame, $k=\max(1,\operatorname{round}(\text{native}/r))$, so 300 is a target, not a cap (366 frames on VET-Bench)\\
\midrule
3 Evidence, timeline & fully visible / partly hidden & mask area $\geq 0.6\,p_{90}$ / $\geq 0.12\,p_{90}$\\
 & shortest segment & 0.5\,s (shorter segments are merged into a neighbour)\\
 & left the frame & last box within 2\% of the image border\\
 & coarse motion & box-centre shift between the first and last third of the visible frames; ``stays'' below 0.1 of the frame\\
\midrule
4 Bind & clip & 4\,fps, 3\,s around a moment ($\delta=1.5$\,s) or the whole video, at most 96 frames\\
 & several moments & the last one with a valid number decides\\
\midrule
5 Read out & viewpoint question & 4 uniform frames\\
 & benchmark frames (timeline read-out) & PT and VET video at 8\,fps; TOC-Bench 32 uniform frames\\
 & unparsable answer & falls back to the direct answer\\
\midrule
All & decoding & greedy\\
\bottomrule
\end{tabularx}
\end{table}

\begin{table}[h!]
\centering
\scriptsize
\caption{\textbf{Cost per question on the test sets} (means). VLM calls by input: \emph{video}, the plan call and a read-out on the benchmark's own frames; \emph{frame}, grounding and the viewpoint question; \emph{clip}, the bind call; \emph{text}, the final-location question and the text-only read-out. Times are wall-clock seconds on A800 GPUs, one GPU per process at 8B and two at 32B, with other jobs sharing the node. The tracker time includes the dense pass.}
\label{tab:cost}
\input{figs/app_cost.tex}
\end{table}

Table~\ref{tab:cost} breaks down the cost of Table~\ref{tab:budget}. Grounding dominates the number of calls. On the VET cup game the ball is found on none of the first 8 frames, so a second round of 7 calls follows. Each call sees one frame, and the input stays at 7--27k tokens per question. Tracking takes 60--70\,s per cup or card video because the dense pass runs on 88--100\% of these videos. On TOC-Bench it runs on 1\% of the questions, and tracking takes 12--14\,s.

\section{Benchmarks, Splits and Evaluation}\label{app:data}
\begin{table}[h!]
\centering
\scriptsize
\caption{\textbf{Evaluation data.} Duration: mean video length in seconds. Dev: used to select the design and the thresholds. Test: run once with the frozen configuration.}
\label{tab:splits}
\begin{tabular}{lllrrrl}
\toprule
Benchmark & Split & Use & Videos & Questions & Duration & Source\\
\midrule
VET-Bench cup & test & test & 50 & 50 & 12.4 & VET-Bench~\cite{vetbench} (synthetic)\\
VET-Bench card & dev & dev & 56 & 56 & 12.1 & rendered with the VET-Bench generator\\
VET-Bench card & test & test & 50 & 50 & 12.1 & VET-Bench~\cite{vetbench} (synthetic)\\
Perception Test cup games & validation & dev & 195 & 195 & 23.7 & Perception Test~\cite{perceptiontest}\\
Perception Test cup games & PT & test & 107 & 107 & 15.6 & filtered by the VET-Bench authors\\
\quad three cups, $\geq 1$ swap & PT-65 & test & 65 & 65 & -- & subset of PT\\
TOC-Bench & dev & dev & 389 & 478 & 28.6 & TOC-Bench~\cite{tocbench}, 20\% of videos\\
TOC-Bench & test & test & 1,561 & 1,845 & 28.9 & TOC-Bench~\cite{tocbench}, the other 80\%\\
Perception Test, all areas & validation & test & 1,602 & 2,000 & 23.3 & stratified sample, Table~\ref{tab:general}\\
Video-MME short + medium & -- & test & 600 & 1,800 & 298.7 & Video-MME~\cite{videomme}, Table~\ref{tab:general}\\
\bottomrule
\end{tabular}
\end{table}

\textbf{Development sets.} The cup-game development set is the Perception Test validation question \emph{``Where is the hidden object at the end of the game from the person's point of view?''} on 195 of its 198 videos; three videos were unavailable. The card-monte development set consists of 56 videos rendered with the VET-Bench generator. We changed only the encoder, from H.264 to VP9 and back to H.264 at a constant rate factor (CRF) of 18, because headless Chromium has no H.264 encoder. Each video carries the test-set question about the Queen of Hearts, with the answer taken from the generator's metadata. VET-Bench has no cup-game development set, so the cup game is tuned only through the real cup games and the card videos.

\textbf{TOC-Bench split.} We use the released answer subset of TOC-Bench, 2,323 questions over 1,951 videos. One development video could not be downloaded; its single question counts as wrong, and the development set has 389 usable videos. Its videos are assigned at random (seed 20261017) to development (20\%) and test (80\%), so that the questions of one video stay in one split. The test split has 1,068 four-way multiple-choice questions, 447 two-statement questions, 241 numerical questions and 89 ordering questions. TOC-Bench scores each format with its own parser, and an unparsable answer counts as wrong.

\textbf{Disjointness.} No video appears in two of the following sets: the cup-game development and test sets, the 2,000 general Perception Test questions, and the TOC-Bench development and test splits. The rendered card videos are new renders and share no file with the VET-Bench test videos.

\textbf{Scoring and statistics.} Perception Test and VET-Bench answers are parsed with the VET-Bench scorer, which reads the boxed letter. Every comparison in the paper is paired over the same questions. The confidence interval is the 2.5th and 97.5th percentile of 10,000 bootstrap resamples of the questions (seed 0). The $p$-value is the exact two-sided McNemar test on the questions that exactly one of the two methods answers correctly.

\section{Baselines}\label{app:baselines}
\begin{table}[h!]
\centering
\scriptsize
\caption{\textbf{Baseline settings.} All baselines receive the benchmark's prompt and are scored with its parser. Self-consistency and the larger input appear in Table~\ref{tab:budget}.}
\label{tab:baseline_settings}
\begin{tabularx}{\textwidth}{@{}>{\raggedright\arraybackslash}p{3.0cm}>{\raggedright\arraybackslash}p{3.2cm}>{\raggedright\arraybackslash}p{4.0cm}>{\raggedright\arraybackslash}X@{}}
\toprule
Baseline & Engine & Input & Decoding\\
\midrule
Qwen3-VL, direct & vLLM 0.15.1~\cite{vllm} & PT and VET: video at 8\,fps, default resizing of qwen-vl-utils; TOC: 32 uniform frames & greedy; 2,048 new tokens (TOC: the benchmark's budget per format)\\
Qwen3-VL, direct & Transformers 4.57.6 & same & greedy\\
Self-consistency & vLLM; Transformers (Table~\ref{tab:budgethf}) & same & $k$ samples at temperature 0.7, top-$p$ 0.8, top-$k$ 20 (Qwen3-VL's recommended settings), majority vote; $k$ equals the agent's mean VLM calls, at most 16\\
More input tokens & vLLM; Transformers (Table~\ref{tab:budgethf}) & PT and VET: 16\,fps, at most 288,000 pixels per frame, no resizing; TOC: 96 frames & greedy\\
Qwen3-VL-Thinking & vLLM (VET, PT); TOC-Bench runner (TOC) & as the direct answer & greedy; 16,384 new tokens\\
Qwen3-VL-235B-A22B & vLLM, 8 GPUs & video at 8\,fps & greedy; 2,048 new tokens\\
Molmo2-8B, Molmo2-SGCoT & Transformers, the authors' code & at most 8\,fps & greedy; 2,048 new tokens; SGCoT is the authors' low-rank adaptation (LoRA) weights with their prompt\\
Open-o3-Video-v2-8B & the official multi-turn evaluator & 1\,fps, 96 frames, crop tool, at most 5 turns & the evaluator's settings; it cannot express ordering questions, which count as wrong\\
SpatialClaw & the official code, Qwen3-VL-32B-Instruct (bf16) in vLLM & all frames at the native rate to the code, 32 key frames to the model & SpatialClaw's configuration for this model: temperature 0.7, top-$p$ 0.8, top-$k$ 20; at most 30 steps\\
VideoAgent & the official code; Qwen3-VL-32B-Instruct in vLLM as the agent LLM, Video-LLaVA-7B (4-bit) as the visual question answering tool & 2\,s segments; tracking at 15\,fps & temperature 0; the official ReAct prompts\\
VideoINSTA & the official code; Qwen3-VL-32B-Instruct in vLLM as the LLM, CogAgent for captions and objects & 1\,fps; 4 event clips (CLIP ViT-L/14); UniVTG grounding & temperature 0; the official NExT-QA configuration\\
\bottomrule
\end{tabularx}
\end{table}

\textbf{SpatialClaw.} We run the official code with its prompts, tools and model configuration. The SAM3 and Depth Anything 3 tools are checked out at the commits the code pins. An adapter passes each benchmark prompt verbatim as the question and scores the returned answer with the benchmark's parser. Some sessions met a tool server that was temporarily unavailable. We reran every such question once, 12 on the real cup games and 28 on TOC-Bench, and score the rerun. Table~\ref{tab:spatialclaw} summarises the runs.

\begin{table}[h!]
\centering
\scriptsize
\caption{\textbf{SpatialClaw with Qwen3-VL-32B on the test sets.} Calls and input tokens are means per question. Mask propagation: share of questions whose code propagated a SAM3 mask through the video at least once without an error.}
\label{tab:spatialclaw}
\input{figs/app_spatialclaw.tex}
\end{table}

SpatialClaw propagates masks through the video in about a quarter of the VET-Bench questions, and the other sessions work on single frames. On the real cup games it propagates masks in 77\% of the questions and still stays below chance. Fig.~\ref{fig:gallery_sc} shows one session.

\begin{figure}[h!]
\centering
\input{figs/gallery_sc.tex}
\caption{\textbf{SpatialClaw on VET card monte} \texttt{card\_001} (Qwen3-VL-32B, logged test run). The images are the session's own \texttt{show()} outputs. The queen's mask is empty once the card turns over, and the model answers from the last frame. Track, Then Bind answers the same video correctly (Fig.~\ref{fig:qual}a).}
\label{fig:gallery_sc}
\end{figure}

\textbf{VideoAgent.} We run the official code (commit 42e1b70) with its prompts, tools and memory construction. The memory holds LaViLa captions of 2\,s segments, ViCLIP segment features, RT-DETR-L tracks from ByteTrack at 15\,fps, and CLIP and DINOv2 features for re-identification. Video-LLaVA-7B answers its visual questions. Four parts need a closed API or the network, and we replace them. Qwen3-VL-32B-Instruct at temperature 0 takes the place of GPT-4 as the agent LLM. The OpenAI text embedding is replaced by all-mpnet-base-v2, a sentence model that the official encoder already offers. LangChain's prompt download is replaced by the same template held locally, and an unused caption-metric import by a stub. The official multiple-choice prompt fixes five choices, so it states the actual number of choices. LangChain returns format errors in the ReAct output to the agent instead of stopping (handle\_parsing\_errors). TOC-Bench ordering questions go to the official free-form runner. A question takes 9.8 LLM calls and 9.0k LLM input tokens on average. Eleven of the 2,052 questions end with a malformed tool input and count as wrong.

\textbf{VideoINSTA.} We run the official code (publication branch) with the authors' NExT-QA configuration. It samples one frame per second and splits the video into four event clips by clustering CLIP ViT-L/14 features. CogAgent describes the actions and objects of each clip, and UniVTG locates the spans that match the question. The LLM summarises each clip, rates how well it answers the question, and merges the ratings into an answer. Three changes are needed. Qwen3-VL-32B-Instruct at temperature 0 takes the place of GPT-3.5 as the LLM. The authors release CogAgent captions and object lists only for their own benchmarks, so we compute them with their extraction configuration. The prompts state the actual number of choices instead of five, and ordering questions use the authors' free-form prompts. A question takes 14.8 LLM calls and 11.2k LLM input tokens on average, measured from the server logs. One answer names a choice that the question does not have and counts as wrong.

\section{Other VLM Families}\label{app:families}
\textbf{Open models on VET-Bench.} Table~\ref{tab:vetopen} adds the open models whose VET-Bench predictions the VET-Bench authors released, rescored with the same parser. With our own runs, the open models cover eight families. All of them stay between 22\% and 40\% on both tasks.

\begin{table}[h!]
\centering
\scriptsize
\caption{\textbf{Open Video-LMs on VET-Bench} (accuracy, \%). Top, \cited{gray}: cited, rescored with the VET-Bench parser from the predictions released by the VET-Bench authors~\cite{vetbench}. Bottom: our runs (Table~\ref{tab:main}).}
\label{tab:vetopen}
\begin{tabular}{lcc}
\toprule
Model & cup & card\\
\midrule
Qwen3.5-397B-A17B & \cited{38.0} & \cited{32.0}\\
Kimi-K2.5 & \cited{28.0} & \cited{32.0}\\
ERNIE-4.5-VL-28B-A3B-Thinking & \cited{28.0} & \cited{30.0}\\
Perception-LM-8B~\cite{perceptionlm} & \cited{40.0} & \cited{34.0}\\
GLM-4.6V-Flash~\cite{glm46v} & \cited{34.0} & \cited{28.0}\\
\midrule
Qwen3-VL-8B / 32B / 235B-A22B & 28.0 / 32.0 / 22.0 & 28.0 / 38.0 / 38.0\\
InternVL3.5-8B & 32.0 & 32.0\\
Gemma 4 31B & 36.0 & 28.0\\
Molmo2-8B & 30.0 & 40.0\\
\bottomrule
\end{tabular}
\end{table}

\textbf{Backends.} The agent, its prompts and its configuration are the same for every VLM. Each backend adapts the input format and the box convention of its model.
\begin{itemize}
\item \emph{InternVL3.5-8B} is served by vLLM. The frames of a clip or video are separate images of one tile each, each preceded by ``Frame $i$ ($t$\,s):'', at most 128 per request. TOC-Bench frames follow the benchmark's own InternVL runner.
\item \emph{Gemma 4 31B} runs in Transformers with thinking disabled. It returns boxes as $[y_1, x_1, y_2, x_2]$, which the backend reorders. A request with several frames shares 4,480 visual tokens, and each frame receives the largest supported budget that fits (70 to 1,120). At most 64 frames are sent.
\item \emph{GPT-5.5} (\texttt{gpt-5.5-2026-04-23}) and \emph{GPT-6.1} (\texttt{gpt-6.1-sol}) answer through the OpenAI Chat Completions API. Both are reasoning models and reason at the API's default effort, medium, in every call, direct or within the agent. A direct answer on a real cup game uses on average 2,740 reasoning tokens with GPT-5.5 and 125 with GPT-6.1, and no answer is cut off. Frames are images of at most $360{\times}640$ pixels, each preceded by its time. Boxes come back in pixels and are converted to the 0--1000 grid. TOC-Bench frames follow the benchmark's OpenAI runner. With GPT-6.1, one of the 107 agent runs on the real cup games ends in an API error after one rerun and counts as wrong.
\end{itemize}
On VET-Bench and the Perception Test, direct answers use the frame rate of the Qwen3-VL baselines, 8\,fps, within each backend's frame limit. GPT-5.5 answers 47.6\% of our 1,845 TOC-Bench test questions, close to the 47.1\% that the TOC-Bench authors report on all 2,323~\cite{tocbench}.

\textbf{GPT-6.1 on unseen videos.} The videos of VET-Bench and the Perception Test were public before GPT-6.1 was trained. Two tests argue against memorisation. On the 56 card-monte development videos, which we render with the VET-Bench generator and have not released, GPT-6.1 answers all 56. We also mirror the real cup games left to right and remap the answers. Four of the 107 videos have no mirrored answer among their options. On the other 103, GPT-6.1 answers 91.3\% of the mirrored videos and 92.2\% of the originals ($-1.0$, CI [$-4.9$, 1.9]). On the 56 videos whose answer changes under the mirror, only four answers keep the original option.

\textbf{Viewpoint annotation.}\label{app:viewpoint} The real cup-game answers count positions from the left of the person who moves the objects, so the read-out must know whether that person faces the camera. One author labelled this for each of the 107 test videos from eight uniformly sampled frames, without the question or any answer. The person faces the camera in 73 videos and sits on the camera's side in 19; the 15 side-on videos are left unlabelled. With the annotated viewpoint in place of the agent's viewpoint fact, the read-out answers 56.1\% (8B) and 57.0\% (32B) of the real cup games, against 51.4\% and 52.3\%.

\textbf{Errors on real cup games by VLM.} Table~\ref{tab:errfam} applies the error analysis of Sec.~\ref{sec:qual} to every VLM. A wrong answer is a \emph{bind/track} error if the gold option disagrees with the bound track under the annotated viewpoint, a \emph{viewpoint} error if the agent's viewpoint fact disagrees with the annotation, and a \emph{read-out} error otherwise. For the 15 unlabelled videos, the gold option is checked under both viewpoints. With every VLM except InternVL3.5, most errors come from the bound track; with InternVL3.5, most errors are questions that the plan never binds.

\begin{table}[h!]
\centering
\scriptsize
\caption{\textbf{Errors of the agent on the 107 real cup games} (counts). Errors are attributed with the annotated viewpoint (Appendix~\ref{app:viewpoint}). Unbound: the plan binds no target. Other: options that are not positions, or an API error. $^{\ddagger}$First option letter, as in Table~\ref{tab:main}.}
\label{tab:errfam}
\begin{tabular}{lcccccc}
\toprule
VLM & errors & bind/track & viewpoint & unbound & read-out & other\\
\midrule
Qwen3-VL-8B & 52 & 38 & 7 & 5 & 0 & 2\\
Qwen3-VL-32B & 51 & 22 & 14 & 12 & 1 & 2\\
InternVL3.5-8B$^{\ddagger}$ & 69 & 11 & 4 & 53 & 1 & 0\\
Gemma 4 31B & 55 & 38 & 2 & 14 & 0 & 1\\
GPT-5.5 & 41 & 34 & 4 & 1 & 0 & 2\\
GPT-6.1 & 31 & 26 & 3 & 0 & 0 & 2\\
\bottomrule
\end{tabular}
\end{table}

\textbf{InternVL3.5 with the plans of Qwen3-VL-8B.} On card monte, the plan of InternVL3.5 lists the queen but no face-down cards, so the cards are never tracked. With the plans of Qwen3-VL-8B and all other steps by InternVL3.5, card monte rises from 32.0\% to 98.0\%. The cup game rises from 48.0\% to 54.0\%. On 40 of its 50 videos the stated facts imply the correct answer, and InternVL3.5 contradicts them in 17 of these 40. On TOC-Bench, the parser rejects 176 direct and 74 agent answers of InternVL3.5. On the 1,668 questions where both parse, the agent is 9.1 points above the direct answer (CI [6.5, 11.6]).

\section{Additional Results}\label{app:results}
\begin{table}[h!]
\centering
\scriptsize
\setlength{\tabcolsep}{3pt}
\caption{\textbf{TOC-Bench test accuracy by dimension} (\%). D: Qwen3-VL answering directly (vLLM); T: Track, Then Bind with the same model.}
\label{tab:dims}
\input{figs/tab_dims.tex}
\end{table}

\textbf{TOC-Bench by dimension.} The agent improves the direct answer in nine of ten dimensions at 8B and in eight at 32B (Table~\ref{tab:dims}). The largest gains, 21--34 points, are in event existence, reappearance versus disappearance, and relative spatial change. A visibility timeline answers these questions almost directly. Conditional-state questions ask about one object at an event of another. On them the 32B agent drops from 49.8\% to 44.7\%.

\begin{table}[h!]
\centering
\scriptsize
\setlength{\tabcolsep}{3.2pt}
\caption{\textbf{One configuration versus per-benchmark configurations} (test sets): accuracy (\%) and VLM calls per question.}
\label{tab:onecfg}
\begin{tabular}{llcccccc}
\toprule
 & & \multicolumn{4}{c}{Accuracy} & \multicolumn{2}{c}{Calls}\\
\cmidrule(lr){3-6}\cmidrule(lr){7-8}
VLM & Configuration & cup & card & PT & TOC & PT & TOC\\
\midrule
8B & per benchmark & 100 & 100 & 44.9 & \textbf{43.4} & 20 & 20.2\\
8B & \cellcolor{ours}single & \cellcolor{ours}100 & \cellcolor{ours}100 & \cellcolor{ours}\textbf{51.4} & \cellcolor{ours}43.0 & \cellcolor{ours}\textbf{13.4} & \cellcolor{ours}\textbf{10.8}\\
\midrule
32B & per benchmark & 100 & 100 & 50.5 & 48.0 & 20 & 22.7\\
32B & \cellcolor{ours}single & \cellcolor{ours}100 & \cellcolor{ours}100 & \cellcolor{ours}\textbf{52.3} & \cellcolor{ours}\textbf{48.6} & \cellcolor{ours}\textbf{14.7} & \cellcolor{ours}\textbf{11.1}\\
\bottomrule
\end{tabular}
\end{table}

\textbf{One configuration for all benchmarks.} Table~\ref{tab:onecfg} compares the agent with three variants of itself, each tuned on one benchmark. They fix the bind question, the grounding frames, the tracking rate and the read-out per benchmark. For example, the card variant grounds 4 frames from the first 30\% of each video and always tracks at 30\,fps. On all four test sets the single configuration lies within the CI of the tuned variants. The single configuration issues 10.8 instead of 20.2 VLM calls per TOC-Bench question at 8B, and 13.4 instead of 20 per real cup game.

\begin{table}[h!]
\centering
\scriptsize
\setlength{\tabcolsep}{4pt}
\caption{\textbf{General video questions} (accuracy, \%, Qwen3-VL-8B). Top: 2,000 Perception Test validation questions, stratified by area and type of reasoning. Bottom: Video-MME, short and medium videos. The direct answer is run in the same engine and the same pass as the agent.}
\label{tab:general}
\begin{tabular}{lrccc}
\toprule
Area & $n$ & Direct & Track, Then Bind & $\Delta$\\
\midrule
physics & 689 & 60.1 & 64.0 & $+3.9$\\
semantics & 634 & 86.4 & 86.9 & $+0.5$\\
abstraction & 593 & 63.4 & 64.2 & $+0.8$\\
memory & 84 & 58.3 & 66.7 & $+8.3$\\
\midrule
all & 2,000 & 69.3 & \textbf{71.5} & $+2.1$\\
\midrule
Video-MME short & 900 & \textbf{78.7} & 78.1 & $-0.6$\\
Video-MME medium & 900 & \textbf{65.1} & 63.7 & $-1.4$\\
Video-MME all & 1,800 & \textbf{71.9} & 70.9 & $-1.0$\\
\bottomrule
\end{tabular}
\end{table}

\textbf{General video questions.} Table~\ref{tab:general} tests the agent on questions that are not about object identity. We sample 2,000 questions from the Perception Test validation split, stratified by area and type of reasoning, and exclude the development videos. The agent answers 71.5\% of them and the direct answer 69.3\% ($+2.1$, CI [0.5, 3.7], $p{=}0.011$). The gain is largest on physics and memory questions, and semantic questions are unchanged. On 1,800 short and medium Video-MME~\cite{videomme} questions, the agent answers 70.9\% and the direct answer 71.9\% ($-1.0$, CI [$-2.1$, 0.1], $p{=}0.09$). These questions rarely concern a single object. The plan binds none of them, and the appended timeline adds little information.

\textbf{All paired comparisons.} Table~\ref{tab:paired} lists the difference between our agent and every baseline that uses the same VLM, on the same questions. The agent is ahead in every row. Three intervals include zero, all on the real cup games: the 8B self-consistency vote in either engine and VideoINSTA.

\begin{table}[h!]
\centering
\scriptsize
\caption{\textbf{Ours minus each baseline with the same VLM} (test sets, percentage points). CI: paired bootstrap 95\% interval; $p$: exact McNemar test.}
\label{tab:paired}
\input{figs/app_paired.tex}
\end{table}

\textbf{Budget baselines in Transformers.} Table~\ref{tab:budget} runs the call-matched and increased-input baselines in vLLM. Table~\ref{tab:budgethf} runs them in Hugging Face Transformers, the engine of our agent, with the same prompts, frames and sampling settings. The two engines lead to the same conclusions. On VET-Bench every baseline stays at 16--34\%. On the real cup games the 8B vote reaches 43.0\%, and the agent is 8.4 points higher (CI [$-4.7$, 21.5]). At 32B the vote reaches 37.4\%, 15.0 points below the agent (CI [1.9, 29.0]). On TOC-Bench the vote changes the direct answer by at most one point, and the agent is 10.7--18.6 points above every baseline. Table~\ref{tab:paired} gives all paired intervals.

\begin{table}[h!]
\centering
\scriptsize
\caption{\textbf{Call-matched and increased-input baselines in Transformers} (test sets, \%). Settings as in Table~\ref{tab:budget}. Self-consistency votes over 16, 12 and 13 (8B) or 15 (32B) samples on cup, card and PT, and over 11 samples on TOC.}
\label{tab:budgethf}
\input{figs/app_budget_hf.tex}
\end{table}

\textbf{TOC-Bench by dimension at 32B.} Table~\ref{tab:dims32} adds the thinking variant, SpatialClaw, VideoAgent and VideoINSTA to the 32B columns of Table~\ref{tab:dims}. The thinking variant moves no dimension by more than 10.1 points, and its overall accuracy is within 0.2 points of the direct answer. SpatialClaw gains 14.2 points on reappearance questions and loses 12.4 points on event counts. VideoINSTA and VideoAgent are the two best methods on event counts (39.4\% and 35.3\% against 32.0\% for ours). VideoAgent is below the direct answer in seven of the ten dimensions, and VideoINSTA in six.

\begin{table}[h!]
\centering
\scriptsize
\caption{\textbf{TOC-Bench test accuracy by dimension, Qwen3-VL-32B} (\%). Direct: answering directly in vLLM. Ours: Track, Then Bind. Best per row in bold.}
\label{tab:dims32}
\input{figs/app_dims32.tex}
\end{table}

\textbf{Perception Test cup games by subset.} Table~\ref{tab:ptgroups} splits the 107 real cup games by the number of cups and swaps given in the VET-Bench metadata. By the number of cups, the gain over the direct answer comes from the three-cup games. By the number of swaps, it is largest with one to three swaps. With four or more swaps the 8B agent gains 2.8 points and the 32B agent 11.4 points. The four four-cup games are answered once at 8B and never at 32B.

\begin{table}[h!]
\centering
\scriptsize
\caption{\textbf{Perception Test cup games by subset} (accuracy, \%). D: direct answer in vLLM.}
\label{tab:ptgroups}
\input{figs/app_ptgroups.tex}
\end{table}

\section{Design Choices on Development Data}\label{app:design}
\begin{table}[h!]
\centering
\scriptsize
\setlength{\tabcolsep}{3pt}
\caption{\textbf{Design choices on development data} (accuracy, \%). Rows 1--3 hold the tracks and the numbered overlays fixed and change only the question to the VLM.}
\label{tab:ablation}
\begin{tabular}{llcc}
\toprule
Variant & Set & 8B & 32B\\
\midrule
Direct answer & PT & 33.3 & 40.0\\
Tracks, asked ``where at the end?'' & PT & 40.0 & 44.1\\
Tracks, asked ``under which, right after hidden?'' & PT & \textbf{48.7} & \textbf{52.3}\\
\midrule
Card tracking at 10\,fps / 30\,fps & card & 37.5 / \textbf{100} & --\\
Viewpoint: ``faces the camera'' / ``opposite the camera'' & PT & 48.2 / \textbf{57.9} & --\\
Tracks, benchmark question, no binding & PT / card & 40.0 / 62.5 & 42.1 / \textbf{100}\\
Timeline: visibility only / + events and motion & TOC & 37.4 / \textbf{42.1} & --\\
Grounding frames: 16 / 8 & TOC & 40.2 / \textbf{42.7} & --\\
\bottomrule
\end{tabular}
\end{table}

Table~\ref{tab:ablation} lists the comparisons made while the design was chosen. Their $p$-values are exploratory. The moment of the question matters most. With the same tracks and overlays, asking where the target is at the end of the video gives 40.0\% at 8B. Asking which cup it is under right after it is hidden gives 48.7\% ($+8.7$, CI [0.5, 16.9], $p{=}0.046$). At 32B the two questions give 44.1\% and 52.3\% ($+8.2$, CI [1.5, 14.9], $p{=}0.026$). The overlay alone adds 6.7 and 4.1 points over the direct answer, and the moment of the question adds the rest. Drawing the tracks and asking the benchmark question, without any bind question, gives 42.1\% at 32B. This is 17.9 points below the single configuration on the same videos (60.0\%, CI [9.2, 27.2]). On card monte the same variant reaches 100\% at 32B, as on the test set. The bind answer itself comes from the VLM. The target's own track hands its identity to a cup in only 56 of 195 videos, and that hand-off is right in 48\% of them.

Tracking rate and grounding decide whether the tracks are usable. On card monte, tracking at 10\,fps instead of 30\,fps lowers accuracy from 56 to 21 of 56 videos, close to the direct answer (16). Identity breaks where identical cards cross quickly. The dense pass of step~2 reproduces the 30\,fps result without a fixed rate. Grounding on 16 frames instead of 8 lowers TOC-Bench accuracy by 2.5 points and card-monte accuracy by 3 videos, and it nearly doubles the VLM calls.

Two choices concern the read-out. Asking whether the person is opposite the camera, on four frames, gives 57.9\% on the development cup games. Asking whether the person faces the camera, on the last frame, gives 48.2\%. This wording was selected among six on the same videos, and on the test videos the single configuration is 6.5 (8B) and 1.9 points (32B) above the tuned variants. On TOC-Bench, visibility segments alone reach 37.4\% of the development questions. Adding the event list and the direction of motion raises this to 42.1\% ($+4.6$, CI [1.5, 7.7]).

\section{Additional Qualitative Examples}\label{app:gallery}
Figures~\ref{fig:gallery_bind} and~\ref{fig:gallery_toc} show six more test items from the logged Qwen3-VL-8B run. The images are the agent's own tracks on the frames it read. The direct answers are those of Table~\ref{tab:main}.

Rows (a) and (b) are the two branches of a located moment. In (a) the bind clip is synthetic and short. In (b) the hidden object is a crumpled paper in a real video, and the viewpoint fact mirrors the camera order. Row (c) is one of the five real cup games that are never bound (Table~\ref{tab:branches}). The VLM grounds the pen holder as the only cup, so the agent never sees two carrier tracks.

Rows (d) and (e) are typical TOC-Bench gains. The timeline states when the object disappears and whether it vanished at the border or inside the frame, which the direct answer misses on its 32 sampled frames. Row (f) is a conditional-state question about two objects. The plan names only one of them, so the timeline cannot place the other object's event. This failure fits the conditional-state results in Table~\ref{tab:dims}.

\begin{figure*}[h!]
\centering
\input{figs/gallery_bind.tex}
\caption{\textbf{Identity questions} (test videos, Qwen3-VL-8B). Numbered boxes are the agent's carrier tracks; magenta is the target's own track. (a, b) Located evidence moments. (c) A grounding failure that leaves one carrier track, so the question is not bound.}
\label{fig:gallery_bind}
\end{figure*}

\begin{figure*}[h!]
\centering
\input{figs/gallery_toc.tex}
\caption{\textbf{TOC-Bench timeline questions} (test videos, Qwen3-VL-8B). Magenta: the subject's track box. Bars: mask area over its 90th-percentile area, coloured by the visibility state (dark green: fully visible; light green: partly hidden; grey: not visible); dashed and dotted lines mark the thresholds 0.6 and 0.12.}
\label{fig:gallery_toc}
\end{figure*}

\clearpage
\section{Worked Examples}\label{app:examples}
Figures~\ref{fig:walk}--\ref{fig:walktoc} follow three test items through the agent, next to the direct answer of the same VLM (Qwen3-VL-8B). Each panel lists what the VLM sees, the prompt it receives, its reply and the number of calls. The prompts are listed verbatim in Table~\ref{tab:uprompts}.

\begin{figure}[h!]
\centering
\resizebox{\linewidth}{!}{\input{figs/walk_pt.tex}}
\caption{\textbf{Hidden target} (Perception Test cup game \texttt{video\_6159}, test set). \emph{Grey:} Qwen3-VL-8B answering directly from the video, one VLM call. \emph{Blue:} Track, Then Bind with the same VLM and the SAM2.1 tracker. Badges give the calls of each step: orange for the VLM, green for the tracker sessions, grey for plain code. Prompts are quoted from the run and shortened with ``\ldots''. This item takes the most common branch on real cup games: the target's own track gives no moment, and the bind question is asked over the whole numbered clip.}
\label{fig:walk}
\end{figure}

\clearpage
\begin{figure}[h!]
\centering
\resizebox{\linewidth}{!}{\input{figs/walk_card.tex}}
\caption{\textbf{Target recognised by its look} (VET-Bench card monte \texttt{card\_003}, test set). The VLM is asked once, on a 1.3\,s clip while the queen is face up; the 30\,fps tracks carry the identity through the shuffle. Track \#1 (red) is hard to see against the red card backs; this is the overlay the VLM receives.}
\label{fig:walkcard}
\end{figure}

\begin{figure}[h!]
\centering
\resizebox{\linewidth}{!}{\input{figs/walk_toc.tex}}
\caption{\textbf{Timeline question} (TOC-Bench \texttt{8b0c33d38075}, test set). The plan finds no target to tell apart, so the agent skips binding and answers with a timeline. Both runs see the same 32 frames and the same template; the only difference is the evidence text inserted after the question. It is built from the tracked mask of the eggs, without a model call.}
\label{fig:walktoc}
\end{figure}

\clearpage

\section{Prompts}\label{app:prompts}
\newcommand{\pr}[1]{\textcolor{blue!35!black}{\sffamily\scriptsize #1}}
\begin{table}[h!]
\caption{\textbf{Unified agent's prompts, verbatim} (one set for all benchmarks). Slots in \{braces\} are filled from the plan, the tracks or the benchmark item.}
\label{tab:uprompts}
\scriptsize
\begin{tabularx}{\textwidth}{@{}>{\raggedright\arraybackslash}p{2.6cm}>{\raggedright\arraybackslash}p{2.6cm}>{\raggedright\arraybackslash}X@{}}
\toprule
Step & Input & Prompt\\
\midrule
Plan (1) & whole video, 2\,fps & \pr{You plan how to answer a question about a video. Question: \{question\} Reply with one JSON object and nothing else: \{"entities": [\{"name": \ldots, "role": "target" $|$ "carrier" $|$ "subject"\}], "identity\_cue": "visible" $|$ "contained" $|$ "none", "category": \ldots, "asks\_final\_location": true $|$ false, "viewpoint": "camera" $|$ "person"\}} followed by one rule per field (target; carriers that move, hide or cover the target or look the same as it; subjects; recognised by its look vs.\ hidden in a carrier; final location; whose left and right)\\
Final location (1) & text only & \pr{Question about a video: \{question\} Does this question ask where a particular object is located at the end of the video (for example, which container it is in, or whether it is on the left, in the middle or on the right)? Reply YES or NO.}\\
Ground (8, +8 if missing) & one frame & \pr{Locate every instance of the following objects in the image, if visible: \{entity 1\}; \{entity 2\}; \ldots\ Output a JSON list of \{"bbox\_2d": [x1, y1, x2, y2], "label": "<one of the names above>"\}; output [] if none is visible.}\\
Bind, by its look (1 per moment) & numbered tracks, 4\,fps & \pr{These frames are from a video. Each \{carrier\} has a numbered box (1, 2, 3) drawn on it; a number always stays with the same physical object. Which numbered \{carrier\} is the \{target\}? Reply with a single number, or NONE if the \{target\} cannot be seen.}\\
Bind, hidden (1 per moment) & numbered tracks, 4\,fps & \pr{\ldots\ Which numbered \{carrier\} is the \{target\} under or inside right after it is hidden? Reply with a single number, or NONE if you cannot tell.}\\
Viewpoint (1, if asked) & 4 uniform frames & \pr{Is the person who moves the objects positioned opposite the camera, on the other side of the table (so the camera looks at them from the front)? Reply YES or NO.}\\
Read out, bound (1) & text only & \pr{Facts established by tracking every object through the whole video (identity was fixed when the target could be seen, and kept by the tracker afterwards): - The \{target\} was hidden under \{carrier\} \#\{k\}. The tracker followed every \{carrier\} through the rest of the video. In the last frame where all \{n\} are tracked (\{t\} s), from the camera's left to right they are: \#\ldots. So the \{target\} is under the \{ordinal\} \{carrier\} from the camera's left. [- viewpoint fact - person-view position] Answer the question using these facts.} then the benchmark prompt, then \pr{The facts above are reliable. Reply with only the final answer in \textbackslash boxed\{X\}.}\\
Read out, not bound (1) & the benchmark's frames & the benchmark's prompt with \pr{Automatic tracker evidence (may contain errors; the video frames take precedence): \{video length; one timeline line per subject\}} after the question\\
\bottomrule
\end{tabularx}
\end{table}

\begin{table}[h!]
\caption{\textbf{Prompts of the per-benchmark configurations, verbatim.} Slots in \{braces\} are filled by the agent from earlier steps or from the benchmark item. These configurations are the comparison of Table~\ref{tab:onecfg}.}
\label{tab:prompts}
\scriptsize
\begin{tabularx}{\textwidth}{@{}>{\raggedright\arraybackslash}p{2.6cm}>{\raggedright\arraybackslash}p{2.6cm}>{\raggedright\arraybackslash}X@{}}
\toprule
Step & Input & Prompt\\
\midrule
\multicolumn{3}{@{}l}{\textit{Direct baselines (one call; the benchmark's own prompt)}}\\
Perception Test & video, 8\,fps & \pr{\{question\}\textbackslash n\textbackslash nA. \{option\}\textbackslash nB. \{option\}\textbackslash nC. \{option\}\textbackslash n\textbackslash nPlease select the correct answer from the options above. Put your final answer in \textbackslash boxed\{X\} where X is the letter of your choice.}\\
VET-Bench card monte & video, 8\,fps & \pr{Where is the \{target\} at the end of the video?\textbackslash n(A) Left\textbackslash n(B) Middle\textbackslash n(C) Right\textbackslash nOutput your final answer (A, B, or C) in the format: \$\textbackslash boxed\{<option>\}\$.}\\
TOC-Bench (two-statement format; other formats use their own templates) & 32 uniform frames & \pr{System: You are a video understanding model. Follow output format exactly.\newline User: You will see a video and two statements. Choose the statement that is TRUE according to the video. Output ONLY the letter A or B.\textbackslash n\textbackslash nQuestion: \{question\}\textbackslash nA. \{statement A\}\textbackslash nB. \{statement B\}\textbackslash nAnswer:}\\
SpatialClaw (its question field) & its own frames & the benchmark prompt above, verbatim; for TOC-Bench the system and user messages joined by a blank line\\
\midrule
\multicolumn{3}{@{}l}{\textit{Containment hand-off (Perception Test cup games; 20 calls per item)}}\\
Parse (1) & video, 2\,fps & \pr{In this video a person plays a shell game: a small object is hidden under one of several similar containers, then the containers are moved around. Reply with two short noun phrases separated by '$|$': first, the containers (plural, no number); second, the small object that gets hidden and disappears from view under a container. Ignore other objects that stay visible. No other text.}\\
Ground carriers (8) & one frame & \pr{Locate every \{containers\} in the image. Output a JSON list of \{"bbox\_2d": [x1, y1, x2, y2], "label": "\{containers\}"\}.}\\
Ground target (8) & one frame & \pr{Locate the \{target\} in the image, if it is visible. Output a JSON list of \{"bbox\_2d": [x1, y1, x2, y2], "label": "\{target\}"\}; output [] if it is not visible.}\\
Viewpoint (1) & last frame & \pr{Is the person playing this game sitting or standing across the table, facing the camera? Reply YES if the person faces the camera, or NO if the camera shows the scene from the person's own viewpoint (for example, held by or mounted on the person).}\\
Bind (1) & numbered-track frames, 2\,fps & \pr{In this video a person hides \{target\} under one of several \{containers\} and then moves them around. Each container has a numbered box (1, 2, 3) drawn on it in every frame; a number always stays with the same physical container. Which numbered container is \{target\} under right after it is hidden? Reply with a single number.}\\
Bind, ablation (1) & same & identical, ending \pr{\ldots\ under at the END of the video? Reply with a single number.}\\
\midrule
\multicolumn{3}{@{}l}{\textit{Visible identity (VET-Bench card monte; 5 calls per item)}}\\
Ground (4) & one frame & \pr{Locate every playing card in the image. Output a JSON list of \{"bbox\_2d": [x1, y1, x2, y2], "label": "card"\}.}\\
Bind (1) & numbered-track frames, 4\,fps, first 30\% & \pr{These frames show the start of a card game. Each card has a numbered box (1, 2, 3) drawn on it; a number always stays with the same physical card. Which numbered card shows the \{target\} face up? Reply with a single number.}\\
\midrule
\multicolumn{3}{@{}l}{\textit{Subject timeline (TOC-Bench; 1 + 16 per subject + 1 calls per item)}}\\
Parse subjects (1) & text only & \pr{List the physical objects or people that this video question is about, as short noun phrases exactly as written, one per line, at most 3. No other text.\textbackslash n\textbackslash n\{question\}\textbackslash n\{statements or events\}}\\
Ground (16 per subject) & one frame & \pr{Locate the \{subject\} in the image, if it is visible. Output a JSON list of \{"bbox\_2d": [x1, y1, x2, y2], "label": "\{subject\}"\}; output [] if not visible.}\\
Answer (1) & 32 uniform frames & the TOC-Bench template above, with \pr{\{question\}} replaced by \pr{\{question\}\textbackslash n\textbackslash nAutomatic tracker evidence (may contain errors; the video frames take precedence):\textbackslash n\{evidence lines\}}\\
\bottomrule
\end{tabularx}
\end{table}

\end{document}

%% file: figs/walk_common.tex
\definecolor{bandbase}{RGB}{243,243,243}
\definecolor{bandours}{RGB}{232,242,252}
\definecolor{okgreen}{RGB}{30,130,40}
\definecolor{badred}{RGB}{190,30,30}
\tikzset{
  wq/.style={draw=black!50, rounded corners=3pt, fill=white, inner sep=3pt, align=left, anchor=north west, font=\fontsize{7}{8.4}\selectfont},
  wband/.style={rounded corners=5pt, draw=#1!60!black, line width=0.6pt},
  whead/.style={anchor=north west, align=left, inner sep=2pt, font=\fontsize{7}{8.4}\selectfont},
  wcard/.style={draw=black!35, rounded corners=2pt, fill=white, inner sep=2.5pt, align=left, anchor=north west, font=\fontsize{6.3}{7.4}\selectfont},
  wbadge/.style={rounded corners=1.5pt, inner sep=1.2pt, font=\fontsize{6}{6.5}\selectfont\bfseries, text=black},
  warrow/.style={-{Stealth[length=4pt]}, line width=0.8pt, black!55},
  wlabel/.style={font=\fontsize{6}{6.5}\selectfont, text=black!70, inner sep=1pt},
}
\newcommand{\wbadgebox}[3]{{\setlength{\fboxsep}{1.2pt}\setlength{\fboxrule}{0.5pt}\fcolorbox{#1}{#2}{\fontsize{6}{6.5}\selectfont\bfseries #3}}}
\newcommand{\vlmb}[1]{\wbadgebox{orange!60!black}{vlm}{VLM $\times$#1}}
\newcommand{\toolb}[1]{\wbadgebox{green!45!black}{trk}{#1}}
\newcommand{\nocall}{\wbadgebox{black!40}{black!8}{no model call}}
\newcommand{\sees}{\textbf{Sees:}~}
\newcommand{\asks}{\textbf{Asks:}~}
\newcommand{\rets}{\textbf{Returns:}~}
\newcommand{\pq}[1]{\textcolor{blue!35!black}{\textsf{``#1''}}}
\newcommand{\cmark}{\textcolor{okgreen}{\ding{51}}}
\newcommand{\xmark}{\textcolor{badred}{\ding{55}}}

%% file: figs/gallery_common.tex
\providecommand{\gthumb}[3][1.55cm]{\tikz[baseline=(i.center)]{\node[inner sep=0] (i) {\includegraphics[height=#1]{#2}};
  \node[anchor=north east, fill=black!60, text=white, font=\sffamily\fontsize{5.4}{6}\selectfont, inner sep=1pt] at (i.north east) {#3};}}
\providecommand{\gcap}[1]{\begin{minipage}{3.05cm}\raggedright\sffamily\fontsize{6.3}{7.5}\selectfont #1\end{minipage}}
\providecommand{\gnote}[2][4.3cm]{\fbox{\begin{minipage}{#1}\raggedright\sffamily\fontsize{6.1}{7.3}\selectfont #2\end{minipage}}}

%% file: figs/overview.tex
\definecolor{tairInk}{HTML}{183449}
\definecolor{tairBlue}{HTML}{28618C}
\definecolor{tairBlueLight}{HTML}{EEF4F9}
\definecolor{tairGreen}{HTML}{34785B}
\definecolor{tairGreenLight}{HTML}{EFF6F2}
\definecolor{tairAmber}{HTML}{996329}
\definecolor{tairAmberLight}{HTML}{FBF4EA}
\definecolor{tairMuted}{HTML}{5D6973}
\definecolor{tairRule}{HTML}{D8E0E6}
\definecolor{tairRed}{HTML}{A35050}
\begin{tikzpicture}[x=1cm,y=1cm,
  font=\sffamily\fontsize{7.6}{9.2}\selectfont,text=tairInk,
  image/.style={anchor=north west,inner sep=0},
  textblock/.style={anchor=north west,inner sep=0,align=left,execute at begin node={\hyphenpenalty=10000\exhyphenpenalty=10000}},
  small/.style={font=\sffamily\fontsize{7}{8.4}\selectfont},
  outcome/.style={anchor=north west,rounded corners=3pt,inner sep=2pt,
    align=left,text width=2.55cm,minimum height=0.65cm,font=\sffamily\fontsize{7.1}{8}\selectfont}]
\path[use as bounding box] (0,0.04) rectangle (17.8,-6.90);
\def\D{figs/ex3/pt}
\fill[tairBlueLight,rounded corners=3pt] (0,0) rectangle (17.8,-0.58);
\node[anchor=west,inner sep=0] at (0.18,-0.29)
  {\textbf{Question}\quad Where is the hidden object at the end of the game, from the person's point of view?};
\node[textblock,text=tairMuted,text width=3.0cm] at (0,-1.10)
  {\textbf{One object, similar mugs}\\[4pt]The lime is visible early.\\Its identity must survive\\the shuffle.};
\foreach \i/\x/\time/\label in {
  1/3.45/4.0/Identifying evidence,
  3/7.25/10.3/Shuffle,
  4/11.05/13.5/Final state} {
  \node[anchor=south west,inner sep=0,font=\sffamily\bfseries\fontsize{8}{9}\selectfont]
    at (\x,-0.97) {\label};
  \node[image] (raw\i) at (\x,-1.10) {\includegraphics[width=3.5cm]{\D/in\i.jpg}};
  \node[anchor=north east,fill=tairInk!85,text=white,inner sep=2pt,small]
    at (raw\i.north east) {\time\,s};
}
\node[textblock,small,text=tairMuted,text width=2.85cm] at (14.85,-1.20)
  {The lime starts under\\the \textbf{middle mug}.\\[6pt]After the shuffle,\\the mugs alone do not\\reveal its identity.};
\fill[black!3,rounded corners=3pt] (3.30,-3.38) rectangle (14.65,-4.03);
\node[textblock,font=\sffamily\bfseries\fontsize{9}{10}\selectfont,text=tairMuted]
  at (0,-3.47) {Direct Video-LM};
\node[textblock,small,text=tairMuted] at (0,-3.88) {Identity held inside the model};
\draw[tairMuted,line width=1pt] (3.50,-3.68) -- (6.83,-3.68);
\draw[tairMuted,line width=1pt,dash pattern=on 3pt off 2pt,-{Stealth[length=4pt]}]
  (6.83,-3.68) -- (14.43,-3.68);
\node[anchor=west,small,inner sep=0,text=tairMuted] at (3.50,-3.98)
  {Reconstruct the target's identity through the shuffle};
\node[outcome,draw=tairRed!45,fill=tairRed!4,text=tairRed] at (14.85,-3.38)
  {\textbf{B\enspace Second}\hfill\ding{55}\\[-1pt]{\fontsize{7}{8}\selectfont``Middle mug''}};
\node[textblock,font=\sffamily\bfseries\fontsize{9}{10}\selectfont,text=tairBlue]
  at (0,-4.31) {Track, Then Bind};
\node[textblock,text width=3.0cm] at (0,-4.84)
  {\textcolor{tairAmber}{\textbf{Bind identity}}\\[3pt]
   Which numbered mug held the lime just after it was hidden?\\[5pt]
   \textbf{Lime $\rightarrow$ track \#2}};
\fill[tairBlueLight,rounded corners=3pt] (3.30,-4.20) rectangle (14.65,-6.84);
\foreach \i/\x in {1/3.45,3/7.25,4/11.05} {
  \node[image] at (\x,-4.34) {\includegraphics[width=3.5cm]{\D/ov\i.jpg}};
}
\draw[tairBlue,line width=1.1pt,-{Stealth[length=4pt]}] (3.53,-6.52) -- (14.40,-6.52);
\node[anchor=north,small,inner sep=0,text=tairBlue] at (9.00,-6.62)
  {Persistent track \#2 carries the bound identity};
\fill[tairGreenLight,rounded corners=3pt] (14.85,-4.20) rectangle (17.8,-6.84);
\node[textblock,text width=2.60cm] at (15.02,-4.38)
  {\textcolor{tairGreen}{\textbf{Read tracked facts}}\\[5pt]
   Camera order:\\ \textbf{\#2\quad\#3\quad\#1}\\[5pt]
   The person faces the\\camera: \#2 is third.\\[7pt]
   \textcolor{tairGreen}{\textbf{C\enspace Third}\quad\ding{51}}};
\end{tikzpicture}

%% file: figs/pipeline.tex
\definecolor{tairInk}{HTML}{183449}
\definecolor{tairBlue}{HTML}{28618C}
\definecolor{tairBlueLight}{HTML}{EEF4F9}
\definecolor{tairGreen}{HTML}{34785B}
\definecolor{tairGreenLight}{HTML}{EFF6F2}
\definecolor{tairAmber}{HTML}{996329}
\definecolor{tairAmberLight}{HTML}{FBF4EA}
\definecolor{tairMuted}{HTML}{5D6973}
\definecolor{tairRule}{HTML}{D8E0E6}
\definecolor{tairPurple}{HTML}{6B5A84}
\begin{tikzpicture}[x=1cm,y=1cm,
  font=\sffamily\fontsize{7.5}{9}\selectfont,text=tairInk,
  txt/.style={anchor=north west,inner sep=0,align=left,execute at begin node={\hyphenpenalty=10000\exhyphenpenalty=10000}},
  image/.style={anchor=north west,inner sep=0},
  head/.style={txt,font=\sffamily\bfseries\fontsize{8.6}{10}\selectfont},
  small/.style={font=\sffamily\fontsize{7}{8.4}\selectfont},
  flow/.style={-{Stealth[length=4pt,width=3.5pt]},line width=0.8pt,draw=tairMuted},
  badge/.style={anchor=north west,rounded corners=2pt,inner sep=3pt,
    font=\sffamily\bfseries\fontsize{7}{8}\selectfont}]
\path[use as bounding box] (0,0.05) rectangle (18.22,-7.98);
\def\M{figs/ex3/m}
\fill[black!2,rounded corners=3pt] (0,0) rectangle (2.75,-2.00);
\fill[tairAmberLight,rounded corners=3pt] (3.12,0) rectangle (6.90,-2.00);
\fill[tairGreenLight,rounded corners=3pt] (7.27,0) rectangle (12.17,-2.00);
\fill[tairBlueLight,rounded corners=3pt] (12.54,0) rectangle (17.8,-2.00);
\node[head] at (0.17,-0.16) {Video + question};
\node[image] at (0.17,-0.65) {\includegraphics[width=1.10cm]{\M/card_in.jpg}};
\node[image] at (1.43,-0.65) {\includegraphics[width=1.10cm]{\M/egg_in.jpg}};
\node[txt,small,text=tairMuted] at (0.17,-1.45) {Identity, visibility or events};
\node[head,text=tairAmber] at (3.30,-0.16) {1\quad Plan};
\node[txt,text width=3.36cm] at (3.30,-0.65)
  {Target, carriers, subjects\\Identity cue and viewpoint\\Final location?};
\node[txt,small,text=tairMuted] at (3.30,-1.64) {Frozen VLM\enspace$\cdot$\enspace 2\,fps};
\node[head,text=tairGreen] at (7.45,-0.16) {2\quad Perceive};
\node[txt,text width=4.47cm] at (7.45,-0.65)
  {VLM boxes on 8 sampled frames\\SAM2.1 instance tracking at 6\,fps\\Adaptive dense pass};
\node[txt,small,text=tairMuted] at (7.45,-1.64) {Boxes and tracks shared across questions};
\node[head,text=tairBlue] at (12.72,-0.16) {Persistent, numbered tracks};
\node[image] at (12.72,-0.65) {\includegraphics[width=1.62cm]{\M/card_trk.jpg}};
\node[image] at (14.78,-0.65) {\includegraphics[width=1.62cm]{\M/egg_trk.jpg}};
\node[txt,small,text=tairMuted] at (12.72,-1.75) {Object masks, IDs and positions over time};
\foreach \xa/\xb in {2.80/3.07,6.95/7.22,12.22/12.49}
  \draw[flow] (\xa,-1.00) -- (\xb,-1.00);
\draw[tairMuted!65,line width=0.7pt] (17.82,-1.00) -- (18.12,-1.00) -- (18.12,-5.60);
\draw[flow,tairBlue] (18.12,-2.72) -- (17.82,-2.72);
\draw[flow,tairGreen] (18.12,-5.60) -- (17.82,-5.60);
\fill[tairBlueLight,rounded corners=3pt] (0,-2.36) rectangle (17.8,-5.03);
\node[head,text=tairBlue] at (0.18,-2.53) {Identity binding};
\node[txt,small,text width=2.43cm] at (0.18,-3.08)
  {\emph{Where is the Queen of Hearts at the end?}\\[6pt]
   \textcolor{tairMuted}{Bind the target to one\\persistent carrier ID.}};
\draw[tairBlue!20] (2.83,-2.60) -- (2.83,-4.82);
\node[head,text=tairBlue] at (3.07,-2.53) {3\quad Find evidence};
\node[image] (bindframe) at (3.07,-3.06) {\includegraphics[width=2.44cm]{\M/card_bind.jpg}};
\node[txt,small,text=tairMuted,text width=2.85cm] at (3.07,-4.52)
  {0.0\,s: queen visible\\1.3\,s bind clip};
\node[head,text=tairAmber] at (6.13,-2.53) {4\quad Bind};
\node[txt,text width=3.01cm] at (6.13,-3.15)
  {\emph{Which numbered card\\is the Queen of Hearts?}};
\node[badge,fill=white,text=tairBlue] at (6.13,-4.02) {Queen $\rightarrow$ track \#3};
\node[txt,small,text=tairMuted] at (6.13,-4.61) {Frozen VLM\enspace$\cdot$\enspace numbered clip};
\draw[flow,tairBlue] (5.57,-3.72) -- (5.97,-3.72);
\draw[flow,tairBlue] (9.26,-3.72) -- (9.66,-3.72);
\node[head,text=tairPurple] at (9.87,-2.53) {5a\quad Read tracked facts};
\node[image] at (9.87,-3.12) {\includegraphics[width=1.88cm]{\M/card_final.jpg}};
\node[txt,text width=2.25cm] at (12.00,-3.20)
  {Last-frame order:\\ \textbf{\#3\quad\#2\quad\#1}\\[4pt]Leftmost: \textbf{\#3}};
\node[txt,small,text=tairMuted] at (9.87,-4.61) {Frozen VLM\enspace$\cdot$\enspace text only};
\draw[flow,tairBlue] (14.33,-3.72) -- (14.78,-3.72);
\node[badge,fill=white,text=tairGreen,text width=2.17cm,align=left] at (15.02,-3.25)
  {Answer\\[5pt](A) Left\quad\ding{51}};
\fill[tairGreenLight,rounded corners=3pt] (0,-5.25) rectangle (17.8,-7.62);
\node[head,text=tairGreen] at (0.18,-5.44) {Visibility / events};
\node[txt,small,text width=2.43cm] at (0.18,-6.00)
  {\emph{Do the eggs ever get\\fully hidden?}\\[6pt]
   \textcolor{tairMuted}{No identity binding.}};
\draw[tairGreen!20] (2.83,-5.49) -- (2.83,-7.41);
\node[head,text=tairGreen] at (3.07,-5.44) {Build a visibility timeline};
\begin{scope}[shift={(3.07,-6.10)}]\input{figs/ex3/toc/timeline_fig2.tex}\end{scope}
\node[txt,small,text=tairMuted] at (3.07,-7.19) {Mask area over time\enspace$\rightarrow$\enspace visibility events};
\draw[flow,tairGreen] (7.22,-6.52) -- (7.66,-6.52);
\node[txt,text width=2.73cm] at (7.88,-6.03)
  {\textbf{Tracked events}\\[4pt]2.0\,s: fully hidden\\3.3\,s: visible again};
\draw[flow,tairGreen] (10.73,-6.52) -- (11.17,-6.52);
\node[head,text=tairPurple] at (11.38,-5.44) {5b\quad Read with a timeline};
\node[txt,text width=3.01cm] at (11.38,-6.03)
  {Events + video frames\\[5pt]Frozen VLM};
\node[txt,small,text=tairMuted] at (11.38,-7.19) {Benchmark prompt; 32 frames};
\draw[flow,tairGreen] (14.33,-6.52) -- (14.78,-6.52);
\node[badge,fill=white,text=tairGreen,text width=2.17cm,align=left] at (15.02,-6.03)
  {Answer\\[5pt](B) Gets hidden\ \ding{51}};
\node[anchor=north west,inner sep=0,small,text=tairMuted] at (0,-7.75)
  {All models frozen\quad$\cdot$\quad One configuration across benchmarks\quad$\cdot$\quad Examples from logged Qwen3-VL-8B test runs};
\end{tikzpicture}

%% file: figs/tab_budget.tex
\begin{tabular}{llccccrr}
\toprule
VLM & Baseline & cup & card & PT & TOC & Calls & Tokens\\
\midrule
8B & direct answer & 28.0 & 28.0 & 33.6 & 25.1 & 1 & 12k\\
 & self-consistency, as many calls & 28.0 & 30.0 & 46.7 & 25.9 & 16/12/13/11 & 157k\\
 & direct answer, more input tokens & 24.0 & 30.0 & 27.1 & 24.7 & 1 & 34k\\
 & \cellcolor{ours}Track, Then Bind & \cellcolor{ours}\textbf{100} & \cellcolor{ours}\textbf{100} & \cellcolor{ours}\textbf{51.4} & \cellcolor{ours}\textbf{43.0} & \cellcolor{ours}19/12/13/11 & \cellcolor{ours}25k\\
\midrule
32B & direct answer & 32.0 & 38.0 & 35.5 & 37.1 & 1 & 12k\\
 & self-consistency, as many calls & 32.0 & 30.0 & 37.4 & 37.3 & 16/12/15/11 & 181k\\
 & direct answer, more input tokens & 30.0 & 24.0 & 31.8 & 36.7 & 1 & 34k\\
 & SpatialClaw code agent & 28.0 & 36.0 & 26.2 & 35.7 & 14/9/18/10 & 508k\\
 & \cellcolor{ours}Track, Then Bind & \cellcolor{ours}\textbf{100} & \cellcolor{ours}\textbf{100} & \cellcolor{ours}\textbf{52.3} & \cellcolor{ours}\textbf{48.6} & \cellcolor{ours}19/12/15/11 & \cellcolor{ours}27k\\
\bottomrule
\end{tabular}

%% file: figs/tab_mech.tex
\begin{tabular}{llccc}
\toprule
VLM & Variant & cup & card & PT\\
\midrule
8B & tracks drawn, question asked directly & 44.0 & 40.0 & 32.7\\
 & bind question on the whole video & 100 & 100 & 53.3\\
 & bind question about the end & 28.0 & 36.0 & 38.3\\
 & facts read out with the video & 38.0 & 72.0 & 46.7\\
 & \cellcolor{ours}Track, Then Bind & \cellcolor{ours}100 & \cellcolor{ours}100 & \cellcolor{ours}51.4\\
\midrule
32B & tracks drawn, question asked directly & 78.0 & 100 & 40.2\\
 & bind question on the whole video & 72.0 & 100 & 54.2\\
 & bind question about the end & 46.0 & 88.0 & 49.5\\
 & facts read out with the video & 76.0 & 96.0 & 48.6\\
 & \cellcolor{ours}Track, Then Bind & \cellcolor{ours}100 & \cellcolor{ours}100 & \cellcolor{ours}52.3\\
\bottomrule
\end{tabular}

%% file: figs/qual.tex
\begin{tikzpicture}[x=1cm, y=1cm, font=\sffamily\fontsize{6.4}{7.6}\selectfont,
  qcap/.style={anchor=north west, align=left, inner sep=2pt, text width=3.1cm},
  note/.style={anchor=north west, align=left, inner sep=2.5pt, text width=5.55cm, draw=black!30, rounded corners=2pt, fill=white},
  tl/.style={anchor=north east, font=\sffamily\fontsize{5.6}{6}\selectfont, text=white, fill=black!60, inner sep=1pt}]
\def\D{figs/ex3}
\def\TW{2.35}
\node[qcap] at (0,0) {\textbf{(a) Identity carried through the shuffle} (VET card \texttt{card\_001}). Queen of Hearts at the end? Gold: \textbf{right}.};
\foreach \n/\t [count=\i from 0] in {t0/0.0\,s, t1/4.0\,s, t2/12.0\,s} {
  \node[anchor=north west, inner sep=0] (a\i) at (3.25+2.45*\i, 0) {\includegraphics[width=\TW cm]{\D/q_card/\n.jpg}};
  \node[tl] at (a\i.north east) {\t};
}
\node[note] at (10.65,0) {The 16-sample vote, the direct answer with more input tokens, SpatialClaw (32B), the numbered tracks drawn on the video and the same facts read out with the video all answer \emph{middle}; the plain direct answer names no option. Ours binds the queen to \#2 at 0.0\,s, and \#2 ends on the right. \textcolor{okgreen}{\ding{51}}};
\node[qcap] at (0,-2.2) {\textbf{(b) Viewpoint error} (PT \texttt{video\_2453}). Hidden object at the end, from the person's point of view? Gold: \textbf{first}.};
\foreach \n/\t [count=\i from 0] in {t0/0.5\,s, t1/1.5\,s, t2/14.5\,s, view/viewpoint frame} {
  \node[anchor=north west, inner sep=0] (b\i) at (3.25+2.45*\i, -2.2) {\includegraphics[width=\TW cm]{\D/q_view/\n.jpg}};
  \node[tl] at (b\i.north east) {\t};
}
\node[note, text width=3.1cm] at (13.1,-2.2) {Bind and tracks are right: lime under \#1, last frame \#2,\#3,\#1. The viewpoint question returns \pq{NO} although the person faces the camera, so the answer is mirrored. \textcolor{badred}{\ding{55}}};
\node[qcap] at (0,-4.4) {\textbf{(c) Tracking error} (PT \texttt{video\_10121}). Gold: \textbf{first} from the person's left.};
\foreach \n/\t [count=\i from 0] in {t0/4.0\,s, t1/10.1\,s, t2/12.1\,s, t3/17.3\,s} {
  \node[anchor=north west, inner sep=0] (c\i) at (3.25+2.45*\i, -4.4) {\includegraphics[width=\TW cm]{\D/q_track/\n.jpg}};
  \node[tl] at (c\i.north east) {\t};
}
\node[note, text width=3.1cm] at (13.1,-4.4) {The cube is bound to \#2 at 4.9\,s, correctly, and \#2 follows it to the right. When both hands cross the cups, \#3 merges onto the same cup, so the middle cup has no track and the count from the left is off. \textcolor{badred}{\ding{55}}};
\end{tikzpicture}

%% file: figs/app_cost.tex
\begin{tabular}{llrrrrrrrrrr}
\toprule
 & & \multicolumn{5}{c}{VLM calls} & \multicolumn{2}{c}{VLM tokens} & \multicolumn{3}{c}{Time (s)}\\
\cmidrule(lr){3-7}\cmidrule(lr){8-9}\cmidrule(lr){10-12}
VLM & Set & video & frame & clip & text & all & in (k) & out & VLM & tracker & total\\
\midrule
8B & cup & 1.0 & 15.0 & 1.0 & 2.0 & 19.0 & 10.8 & 1058 & 40 & 69 & 110\\
 & card & 1.0 & 8.0 & 1.0 & 2.0 & 12.0 & 7.3 & 987 & 37 & 64 & 103\\
 & PT & 1.1 & 9.3 & 1.0 & 1.9 & 13.4 & 24.7 & 1289 & 50 & 71 & 126\\
 & TOC & 2.1 & 7.8 & 0.0 & 1.0 & 10.8 & 15.3 & 443 & 18 & 14 & 35\\
\midrule
32B & cup & 1.0 & 15.0 & 1.0 & 2.0 & 19.0 & 10.8 & 984 & 91 & 62 & 154\\
 & card & 1.0 & 8.0 & 1.0 & 2.0 & 12.0 & 7.3 & 216 & 22 & 64 & 87\\
 & PT & 1.2 & 10.7 & 0.9 & 1.9 & 14.7 & 27.0 & 929 & 98 & 56 & 159\\
 & TOC & 2.0 & 8.1 & 0.0 & 1.0 & 11.1 & 15.3 & 373 & 39 & 12 & 54\\
\bottomrule
\end{tabular}

%% file: figs/app_spatialclaw.tex
\begin{tabular}{lrrrrrr}
\toprule
Set & $n$ & Acc.\ (\%) & VLM calls & Input tokens & Mask propagation (\%) & Rerun after tool failure\\
\midrule
cup & 50 & 28.0 & 14.5 & 419k & 26 & 0\\
card & 50 & 36.0 & 9.0 & 229k & 26 & 0\\
PT & 107 & 26.2 & 17.7 & 508k & 77 & 12\\
TOC & 1845 & 35.7 & 10.2 & 300k & 62 & 28\\
\bottomrule
\end{tabular}

%% file: figs/gallery_sc.tex
\setlength{\fboxsep}{2.5pt}\setlength{\fboxrule}{0.4pt}
\begin{tabular}{@{}m{2.55cm}@{\hspace{3pt}}m{2.55cm}@{\hspace{3pt}}m{2.55cm}@{\hspace{6pt}}m{8.6cm}@{}}
\includegraphics[width=2.55cm]{figs/ex3/g_sc_card/s1.jpg} & \includegraphics[width=2.55cm]{figs/ex3/g_sc_card/s3.jpg} & \includegraphics[width=2.55cm]{figs/ex3/g_sc_card/s4.jpg} &
\gnote[8.4cm]{\textbf{1}\enspace \texttt{SAM3.segment\_image\_by\_text(last frame, "playing card")}: three masks (left image).\\
\textbf{2}\enspace \texttt{SAM3.segment\_image\_by\_text(first frame, "Queen of Hearts")}: error, no usable mask.\\
\textbf{3}\enspace \texttt{vlm.locate(first frame, "Queen of Hearts")} $\rightarrow$ (500, 496); a point prompt segments the queen (middle image).\\
\textbf{4}\enspace \texttt{SAM3.segment\_video\_by\_points(\ldots, prompt\_frame\_idx=0)}: \pq{Segmented mask area in final frame: 0 pixels} (right image).\\
\textbf{5}\enspace \texttt{vlm.ask\_with\_thinking(last frame, \ldots)}: \pq{Cannot determine from the images.}\\
\textbf{6}\enspace Code comment: \pq{it is most likely still in the middle}; \texttt{ReturnAnswer("B")}.~\xmark\ Gold: C.}\\[-2pt]
{\sffamily\fontsize{5.8}{6.5}\selectfont step 1, last frame} & {\sffamily\fontsize{5.8}{6.5}\selectfont step 3, first frame} & {\sffamily\fontsize{5.8}{6.5}\selectfont step 4, last frame} & \\
\end{tabular}

%% file: figs/tab_dims.tex
\begin{tabular}{lrcccc}
\toprule
 & & \multicolumn{2}{c}{8B} & \multicolumn{2}{c}{32B}\\
\cmidrule(lr){3-4}\cmidrule(lr){5-6}
Dimension & $n$ & D & T & D & T\\
\midrule
event existence & 170 & 34.7 & \textbf{67.1} & 42.9 & \textbf{66.5}\\
reappear vs.\ disappear & 106 & 41.5 & \textbf{71.7} & 41.5 & \textbf{75.5}\\
relative spatial change & 131 & 24.4 & \textbf{51.9} & 31.3 & \textbf{52.7}\\
event count & 241 & 13.3 & \textbf{29.0} & 20.3 & \textbf{32.0}\\
event ordering & 89 & 12.4 & \textbf{27.0} & 13.5 & \textbf{32.6}\\
duration category & 245 & 18.0 & \textbf{34.7} & 31.0 & \textbf{36.7}\\
temporal location & 473 & 20.1 & \textbf{38.7} & 41.4 & \textbf{51.4}\\
cross-object order & 155 & 40.6 & \textbf{49.7} & 49.7 & \textbf{58.1}\\
conditional state & 219 & 33.3 & \textbf{40.2} & \textbf{49.8} & 44.7\\
reappear identity & 16 & \textbf{62.5} & 56.2 & 50.0 & 50.0\\
\bottomrule
\end{tabular}

%% file: figs/app_paired.tex
\begin{tabular}{lllrrr}
\toprule
VLM & Baseline & Set & Baseline (\%) & Ours $-$ baseline [95\% CI] & $p$\\
\midrule
8B & Direct answer (vLLM) & cup & 28.0 & +72.0 [60.0, 84.0] & $<$0.001\\
 &  & card & 28.0 & +72.0 [60.0, 84.0] & $<$0.001\\
 &  & PT & 33.6 & +17.8 [3.7, 30.8] & 0.02\\
 &  & TOC & 25.1 & +17.9 [15.4, 20.4] & $<$0.001\\
 & Direct answer (Transformers) & cup & 22.0 & +78.0 [66.0, 88.0] & $<$0.001\\
 &  & card & 34.0 & +66.0 [52.0, 78.0] & $<$0.001\\
 &  & PT & 31.8 & +19.6 [5.6, 32.7] & 0.009\\
 &  & TOC & 25.2 & +17.8 [15.3, 20.3] & $<$0.001\\
 & Self-consistency, as many calls & cup & 28.0 & +72.0 [60.0, 84.0] & $<$0.001\\
 &  & card & 30.0 & +70.0 [56.0, 82.0] & $<$0.001\\
 &  & PT & 46.7 & +4.7 [$-$9.3, 18.7] & 0.59\\
 &  & TOC & 25.9 & +17.2 [14.7, 19.6] & $<$0.001\\
 & Direct answer, more input tokens & cup & 24.0 & +76.0 [64.0, 88.0] & $<$0.001\\
 &  & card & 30.0 & +70.0 [56.0, 82.0] & $<$0.001\\
 &  & PT & 27.1 & +24.3 [11.2, 36.4] & $<$0.001\\
 &  & TOC & 24.7 & +18.4 [15.9, 20.8] & $<$0.001\\
 & Self-consistency (Transformers) & cup & 28.0 & +72.0 [60.0, 84.0] & $<$0.001\\
 &  & card & 30.0 & +70.0 [56.0, 82.0] & $<$0.001\\
 &  & PT & 43.0 & +8.4 [$-$4.7, 21.5] & 0.25\\
 &  & TOC & 26.0 & +17.0 [14.5, 19.5] & $<$0.001\\
 & More input tokens (Transformers) & cup & 28.0 & +72.0 [60.0, 84.0] & $<$0.001\\
 &  & card & 28.0 & +72.0 [60.0, 84.0] & $<$0.001\\
 &  & PT & 29.0 & +22.4 [9.3, 35.5] & 0.002\\
 &  & TOC & 24.4 & +18.6 [16.2, 21.0] & $<$0.001\\
 & Thinking variant & cup & 28.0 & +72.0 [60.0, 84.0] & $<$0.001\\
 &  & card & 24.0 & +76.0 [64.0, 88.0] & $<$0.001\\
 &  & PT & 37.4 & +14.0 [1.9, 26.2] & 0.04\\
 &  & TOC & 30.0 & +13.0 [10.3, 15.7] & $<$0.001\\
 & Open-o3-Video-v2-8B & TOC & 33.9 & +9.2 [6.4, 11.8] & $<$0.001\\
\midrule
32B & Direct answer (vLLM) & cup & 32.0 & +68.0 [56.0, 80.0] & $<$0.001\\
 &  & card & 38.0 & +62.0 [48.0, 76.0] & $<$0.001\\
 &  & PT & 35.5 & +16.8 [3.7, 29.9] & 0.02\\
 &  & TOC & 37.1 & +11.5 [9.0, 14.0] & $<$0.001\\
 & Direct answer (Transformers) & cup & 24.0 & +76.0 [64.0, 86.0] & $<$0.001\\
 &  & card & 28.0 & +72.0 [58.0, 84.0] & $<$0.001\\
 &  & PT & 27.1 & +25.2 [13.1, 36.4] & $<$0.001\\
 &  & TOC & 36.9 & +11.7 [9.2, 14.2] & $<$0.001\\
 & Self-consistency, as many calls & cup & 32.0 & +68.0 [54.0, 80.0] & $<$0.001\\
 &  & card & 30.0 & +70.0 [56.0, 82.0] & $<$0.001\\
 &  & PT & 37.4 & +15.0 [2.8, 27.1] & 0.03\\
 &  & TOC & 37.3 & +11.3 [8.8, 13.8] & $<$0.001\\
 & Direct answer, more input tokens & cup & 30.0 & +70.0 [56.0, 82.0] & $<$0.001\\
 &  & card & 24.0 & +76.0 [64.0, 88.0] & $<$0.001\\
 &  & PT & 31.8 & +20.6 [7.5, 33.6] & 0.005\\
 &  & TOC & 36.7 & +11.9 [9.4, 14.5] & $<$0.001\\
 & Self-consistency (Transformers) & cup & 28.0 & +72.0 [60.0, 84.0] & $<$0.001\\
 &  & card & 30.0 & +70.0 [56.0, 82.0] & $<$0.001\\
 &  & PT & 37.4 & +15.0 [1.9, 29.0] & 0.05\\
 &  & TOC & 37.9 & +10.7 [8.2, 13.2] & $<$0.001\\
 & More input tokens (Transformers) & cup & 34.0 & +66.0 [52.0, 78.0] & $<$0.001\\
 &  & card & 16.0 & +84.0 [74.0, 94.0] & $<$0.001\\
 &  & PT & 31.8 & +20.6 [7.5, 33.6] & 0.005\\
 &  & TOC & 36.7 & +11.9 [9.4, 14.5] & $<$0.001\\
 & Thinking variant & cup & 24.0 & +76.0 [64.0, 88.0] & $<$0.001\\
 &  & card & 22.0 & +78.0 [66.0, 88.0] & $<$0.001\\
 &  & PT & 33.6 & +18.7 [6.5, 30.8] & 0.006\\
 &  & TOC & 36.9 & +11.7 [9.2, 14.3] & $<$0.001\\
 & SpatialClaw & cup & 28.0 & +72.0 [60.0, 84.0] & $<$0.001\\
 &  & card & 36.0 & +64.0 [50.0, 76.0] & $<$0.001\\
 &  & PT & 26.2 & +26.2 [14.0, 38.3] & $<$0.001\\
 &  & TOC & 35.7 & +12.9 [10.1, 15.7] & $<$0.001\\
 & VideoAgent & cup & 40.0 & +60.0 [46.0, 74.0] & $<$0.001\\
 &  & card & 28.0 & +72.0 [60.0, 84.0] & $<$0.001\\
 &  & PT & 37.4 & +15.0 [1.9, 28.0] & 0.03\\
 &  & TOC & 33.0 & +15.7 [12.8, 18.6] & $<$0.001\\
 & VideoINSTA & cup & 42.0 & +58.0 [44.0, 72.0] & $<$0.001\\
 &  & card & 28.0 & +72.0 [60.0, 84.0] & $<$0.001\\
 &  & PT & 42.1 & +10.3 [$-$1.9, 23.4] & 0.15\\
 &  & TOC & 34.5 & +14.1 [11.3, 17.0] & $<$0.001\\
\bottomrule
\end{tabular}

%% file: figs/app_budget_hf.tex
\begin{tabular}{llcccc}
\toprule
VLM & Baseline (Transformers) & cup & card & PT & TOC\\
\midrule
8B & direct answer & 22.0 & 34.0 & 31.8 & 25.2\\
 & self-consistency, as many calls & 28.0 & 30.0 & 43.0 & 26.0\\
 & direct answer, more input tokens & 28.0 & 28.0 & 29.0 & 24.4\\
 & \cellcolor{ours}Track, Then Bind & \cellcolor{ours}100 & \cellcolor{ours}100 & \cellcolor{ours}51.4 & \cellcolor{ours}43.0\\
\midrule
32B & direct answer & 24.0 & 28.0 & 27.1 & 36.9\\
 & self-consistency, as many calls & 28.0 & 30.0 & 37.4 & 37.9\\
 & direct answer, more input tokens & 34.0 & 16.0 & 31.8 & 36.7\\
 & \cellcolor{ours}Track, Then Bind & \cellcolor{ours}100 & \cellcolor{ours}100 & \cellcolor{ours}52.3 & \cellcolor{ours}48.6\\
\bottomrule
\end{tabular}

%% file: figs/app_dims32.tex
\begin{tabular}{lrcccccc}
\toprule
Dimension & $n$ & Direct & Thinking & SpatialClaw & VideoAgent & VideoINSTA & Ours\\
\midrule
event existence & 170 & 42.9 & 42.9 & 45.9 & 38.2 & 40.6 & \textbf{66.5}\\
reappear vs.\ disappear & 106 & 41.5 & 43.4 & 55.7 & 34.0 & 33.0 & \textbf{75.5}\\
relative spatial change & 131 & 31.3 & 33.6 & 29.8 & 26.7 & 29.0 & \textbf{52.7}\\
event count & 241 & 20.3 & 19.9 & 7.9 & 35.3 & \textbf{39.4} & 32.0\\
event ordering & 89 & 13.5 & 23.6 & 16.9 & 13.5 & 14.6 & \textbf{32.6}\\
duration category & 245 & 31.0 & 24.9 & 29.0 & 26.1 & 31.4 & \textbf{36.7}\\
temporal location & 473 & 41.4 & 40.2 & 40.2 & 31.3 & 31.1 & \textbf{51.4}\\
cross-object order & 155 & 49.7 & 55.5 & 47.7 & 47.1 & 47.7 & \textbf{58.1}\\
conditional state & 219 & \textbf{49.8} & 47.0 & 48.4 & 37.0 & 36.1 & 44.7\\
reappear identity & 16 & 50.0 & \textbf{56.2} & 50.0 & \textbf{56.2} & \textbf{56.2} & 50.0\\
\midrule
all & 1,845 & 37.1 & 36.9 & 35.7 & 33.0 & 34.5 & \textbf{48.6}\\
\bottomrule
\end{tabular}

%% file: figs/app_ptgroups.tex
\begin{tabular}{lrccccc}
\toprule
Subset & $n$ & D 8B & Ours 8B & D 32B & SpatialClaw & Ours 32B\\
\midrule
2 cups & 13 & 53.8 & 53.8 & 46.2 & 23.1 & 53.8\\
3 cups & 90 & 30.0 & 52.2 & 33.3 & 27.8 & 54.4\\
4 cups & 4 & 50.0 & 25.0 & 50.0 & 0.0 & 0.0\\
\midrule
no swap & 27 & 40.7 & 55.6 & 37.0 & 18.5 & 51.9\\
1--3 swaps & 45 & 22.2 & 53.3 & 28.9 & 28.9 & 51.1\\
$\geq$4 swaps & 35 & 42.9 & 45.7 & 42.9 & 28.6 & 54.3\\
\midrule
PT-65 & 65 & 27.7 & 52.3 & 33.8 & 30.8 & 55.4\\
\bottomrule
\end{tabular}

%% file: figs/gallery_bind.tex
\setlength{\fboxsep}{2.5pt}\setlength{\fboxrule}{0.4pt}
\begin{tabular}{@{}m{3.1cm}@{\hspace{3pt}}m{2.05cm}@{\hspace{2pt}}m{2.05cm}@{\hspace{2pt}}m{2.05cm}@{\hspace{2pt}}m{2.05cm}@{\hspace{4pt}}m{4.5cm}@{}}
\gcap{\textbf{(a) Located moment, synthetic} (VET cup \texttt{cup\_001}). Where is the ball at the end? Gold: \textbf{(C) right}.} &
\gthumb[1.5cm]{figs/ex3/g_cup_001/f0.jpg}{0.0\,s} & \gthumb[1.5cm]{figs/ex3/g_cup_001/f1.jpg}{1.4\,s} & \gthumb[1.5cm]{figs/ex3/g_cup_001/f2.jpg}{6.9\,s} & \gthumb[1.5cm]{figs/ex3/g_cup_001/final.jpg}{12.4\,s} &
\gnote{The ball's own track vanishes under \#1 at 1.5\,s, so the bind clip has 12 frames around that moment. The VLM replies \pq{1}. In the last frame the order is \#3, \#2, \#1, and the text-only read-out answers C.~\cmark\ The direct answer keeps the ball under the left cup (A).~\xmark}\\ \addlinespace[7pt]
\gcap{\textbf{(b) Located moment, real video} (PT \texttt{video\_10863}). Hidden object at the end, from the person's point of view? Gold: \textbf{(A) first}.} &
\gthumb[1.13cm]{figs/ex3/g_video_10863/f0.jpg}{0.4\,s} & \gthumb[1.13cm]{figs/ex3/g_video_10863/f1.jpg}{1.4\,s} & \gthumb[1.13cm]{figs/ex3/g_video_10863/f2.jpg}{5.5\,s} & \gthumb[1.13cm]{figs/ex3/g_video_10863/final.jpg}{10.7\,s} &
\gnote{The paper vanishes under \#3 at 1.4\,s, and the VLM binds it to \#3. The last frame reads \#2, \#1, \#3 from the camera's left. The viewpoint call answers that the person faces the camera, so \#3 is the person's first cup.~\cmark\ The direct answer follows a pointing gesture to the rightmost cup (B).~\xmark}\\ \addlinespace[7pt]
\gcap{\textbf{(c) Grounding failure, no binding} (PT \texttt{video\_2567}). Same question. Gold: \textbf{(A) first}.} &
\gthumb[1.13cm]{figs/ex3/g_video_2567/f0.jpg}{0.5\,s} & \gthumb[1.13cm]{figs/ex3/g_video_2567/f1.jpg}{6.9\,s} & \gthumb[1.13cm]{figs/ex3/g_video_2567/f2.jpg}{10.5\,s} & &
\gnote{On all 8 grounding frames the VLM returns one box for \pq{cups}, on the pen holder (orange). With a single carrier track there is nothing to bind, and the question falls to the timeline read-out. It states only that the key is hidden at 2.5\,s, and the VLM answers C.~\xmark}
\end{tabular}

%% file: figs/gallery_toc.tex
\setlength{\fboxsep}{2.5pt}\setlength{\fboxrule}{0.4pt}
\begin{tabular}{@{}m{3.1cm}@{\hspace{3pt}}m{5.2cm}@{\hspace{3pt}}m{3.6cm}@{\hspace{4pt}}m{4.6cm}@{}}
\gcap{\textbf{(d) Reappear vs.\ disappear} (\texttt{13a41b8c6b50}). Before the red fire extinguisher comes back into view, was it fully blocked or did it leave the frame? Gold: \textbf{(A) blocked}.} &
\gthumb[0.94cm]{figs/ex3/g_13a41b8c6b50/f0.jpg}{1.0\,s}\,\gthumb[0.94cm]{figs/ex3/g_13a41b8c6b50/f1.jpg}{3.5\,s}\,\gthumb[0.94cm]{figs/ex3/g_13a41b8c6b50/f2.jpg}{6.0\,s} &
\includegraphics[width=3.6cm]{figs/ex3/g_13a41b8c6b50/bar.png} &
\gnote[4.4cm]{Timeline: \pq{2.0--5.3\,s not visible (vanished inside the frame, so it was blocked or hidden) \ldots\ 5.3\,s comes back into view}. A person walks past the camera. Ours: A.~\cmark\ Direct: B.~\xmark}\\ \addlinespace[9pt]
\gcap{\textbf{(e) Event existence} (\texttt{de4a73805c89}). Do the coins ever get fully hidden? Gold: \textbf{(A) yes}.} &
\gthumb[0.94cm]{figs/ex3/g_de4a73805c89/f0.jpg}{3.0\,s}\,\gthumb[0.94cm]{figs/ex3/g_de4a73805c89/f1.jpg}{7.5\,s}\,\gthumb[0.94cm]{figs/ex3/g_de4a73805c89/f2.jpg}{15.0\,s} &
\includegraphics[width=3.6cm]{figs/ex3/g_de4a73805c89/bar.png} &
\gnote[4.4cm]{Timeline: \pq{5.1--10.2\,s partly hidden; 10.2--23.9\,s not visible \ldots\ 10.2\,s gets fully hidden and disappears for good}. The coins go into the piggy bank. Ours: A.~\cmark\ Direct: B.~\xmark}\\ \addlinespace[9pt]
\gcap{\textbf{(f) Conditional state, failure} (\texttt{7b08562c551d}). When the metal pot disappears for good, what is the state of the yellow skirt? Gold: \textbf{(A) fully visible}.} &
\gthumb[1.55cm]{figs/ex3/g_7b08562c551d/f0.jpg}{5.0\,s}\,\gthumb[1.55cm]{figs/ex3/g_7b08562c551d/f1.jpg}{25.0\,s}\,\gthumb[1.55cm]{figs/ex3/g_7b08562c551d/f2.jpg}{42.5\,s} &
\includegraphics[width=3.6cm]{figs/ex3/g_7b08562c551d/bar.png} &
\gnote[4.4cm]{The plan lists only the skirt, so the timeline has no time for the pot's disappearance. Its one event is \pq{41.7\,s gets partly hidden}, and the VLM answers B.~\xmark\ The direct answer is A.~\cmark}
\end{tabular}

%% file: figs/walk_pt.tex
\begin{tikzpicture}[x=1cm,y=1cm]
\def\D{figs/ex3/pt}
\node[wq, text width=17.45cm] (q) at (0,0) {\textbf{Question} (Perception Test, test video \texttt{video\_6159}): \emph{The person uses multiple similar objects to play an occlusion game. Where is the hidden object at the end of the game from the person's point of view?}\quad A.~under the first object from the left \quad B.~second \quad \textbf{C.~third} (ground truth)};
\node[whead] at (0,-0.78) {\textbf{Input video}\\13.6\,s, 30\,fps};
\foreach \i/\t in {0/0.0\,s: three mugs, 1/4.0\,s: lime under the middle mug, 2/6.4\,s: right mug lifted, 3/10.3\,s: shuffling, 4/13.5\,s: end} {
  \node[anchor=north west, inner sep=0] (in\i) at (1.9+3.15*\i,-0.75) {\includegraphics[width=2.75cm]{\D/in\i.jpg}};
  \node[wlabel, anchor=north] at (in\i.south) {\t};
}
\fill[bandbase, rounded corners=5pt] (0,-2.58) rectangle (17.8,-3.88);
\draw[wband=black] (0,-2.58) rectangle (17.8,-3.88);
\node[whead, text width=2.9cm] at (0.05,-2.63) {\textbf{Baseline: direct answer}\\Qwen3-VL-8B\\[1pt]\vlmb{1}\ \ no tool call};
\node[wcard, text width=4.6cm] (b1) at (3.2,-2.68) {\sees the whole video at 8\,fps (12.4k input tokens)\\\asks the benchmark prompt: question, options, \pq{Put your final answer in \textbackslash boxed\{X\}}};
\node[wcard, text width=6.6cm] (b2) at (8.35,-2.68) {\rets \pq{\ldots He lifts the second mug (middle) and reveals a lime. \ldots\ Then he shuffles the mugs around, but the lime is still under the middle mug. At the end, he points to the middle mug \ldots\ \textbackslash boxed\{B\}}};
\draw[warrow] (b1.east) -- (b2.west);
\node[whead, align=center] at (15.35,-2.83) {\textbf{Answer: B} \xmark};
\fill[bandours, rounded corners=5pt] (0,-4.13) rectangle (17.8,-10.55);
\draw[wband=blue] (0,-4.13) rectangle (17.8,-10.55);
\node[whead] at (0.05,-4.16) {\textbf{Ours: Track, Then Bind}\quad same Qwen3-VL-8B + SAM2.1 tracker\quad in total \vlmb{13} and \toolb{tracker $\times$4}, 23.3k input tokens};
\node[wcard, text width=5.5cm, minimum height=2.8cm] (s1) at (0.15,-4.68) {\textbf{1. Plan}\hfill\vlmb{2}\\[1pt]
\sees the whole video at 2\,fps\\
\asks the question and \pq{Reply with one JSON object \ldots\ "entities", "identity\_cue", "category", "viewpoint"}; then, text only, \pq{Does this question ask where a particular object is located at the end of the video?}\\
\rets target \pq{hidden object}, carriers \pq{mugs}, cue \pq{contained}, viewpoint \pq{person}; \pq{YES}};
\node[wcard, text width=5.5cm, minimum height=2.8cm] (s2) at (6.0,-4.68) {\textbf{2. Perceive}\hfill\vlmb{8}\ \toolb{SAM2.1 $\times$4}\\[1pt]
\begin{minipage}[t]{2.5cm}\vspace{0pt}\includegraphics[width=2.5cm]{\D/ground.jpg}\\[-1pt]{\fontsize{5.5}{6}\selectfont 3.8\,s; orange: mugs, pink: hidden object}\end{minipage}\hfill
\begin{minipage}[t]{2.9cm}\vspace{0pt}\sees 8 evenly spaced frames, one at a time\\
\asks \pq{Locate every instance of the following objects \ldots: hidden object; mugs.}\\
\rets boxes; SAM2.1 tracks the mugs and the target at 6\,fps, and again at 30\,fps because the mugs cross\end{minipage}};
\node[wcard, text width=5.5cm, minimum height=2.8cm] (s3) at (11.85,-4.68) {\textbf{3. Evidence moment}\hfill\nocall\\[1pt]
\includegraphics[width=1.77cm]{\D/trk0.jpg}\,\includegraphics[width=1.77cm]{\D/trk1.jpg}\,\includegraphics[width=1.77cm]{\D/trk2.jpg}\\
The target's own track never vanishes inside the frame, so no window is cut. The bind question names the moment, and its clip spans the whole video. Tracks shown at 6.4, 9.5, 11.9\,s.};
\node[wcard, text width=5.5cm, minimum height=2.75cm] (s4) at (0.15,-7.6) {\textbf{4. Bind}\hfill\vlmb{1}\\[1pt]
\begin{minipage}[t]{2.5cm}\vspace{0pt}\includegraphics[width=2.5cm]{\D/bind.jpg}\\[-1pt]{\fontsize{5.5}{6}\selectfont one of the 51 frames (4.0\,s)}\end{minipage}\hfill
\begin{minipage}[t]{2.9cm}\vspace{0pt}\sees 51 frames with numbered tracks, 4\,fps\\
\asks \pq{\ldots Each mug has a numbered box (1, 2, 3) \ldots\ Which numbered mug is the hidden object under or inside \textbf{right after it is hidden}?}\\
\rets \pq{2}\end{minipage}};
\node[wcard, text width=5.5cm, minimum height=2.75cm] (s5) at (6.0,-7.6) {\textbf{5a. Viewpoint fact}\hfill\vlmb{1}\\[1pt]
\begin{minipage}[t]{2.5cm}\vspace{0pt}\includegraphics[width=2.5cm]{\D/view.jpg}\end{minipage}\hfill
\begin{minipage}[t]{2.9cm}\vspace{0pt}\sees 4 evenly spaced frames\\
\asks \pq{Is the person who moves the objects positioned opposite the camera, on the other side of the table \ldots?}\\
\rets \pq{YES}: left and right are mirrored for the person\end{minipage}};
\node[wcard, text width=5.5cm, minimum height=2.75cm] (s6) at (11.85,-7.6) {\textbf{5b. Read out from facts}\hfill\vlmb{1}\\[1pt]
\begin{minipage}[t]{2.5cm}\vspace{0pt}\includegraphics[width=2.5cm]{\D/final.jpg}\end{minipage}\hfill
\begin{minipage}[t]{2.9cm}\vspace{0pt}\sees text only: \pq{The hidden object was hidden under mug \#2 \ldots\ from the camera's left to right: \#2, \#3, \#1 \ldots\ the person's left is the camera's right \ldots}, then the question\\
\rets \pq{\textbackslash boxed\{C\}}\quad\textbf{Answer: C} \cmark\end{minipage}};
\draw[warrow] (s1.east) -- (s2.west);
\draw[warrow] (s2.east) -- (s3.west);
\draw[warrow] (s3.south) -- ++(0,-0.08) -| (s4.north);
\draw[warrow] (s4.east) -- (s5.west);
\draw[warrow] (s5.east) -- (s6.west);
\end{tikzpicture}

%% file: figs/walk_card.tex
\begin{tikzpicture}[x=1cm,y=1cm]
\def\D{figs/ex3/card}
\node[wq, text width=17.45cm] (q) at (0,0) {\textbf{Question} (VET-Bench card monte, test video \texttt{card\_003}): \emph{Where is the Queen of Hearts at the end of the video?}\quad \textbf{(A)~Left} (ground truth) \quad (B)~Middle \quad (C)~Right};
\node[whead] at (0,-0.5) {\textbf{Input video}\\12.1\,s, 30\,fps};
\foreach \i/\t in {0/0.0\,s: queen face up, 1/1.0\,s: turned face down, 2/4.0\,s: shuffling, 3/8.0\,s, 4/12.0\,s: end} {
  \node[anchor=north west, inner sep=0] (in\i) at (1.75+3.15*\i,-0.47) {\includegraphics[width=2.6cm]{\D/in\i.jpg}};
  \node[wlabel, anchor=north] at (in\i.south) {\t};
}
\fill[bandbase, rounded corners=5pt] (0,-2.75) rectangle (17.8,-4.05);
\draw[wband=black] (0,-2.75) rectangle (17.8,-4.05);
\node[whead, text width=2.9cm] at (0.05,-2.8) {\textbf{Baseline: direct answer}\\Qwen3-VL-8B\\[1pt]\vlmb{1}\ \ no tool call};
\node[wcard, text width=4.6cm] (b1) at (3.2,-2.85) {\sees the whole video at 8\,fps (10.2k input tokens)\\\asks the benchmark prompt: question, options, \pq{Output your final answer (A, B, or C) in the format: \$\textbackslash boxed\{$<$option$>$\}\$.}};
\node[wcard, text width=6.6cm] (b2) at (8.35,-2.85) {\rets \pq{The Queen of Hearts is initially on the right side of the screen. At 0:01, it is flipped over \ldots\ the Queen of Hearts remains in the same position on the right side of the screen, even as the other cards move around. \ldots\ \textbackslash boxed\{C\}}};
\draw[warrow] (b1.east) -- (b2.west);
\node[whead, align=center] at (15.35,-3.0) {\textbf{Answer: C} \xmark};
\fill[bandours, rounded corners=5pt] (0,-4.3) rectangle (17.8,-10.6);
\draw[wband=blue] (0,-4.3) rectangle (17.8,-10.6);
\node[whead] at (0.05,-4.33) {\textbf{Ours: Track, Then Bind}\quad same Qwen3-VL-8B + SAM2.1 tracker\quad in total \vlmb{12} and \toolb{tracker $\times$4}, 7.3k input tokens};
\node[wcard, text width=5.5cm, minimum height=2.95cm] (s1) at (0.15,-4.85) {\textbf{1. Plan}\hfill\vlmb{2}\\[1pt]
\sees the whole video at 2\,fps; then the question alone\\
\rets target \pq{Queen of Hearts}, cue \pq{visible}, category \pq{playing cards}, viewpoint \pq{camera}; final location \pq{YES}.\\[2pt]
No carrier is named, so all \emph{playing cards} become carriers (the category rule of step~1).};
\node[wcard, text width=5.5cm, minimum height=2.95cm] (s2) at (6.0,-4.85) {\textbf{2. Perceive}\hfill\vlmb{8}\ \toolb{SAM2.1 $\times$4}\\[1pt]
\begin{minipage}[t]{2.3cm}\vspace{0pt}\includegraphics[width=2.3cm]{\D/ground.jpg}\\[-1pt]{\fontsize{5.5}{6}\selectfont 0.0\,s; pink: Queen of Hearts}\end{minipage}\hfill
\begin{minipage}[t]{3.1cm}\vspace{0pt}\sees 8 evenly spaced frames, one at a time\\
\asks \pq{Locate every instance of \ldots: Queen of Hearts; playing cards.}\\
\rets the queen on frame~0 only. SAM2.1 tracks the cards at 6 and 30\,fps.\end{minipage}};
\node[wcard, text width=5.5cm, minimum height=2.95cm] (s3) at (11.85,-4.85) {\textbf{3. Evidence moment}\hfill\nocall\\[1pt]
\includegraphics[width=1.77cm]{\D/trk0.jpg}\,\includegraphics[width=1.77cm]{\D/trk1.jpg}\,\includegraphics[width=1.77cm]{\D/trk2.jpg}\\
The queen is recognised by its look, so the evidence moments are the frames on which grounding found it: 0.0\,s. Tracks \#1 (red), \#2 (green), \#3 (blue), shown at 1.0, 4.0, 8.0\,s.};
\node[wcard, text width=5.5cm, minimum height=2.6cm] (s4) at (0.15,-7.95) {\textbf{4. Bind}\hfill\vlmb{1}\\[1pt]
\begin{minipage}[t]{2.3cm}\vspace{0pt}\includegraphics[width=2.3cm]{\D/bind.jpg}\\[-1pt]{\fontsize{5.5}{6}\selectfont one of the 6 frames (0.0\,s)}\end{minipage}\hfill
\begin{minipage}[t]{3.1cm}\vspace{0pt}\sees 6 frames with numbered tracks, 4\,fps, 0.0--1.3\,s\\
\asks \pq{\ldots Each playing card has a numbered box (1, 2, 3) \ldots\ Which numbered playing card is the Queen of Hearts?}\\
\rets \pq{3}\end{minipage}};
\node[wcard, text width=5.5cm, minimum height=2.6cm] (s5) at (6.0,-7.95) {\textbf{5. Read out from facts}\hfill\vlmb{1}\\[1pt]
\begin{minipage}[t]{2.3cm}\vspace{0pt}\includegraphics[width=2.3cm]{\D/final.jpg}\end{minipage}\hfill
\begin{minipage}[t]{3.1cm}\vspace{0pt}\sees text only: \pq{The Queen of Hearts is playing card \#3 \ldots\ from the camera's left to right they are: \#3, \#2, \#1 \ldots}, then the question\\
\rets \pq{\textbackslash boxed\{A\}}\quad\textbf{Answer: A} \cmark\end{minipage}};
\node[wcard, text width=5.5cm, minimum height=2.6cm, fill=bandours, draw=none] (s6) at (11.85,-7.95) {\textbf{Why the direct answer fails.} After 1.0\,s no frame tells the three cards apart. The model must carry identity through 11\,s of crossings, and it assumes the queen stayed in place. The agent asks the VLM only while the queen is face up; identity after that comes from the tracker.};
\draw[warrow] (s1.east) -- (s2.west);
\draw[warrow] (s2.east) -- (s3.west);
\draw[warrow] (s3.south) -- ++(0,-0.08) -| (s4.north);
\draw[warrow] (s4.east) -- (s5.west);
\end{tikzpicture}

%% file: figs/walk_toc.tex
\ifdefined\tairTLbox\else\newsavebox{\tairTLbox}\fi
\sbox{\tairTLbox}{\input{figs/ex3/toc/timeline_fig9.tex}}%
\begin{tikzpicture}[x=1cm,y=1cm]
\def\D{figs/ex3/toc}
\node[wq, text width=17.45cm] (q) at (0,0) {\textbf{Question} (TOC-Bench test question \texttt{8b0c33d38075}, dimension \emph{event existence}): \emph{Does eggs ever get fully hidden in the video?}\quad A.~eggs never gets fully hidden\quad \textbf{B.~eggs does get fully hidden} (ground truth)};
\node[whead] at (0,-0.78) {\textbf{Input video}\\26.2\,s, 30\,fps};
\foreach \i/\t in {0/0.5\,s: eggs on a plate, 1/1.6\,s: a bottle passes, 2/2.6\,s: eggs hidden, 3/12.0\,s, 4/23.0\,s: hidden by a hand} {
  \node[anchor=north west, inner sep=0] (in\i) at (1.75+3.15*\i,-0.75) {\includegraphics[width=3.05cm]{\D/in\i.jpg}};
  \node[wlabel, anchor=north] at (in\i.south) {\t};
}
\fill[bandbase, rounded corners=5pt] (0,-2.75) rectangle (17.8,-3.85);
\draw[wband=black] (0,-2.75) rectangle (17.8,-3.85);
\node[whead, text width=2.9cm] at (0.05,-2.8) {\textbf{Baseline: direct answer}\\Qwen3-VL-8B\\[1pt]\vlmb{1}\ \ no tool call};
\node[wcard, text width=7.6cm] (b1) at (3.2,-2.85) {\sees 32 uniformly sampled frames (TOC-Bench's own sampler)\\\asks TOC-Bench's prompt: \pq{You will see a video and two statements. Choose the statement that is TRUE according to the video. Output ONLY the letter A or B. Question: \ldots\ A.~\ldots\ B.~\ldots\ Answer:}};
\node[wcard, text width=3.0cm, anchor=west] (b2) at ([xshift=4mm]b1.east) {\rets \pq{A}};
\draw[warrow] (b1.east) -- (b2.west);
\node[whead, align=center] at (15.35,-3.0) {\textbf{Answer: A} \xmark};
\fill[bandours, rounded corners=5pt] (0,-4.1) rectangle (17.8,-11.3);
\draw[wband=blue] (0,-4.1) rectangle (17.8,-11.3);
\node[whead] at (0.05,-4.13) {\textbf{Ours: Track, Then Bind}\quad same Qwen3-VL-8B + SAM2.1 tracker\quad in total \vlmb{11} and \toolb{tracker $\times$1}, 20.5k input tokens};
\node[wcard, text width=5.5cm, minimum height=2.95cm] (s1) at (0.15,-4.65) {\textbf{1. Plan}\hfill\vlmb{2}\\[1pt]
\sees the whole video at 2\,fps\\
\asks the plan schema (as in Fig.~\ref{fig:walk}); then, text only, whether the question asks for a final location\\
\rets entity \pq{eggs}; final location \pq{NO}. No target has to be told apart from look-alikes, so steps~3 and~4 are skipped.};
\node[wcard, text width=5.5cm, minimum height=2.95cm] (s2) at (6.0,-4.65) {\textbf{2. Perceive: ground}\hfill\vlmb{8}\\[1pt]
\begin{minipage}[t]{2.5cm}\vspace{0pt}\includegraphics[width=2.5cm]{\D/ground.jpg}\end{minipage}\hfill
\begin{minipage}[t]{2.9cm}\vspace{0pt}\sees 8 evenly spaced frames, one at a time\\
\asks \pq{Locate every instance of the following objects in the image, if visible: eggs. \ldots}\\
\rets a box on all 8 frames; SAM2.1 is prompted with one of them\end{minipage}};
\node[wcard, text width=5.5cm, minimum height=2.95cm] (s3) at (11.85,-4.65) {\textbf{2. Perceive: track}\hfill\toolb{SAM2.1 $\times$1}\\[1pt]
\includegraphics[width=1.77cm]{\D/trk0.jpg}\,\includegraphics[width=1.77cm]{\D/trk1.jpg}\,\includegraphics[width=1.77cm]{\D/trk2.jpg}\\
The mask of the eggs at 6\,fps (157 frames), forward and backward from the prompt frame. Shown: 0.5, 2.6 and 12.0\,s; at 2.6\,s the bottle covers the eggs and the mask nearly vanishes.};
\node[wcard, text width=5.5cm, minimum height=3.1cm] (s4) at (0.15,-8.15) {\textbf{5. Read out: timeline}\hfill\nocall\\[1pt]
\usebox{\tairTLbox}\\[2pt]
Mask area against its 90th percentile gives three states: dark green \emph{fully visible} ($\ge$0.6), light green \emph{partly hidden} (0.12--0.6), grey \emph{not visible}. Both gaps end with the eggs back in place, so they were blocked, not removed.};
\node[wcard, text width=11.35cm, minimum height=3.1cm] (s5) at (6.0,-8.15) {\textbf{5. Read out: answer}\hfill\vlmb{1}\\[1pt]
\sees the same 32 frames as the baseline\\
\asks the same TOC-Bench prompt, with the timeline inserted after the question:\\
\pq{\ldots Question: Does eggs ever get fully hidden in the video?\newline Automatic tracker evidence (may contain errors; the video frames take precedence):\newline Video length 26.2 s. 'eggs': 0.0-1.3 s fully visible; 1.3-2.0 s partly hidden; 2.0-3.3 s not visible (vanished inside the frame, so it was blocked or hidden); 3.3-22.2 s fully visible; 22.2-24.3 s not visible (\ldots); \ldots\ Events in order: 1.3 s gets partly hidden; 2.0 s gets fully hidden; 3.3 s comes back into view; 22.2 s gets fully hidden; 24.3 s comes back into view. \ldots\newline A. eggs never gets fully hidden\quad B. eggs does gets fully hidden\quad Answer:}\\[2pt]
\rets \pq{B}\hfill\textbf{Answer: B} \cmark};
\draw[warrow] (s1.east) -- (s2.west);
\draw[warrow] (s2.east) -- (s3.west);
\draw[warrow] (s3.south) -- ++(0,-0.08) -| (s4.north);
\draw[warrow] (s4.east) -- (s5.west);
\end{tikzpicture}